\documentclass[]{servicenow}

\usepackage{amsmath,amsfonts,bm}

\def\eqref#1{equation~\ref{#1}}

\def\1{\bm{1}}

\DeclareMathAlphabet{\mathsfit}{\encodingdefault}{\sfdefault}{m}{sl}
\SetMathAlphabet{\mathsfit}{bold}{\encodingdefault}{\sfdefault}{bx}{n}

\DeclareMathOperator*{\argmax}{arg\,max}

\usepackage{preamble}

\usepackage{hyperref}
\usepackage[inline]{enumitem}
\usepackage{graphicx}
\usepackage{url}
\usepackage{subfig}
\usepackage{nicefrac}
\usepackage[table]{xcolor}
\usepackage{array}
\usepackage{booktabs}

\title{TIPS: A Tiny Inference-time Policy\\
for Scaling in Diffusion Language Models}

\author[1,2,\dagger]{Ya-Shi Zhang}
\author[2,3]{Jian Tang}
\author[1]{Pierre-André Noël}
\author[1]{Oleksiy Ostapenko}

\affiliation[1]{ServiceNow AI Research}
\affiliation[2]{Mila - Quebec AI Institute}
\affiliation[3]{HEC Montréal}

\contribution[\dagger]{Work done at ServiceNow}

\abstract{
Masked diffusion language models expose an inference-time choice: which positions become visible after each denoising pass.
Many existing decoding methods use hand-designed rules or require visibility to grow monotonically.
We present a \textbf{T}iny \textbf{I}nference-time \textbf{P}olicy for \textbf{S}caling (\textbf{TIPS}), an approximately $600$K-parameter controller that learns both unmasking and remasking while keeping the underlying language model frozen.
TIPS uses visibility-aware confidence features and a shallow transformer to select which tokens should be visible, while a fixed schedule controls the expected visible fraction.
We train the controller through supervised warmup followed by reinforcement learning from terminal correctness rewards.
On GSM8K and MATH-500 with LLaDA-8B-Instruct, the full-canvas and two-block TIPS variants achieve $58.46\%$ and $57.97\%$ micro-averaged accuracy at a budget of $32$ decoder evaluations, exceeding the strongest baseline in our main comparison by $12.1$ and $11.6$ percentage points, respectively.
Both variants are trained at a budget of $32$ evaluations and remain competitive at higher budgets without retraining.
These results demonstrate that learning when to reveal and reconsider tokens can improve the accuracy--compute trade-off without updating the decoder backbone.
}

\correspondence{\email{ya-shi.zhang@mila.quebec}}

\begin{document}
\maketitle

\section{Introduction}
\label{sec:introduction}

Masked diffusion language models (dLLMs) generate responses by predicting unresolved positions in parallel and exposing selected predictions as context \citep{sahoo2024simple,nie2025llada}.
This flexibility creates a decoding problem: under a limited number of denoising passes, a sampler must decide which predictions to commit and whether earlier commitments should be reconsidered.
Revealing more positions makes faster progress toward a complete response, but those predictions also condition subsequent generations.
An early commitment can therefore influence later predictions even after the surrounding context changes.
The resulting \emph{visibility trajectory}---which tokens become context, when they appear, and whether they can be revised---is an important component of the accuracy--compute trade-off \citep{kim2025ordering}.
We study how to learn this trajectory while leaving the pretrained token model unchanged.

Existing methods address different parts of this problem.
Confidence-based and uncertainty-aware samplers select positions using hand-designed rules \citep{wu2025fastdllm,benhamu2025entropy,huang2025uncode}.
Learned unmasking policies instead optimize position selection around a frozen denoiser, but typically preserve monotonic visibility: once a token is exposed, the policy cannot withdraw it \citep{jazbec2026learning,xu2026scheduling,hong2026upo}.
Conversely, remasking methods allow revision through prescribed transitions or heuristic criteria \citep{wang2025remasking,hong2025wino,zhai2026core}, while other approaches learn remasking behavior by updating the language model itself \citep{huang2025don}.
These developments motivate a complementary question: can a small auxiliary policy learn both commitment and revision without modifying the dLLM?

We introduce \textbf{T}iny \textbf{I}nference-time \textbf{P}olicy for \textbf{S}caling (\textbf{TIPS}), an approximately $600$K-parameter controller assigning to each response position a binary next-step visibility action.
At masked positions, the action reveals the decoder's proposal or leaves the position masked; at visible positions, it retains the current token or remasks it for subsequent regeneration.
A two-layer transformer predicts visibility preferences from confidence features, mask indicators, and normalized decoding time.

TIPS separates decoding \emph{order} from \emph{pace}.
A fixed power-law schedule specifies the expected visible fraction, while a shared shift of the policy logits matches this target without changing the policy's relative preferences.
Thus, the expected amount of context grows according to the schedule even as individual commitments remain reversible.
Our training recipe first provides dense, reference-derived supervision of visibility decisions, then optimizes final-answer correctness through group-relative policy optimization \citep{shao2024deepseekmath}.
Only the auxiliary controller is updated in either stage; training requires neither gradients through the dLLM nor its sequence likelihood.
The same visibility-control formulation applies to full-canvas and semi-autoregressive decoding (\cref{sec:method}).

We summarize our contributions as follows:
\begin{itemize}
    \item
    We introduce TIPS, an approximately $600$K-parameter policy that learns both unmasking and remasking while keeping the dLLM backbone frozen.

    \item 
    We instantiate TIPS with full-canvas and blockwise decoding on LLaDA-8B-Instruct.
    At most decoder NFE budgets, we surpass previous state-of-the-art methods in accuracy on GSM8K and MATH-500 by up to $13.25$ percentage points, otherwise remaining competitive. 

    \item
    We investigate how TIPS uses commitment and revision through ablations and other supplementary experiments. 
    These analyses highlight the importance of remasking, suggesting its benefits are not only token replacement, but also temporarily withdrawing harmful context.
    We further characterize the scope and limits of TIPS through transfer experiments.
\end{itemize}

\section{Background}
\label{sec:background}

We focus on finite-step decoding with a pretrained masked diffusion language model (dLLM).
For fundamental principles %
and backbone training objectives, we refer readers to
\citet{austin2021structured,sahoo2024simple,shi2024simplified,nie2025llada,ou2025absorbing,arriola2025block}.

\subsection{Finite-step decoding}
\label{sec:background-decoding}

Let $x_t\in\calV^L$ denote the partially generated response after $t$ denoising passes, let
\begin{equation}
    \mathcal{M}_t
    =
    \{i:x_{t,i}=\mask\},
    \qquad
    \mathcal{S}_t
    =
    \{i:x_{t,i}\neq\mask\}
    \label{eq:masked-visible-sets}
\end{equation}
be the masked and visible positions, respectively. 
A dLLM predicts a vocabulary distribution at every position and the reverse process samples a candidate token at each masked position:
\begin{equation}
    p_{t,i}(v)
    =
    p_\theta(v\mid q,x_t),
    \qquad
    \widetilde{x}_{t,i}
    \sim
    p_{t,i}(\cdot),
    \quad
    i\in\mathcal{M}_t.
    \label{eq:default-candidates}
\end{equation}
Note that token sampling is often controlled ad-hoc via temperature scaling, where temperature $0$ denotes top-1 or greedy decoding. 
A decoding algorithm then chooses the set $\mathcal{S}_{t+1}\subseteq\{1,\ldots,L\}$ of positions that will be visible in the next state and applies
\begin{equation}
    x_{t+1,i}
    =
    \begin{cases}
        \mask,
        & i\notin\mathcal{S}_{t+1},\\
        x_{t,i},
        & i\in\mathcal{S}_t\cap\mathcal{S}_{t+1},\\
        \widetilde{x}_{t,i},
        & i\in\mathcal{M}_t\cap\mathcal{S}_{t+1}.
    \end{cases}
    \label{eq:general-visibility-update}
\end{equation}
A visible token may be kept or re-masked, while a masked position may be revealed or left masked. Many decoding algorithms impose the monotonic constraint
\(\mathcal{S}_t\subseteq\mathcal{S}_{t+1}\). 
Under this restriction, tokens that are already unmasked can no longer be modified. Re-masking and editing methods remove this monotonicity constraint, allowing an earlier prediction to return to the masked state or directly transition to another vocabulary token after the surrounding context changes.

Hence, masked-diffusion inference separates two choices: \emph{what} token to place, and \emph{which} positions should be visible. The first choice is supplied by the dLLM backbone. The second is supplied by the decoding algorithm. Because visible tokens become context at subsequent passes, different visibility trajectories can produce different final responses for the same model and number of forward passes.

\subsection{The decoding path as a Markov decision process}
\label{sec:background-mdp}

A decoding path can be viewed as a finite-horizon Markov decision process \citep{jazbec2026learning}. 
At decision step $t$, the state contains the prompt, current partially generated response, and denoising time. One dLLM evaluation produces the candidate distributions in \cref{eq:default-candidates}, and an action is represented by the next visible set $\mathcal{S}_{t+1}$ or -- equivalently -- by a binary next-step visibility decision per position, denoted $a_t\in\{0,1\}^L$, where $1$ denotes unmasked. 
The transition then applies \cref{eq:general-visibility-update}.
For non-zero temperatures, the transition is stochastic when conditioning on the action; with greedy decoding it is deterministic given the state and action. We assume greedy decoding from here on out.

\section{Method}
\label{sec:method}

\begin{algorithm}[t]
    \caption{TIPS: Inference}
    \label{alg:tips-decoding}
    \begin{algorithmic}[1]
        \Require Prompt $q$, response length $L$, decoder budget $T\geq 1$
        \Require Frozen decoder $p_\theta$, TIPS $\pi_\phi$,
                 fixed schedule parameter $\gamma\in\mathbb{R}$
        \Ensure Fully visible response $x_T$ using exactly $T$ dLLM evaluations

        \State $x_0 \gets (\mask,\ldots,\mask)$

        \For{$t=0,\ldots,T-2$}
            \Comment{No early stopping; empty when $T=1$}
            \State $p_t \gets p_\theta(\cdot\mid q,x_t)$
                \Comment{One dLLM evaluation}
            \State $m_{t,i} \gets
                \mathbf{1}\{x_{t,i}=\mask\}$ for all $i$
            \State Compute $C_t$ using \cref{eq:tips-features}
            \State $b_t \gets \pi_\phi(C_t,m_t,t/T)$
            \State $\kappa_t \gets
                \bigl((t+1)/T\bigr)^{\exp(\gamma)}$
            \State Compute $\delta_t$ s.t.
                $\frac{1}{L}\sum_{i=1}^{L}
                \sigma(b_{t,i}+\delta_t)=\kappa_t$
            \State $\rho_{t,i} \gets
                \sigma(b_{t,i}+\delta_t)$ for all $i$
            \State Sample $a_{t,i}\sim
                \operatorname{Bernoulli}(\rho_{t,i})$ for all $i$
            \State Update all positions simultaneously:\vspace{-1.5\bigskipamount}
            \begin{equation*}
                \qquad\qquad\qquad x_{t+1,i}=
                \begin{cases}
                    \mask, & a_{t,i}=0,\\
                    x_{t,i}, & a_{t,i}=1,\ m_{t,i}=0,\\
                    \widehat{x}_{t,i}, & a_{t,i}=1,\ m_{t,i}=1.
                \end{cases}
            \end{equation*}
            \vspace{-\bigskipamount}
        \EndFor

        \State $p_{T-1} \gets
            p_\theta(\cdot\mid q,x_{T-1})$
            \Comment{Final dLLM evaluation}
        \State $\widehat{x}_{T-1,i} \gets
            \argmax_{v\in\mathcal{V}}
            p_{T-1,i}(v)$ for all $i$
        \State Fill remaining masks: \vspace{-1.5\bigskipamount}
        \begin{equation*}
            \qquad\qquad\qquad x_{T,i}
            =
            \begin{cases}
                x_{T-1,i},
                & x_{T-1,i}\neq\mask,\\
                \widehat{x}_{T-1,i},
                & x_{T-1,i}=\mask.
            \end{cases}
        \end{equation*}
        \vspace{-\bigskipamount}
        \State \Return $x_T$
    \end{algorithmic}
\end{algorithm}

TIPS is a lightweight auxiliary policy network $\pi_\phi$ that controls the inference trajectory of a frozen dLLM. 
Following the Bernoulli action parameterization of \citet{jazbec2026learning}, the policy maps decoding-state features to position-wise binary decisions.
TIPS extends unmasking-only decisions to \emph{next-step visibility}: an action can reveal or defer a masked position, or retain or remask an already visible token.
TIPS is also \emph{fixed-horizon}: it does not terminate early when all tokens are unmasked. Subsequent policy actions may remask tokens for revision. 
See \cref{alg:tips-decoding} for details.%

One dLLM forward pass produces probabilities over non-mask tokens 
$p_{t,i}(v)\coloneqq p_\theta(v\mid q,x_t)$.
Define mask indicator $m_{t,i}=\mathbf{1}\{x_{t,i}=\mask\}$ and greedy candidate $\widehat{x}_{t,i}=\argmax_{v\in\calV} p_{t,i}(v)$. 

\paragraph{Featurization. }
For every position, define the \emph{active token}:
\begin{equation}
    v^{\mathrm{act}}_{t,i}
    =
    \begin{cases}
        \widehat{x}_{t,i},
        & m_{t,i}=1,\\
        x_{t,i},
        & m_{t,i}=0,
    \end{cases}
    \label{eq:active-token}
\end{equation}
with corresponding probability $P^{\mathrm{act}}_{t,i}=p_{t,i}(v^{\mathrm{act}}_{t,i})$.
At a masked position, the active token is the decoder's greedy proposal. At an unmasked position, it is the already decoded token, even if the token is not the decoder's argmax. This is done to ensure \emph{visibility consistency}: always measuring support for the token whose visibility the policy controls. 
For more details, refer to \cref{app:exp-visibility-consistency}. 
We also define the strongest alternative
\begin{equation}
    v^{\mathrm{alt}}_{t,i}
    =
    \argmax_{
        v\in
        \mathcal{V}
        \setminus
        \{v^{\mathrm{act}}_{t,i}\}
    }
    p_{t,i}(v),
    \label{eq:alternative-token}
\end{equation}
with probability
\(
    P^{\mathrm{alt}}_{t,i}
    =
    p_{t,i}\!\left(v^{\mathrm{alt}}_{t,i}\right),
\)
which is the most probable token other than the active token. 

Now, we construct the two-dimensional per-token feature 
\begin{equation}
    C_{t,i}
    =
    \left[
        \log
        \frac{
            {P}^{\mathrm{act}}_{t,i}
        }{
            1-{P}^{\mathrm{act}}_{t,i}
        },
        \;
        \log
        \frac{
            {P}^{\mathrm{act}}_{t,i}
        }{
            {P}^{\mathrm{alt}}_{t,i}
        }
    \right].
    \label{eq:tips-features}
\end{equation}
The first coordinate is the active token's log-odds, whose parametrization can distinguish changes near the low- and high-probability extremes more clearly than raw probability. The second coordinate measures \emph{relative support} of the active token against the most likely competing token. This log-ratio is negative if and only if the decoder prefers a different token over the current visible token. It is also important to mention the two features are on the same numerical scale. 

A two-layer RoPE-DiT \citep{peebles2023scalable,su2024roformer} policy network produces per-position visibility logits $b_t=\pi_\phi(C_t,m_t,t/T)$, representing the policy's relative preference for each position's visibility (see \cref{app:policy-network-architecture}). These logits determine \emph{which} positions are preferred while also allowing an external unmasking schedule to control \emph{how many} positions become visible. 

\paragraph{Scheduled visibility. }
We observe in \cref{sec:ablation-schedule} a network that is free to determine the total visible fraction leads to a lazy strategy we hereafter refer to as \emph{finalization deferral}: keeping most positions masked until the final forward pass, where all are unmasked at once. This prevents the decoder from conditioning on progressively richer contexts, and leads to worse task performance. 
Therefore, we introduce a scalar shift to the output logits so that the adjusted logits adhere to the schedule in expectation. 

For normalized decoding time $\tau\in[0,1]$, define the family $\kappa(\tau)=\tau^{\exp(\gamma)},\,\gamma\in\mathbb{R}$. 
For $\gamma>0$, the curve is convex: visibility grows slowly
at first and more rapidly toward the end of decoding.
For $\gamma<0$, it is concave, exposing more tokens early;
$\gamma=0$ recovers a linear schedule.
In all cases, $\kappa(0)=0$ and $\kappa(1)=1$, corresponding to completely masked and completely unmasked states at the beginning and end of decoding, respectively.
Thus, for a budget of T dLLM evaluations and parameter $\gamma$, the target fraction of visible tokens for the next state is $\kappa_t=\left((t+1)/T \right)^{\exp(\gamma)}$, with $t=0,\ldots,T-2$.
To achieve this ratio in practice without changing local preference, we find a shared scalar $\delta_t$ such that $\tfrac{1}{L}\sum_{i=1}^L \sigma(b_{t,i}+\delta_t)=\kappa_t$,
with calibrated next-state visibility probabilities $\rho_{t,i}=\sigma(b_{t,i}+\delta_t).$ 
As the left hand side is continuous and strictly monotonic in $\delta_t$, 
it has a unique solution, which we solve for numerically via Newton iteration. 
For more details, refer to \cref{eq:newton-calibration}.
The same scalar is added to every position, preserving the policy's preference ordering. The realized number of visible positions remains stochastic as TIPS samples $a_{t,i}\sim\mathrm{Bernoulli}(\rho_{t,i})$, where $a_{t,i}$ has the state-independent interpretation of 
\begin{equation}
    \label{eq:action-interp}
    a_{t,i}=1\iff \mathrm{position}\ i\ \mathrm{is\ visible\ in}\ x_{t+1},
\end{equation}
while following the schedule in expectation, i.e., $\mathbb{E}\left[\tfrac{1}{L}\sum_{i=1}^La_{t,i} \middle| x_t\right]=\kappa_t$.
 
\paragraph{Policy optimization. }
As in \citet{jazbec2026learning}, we use conditionally independent Bernoulli actions with a tractable factorized likelihood.
For TIPS's next-step visibility actions, the joint log-probability is
\begin{equation}
    \label{eq:tips-logprob}
    \log\mathrm{Pr}_\phi(a_t\mid s_t)=\sum_{i=1}^L\left[ 
    a_{t,i}\log\rho_{t,i}+(1-a_{t,i})\log(1-\rho_{t,i})
    \right],
\end{equation}
where $s_t=(C_t,m_t,t/T)$ denotes the policy state. During training, we sample a group of $G$ visibility trajectories with final responses to obtain a binary reward $R_g$ from a verifier. 
We can then optimize the policy network with GRPO \citep{schulman2017ppo,shao2024deepseekmath}:
\begin{equation}
\begin{aligned}
    \mathcal{L}_{\mathrm{RLVR}}(\phi)
    =
    -\frac{1}{G(T-1)}
    \sum_{g=1}^{G}
    \sum_{t=0}^{T-2}
    \min\!\Big(
        w_{t,g}(\phi)A_g,\operatorname{clip}
        \left(w_{t,g}(\phi),0.5,1.5\right)A_g
    \Big),
\end{aligned}
    \label{eq:onpolicy-grpo}
\end{equation}
where $w_{t,g}(\phi)= \tfrac{\mathrm{Pr}_\phi(a_t^{(g)}\mid s_t^{(g)})}{\mathrm{Pr}_{\mathrm{old}}(a_t^{(g)}\mid s_t^{(g)})}$ is the likelihood ratio and $A_g= R_g-\frac{1}{G}\sum_{g=1}^G R_g$ is the group-relative advantage. 

\paragraph{Supervised warmup. }
If reference response data is available, then we can optionally warmup the policy network before policy optimization by providing dense supervision of network logits. We collect intermediate decoding states from rollouts, then construct a target using the reference tokens. Given a reference response $y$ and a masked position, we define the probability of the reference token and the strongest non-reference token as $q_{t,i}=p_{t,i}(y_i)$ and $\widehat{q}_{t,i}=\max_{v\in\mathcal{V}\setminus\{y_i\}}p_{t,i}(v)$, respectively. Now, we assign a \emph{visibility utility} to each position:
\begin{equation}
    u_{t,i}
    =
    \begin{cases}
        +1,
        & m_{t,i}=0,\;x_{t,i}=y_i,\\
        -1,
        & m_{t,i}=0,\;x_{t,i}\neq y_i,\\
        \frac{q_{t,i}-\widehat{q}_{t,i}}{q_{t,i}+\widehat{q}_{t,i}},
        & m_{t,i}=1.
    \end{cases}
    \label{eq:warmup-utility}
\end{equation}
At visible positions, this utility favors retaining tokens that match the reference response and remasking those that are mismatching. At masked positions, it measures the decoder's support for the reference token relative to its strongest competitor. The masked utility can also be written as $\tanh\left(\frac{1}{2}\log{q_{t,i}}/{\widehat{q}_{t,i}}\right)\in[-1,1]$. The utilities are converted into visibility targets $\rho^\star_{t,i}=\mathrm{sg}[\sigma(\beta u_{t,i}+\delta^\star_t)]$, where $\beta>0$ controls sharpness and $\mathrm{sg}[\cdot]$ is the stopgrad operator. We minimize the binary cross-entropy between the policy probabilities and the targets (details in \cref{app:supervised-warmup-config})
\begin{equation}
    \label{eq:bce-supervised-loss}
    \mathcal{L}_\mathrm{warmup}(\phi)=
    -\frac{1}{L}\sum_{i=1}^L\left[
    \rho^\star_{t,i}\log\rho_{t,i}
    +(1-\rho^\star_{t,i})\log(1-\rho_{t,i})
    \right].
\end{equation}

\paragraph{Semi-autoregressive decoding. }
TIPS generalizes to blockwise decoding trivially by treating each block as the full canvas \citep{arriola2025block}. For a total budget of $T$ dLLM evaluations split across $B$ blocks, we allocate $T/B$ calls to each block. For each block, TIPS is active for the first $T/B-1$ calls before the deterministic final forward pass that unmasks all remaining mask tokens. We train and evaluate two variants of TIPS, full-canvas and $B=2$, referred to as $\text{TIPS}_{\text{full}}$ and $\text{TIPS}_{\text{block}}$. 

\section{Experiments}
\label{sec:experiments}

\subsection{Setup}
\label{sec:experimental-setup}

\paragraph{Backbone and tasks. }
All experiments use the frozen LLaDA-8B-Instruct decoder with a 256-token response length \citep{nie2025llada} unless stated otherwise. 
We train the models on the GSM8K and MATH training sets and evaluate on the GSM8K test set and MATH-500, containing 1,319 and 500 problems, respectively \citep{cobbe2021gsm8k, hendrycks2021math, lightman2024verify}. For all methods, we report the accuracy over the 1,819 problems (i.e., micro-average over both test sets). 

\paragraph{Training. }
We train TIPS across three seeds with fixed decoding budget $T_{\mathrm{train}}=32$, beginning with the supervised warmup followed by 935 RLVR updates. 
For the choice of hyper-parameter $\gamma$, we first roll out the policy on a validation set without supervised warmup or logit calibration and choose $\gamma$ by least-squares projection of its mean expected visibility trajectory onto the family $\tau^{\exp{(\gamma})}$ (details in \cref{app:mse-projection-gamma}). We get $\gamma_{\mathrm{full}}\approx0.145$ and $\gamma_{\mathrm{block}}\approx0.187$.

\begin{table}[htbp]
\centering
\caption{
Micro-averaged test accuracy (\%) on GSM8K and MATH-500
(1,819 problems) at different decoder-call budgets.
For compactness, we display a subset of baselines that preserves
the highest or competitive accuracy at every evaluated budget across
the full comparison suite.
TIPS accuracy entries show the mean with $\pm$ standard error
underneath, across three training-seed means; each seed mean
averages three inference draws.
AUC is the normalized area under the accuracy versus
$\log_2$-budget curve over budgets 4--256, computed using
trapezoidal quadrature and reported on the same 0--100 scale
as accuracy. TIPS AUC values are computed from the mean
accuracy curves.
Bold indicates the highest accuracy among all evaluated
configurations in each budget column; in the AUC column,
it indicates the highest value among the displayed methods.
Blocks denotes number of blocks $\times$ tokens per block.
The policy from \citet{jazbec2026learning} corresponds to
the expert-steering variant.
Fast-dLLM's unavailable 4-call accuracy is imputed as its 8-call accuracy.}
\label{tab:main-results}
\begingroup
\definecolor{tipsfull}{HTML}{293E40}
\definecolor{tipsblock}{HTML}{80B6A1}
\footnotesize
\setlength{\tabcolsep}{2.8pt}
\renewcommand{\arraystretch}{1.12}
\begin{tabular*}{\linewidth}{@{\extracolsep{\fill}}lc*{8}{r}@{}}
\toprule
Method & Blocks & 4 & 8 & 16 & 32 & 64 & 128 & 256 & AUC \\
\midrule
Fast-dLLM
& $8\!\times\!32$
& 17.48
& 17.48
& 14.73
& 37.93
& \textbf{65.59}
& 67.01
& 67.01
& 40.83 \\
WINO
& $1\!\times\!256$
& \textbf{18.14}
& 21.06
& 34.80
& 46.34
& 49.26
& 49.37
& 49.37
& 39.10 \\
UNCODE
& $1\!\times\!256$
& 12.26
& 14.35
& 21.50
& 29.74
& 53.66
& 63.66
& 67.73
& 37.15 \\
\citet{jazbec2026learning}
& $1\!\times\!256$
& 5.63
& 7.42
& 9.18
& 12.39
& 47.72
& 64.80
& 64.63
& 29.44 \\
\addlinespace[2pt]
\rowcolor{tipsfull!7}
\textcolor{tipsfull}{\textbf{TIPS}$_{\mathrm{full}}$}
& $1\!\times\!256$
& \shortstack[r]{16.46\\[-1.5pt]{\scriptsize $\pm$1.79}}
& \shortstack[r]{\textbf{29.12}\\[-1.5pt]{\scriptsize $\pm$1.68}}
& \shortstack[r]{\textbf{48.05}\\[-1.5pt]{\scriptsize $\pm$0.73}}
& \shortstack[r]{\textbf{58.46}\\[-1.5pt]{\scriptsize $\pm$0.10}}
& \shortstack[r]{63.09\\[-1.5pt]{\scriptsize $\pm$0.50}}
& \shortstack[r]{65.32\\[-1.5pt]{\scriptsize $\pm$0.55}}
& \shortstack[r]{66.28\\[-1.5pt]{\scriptsize $\pm$0.71}}
& \textbf{50.90} \\
\rowcolor{tipsblock!19}
\textcolor{tipsfull}{\textbf{TIPS}$_{\mathrm{block}}$}
& $2\!\times\!128$
& \shortstack[r]{12.53\\[-1.5pt]{\scriptsize $\pm$0.45}}
& \shortstack[r]{20.95\\[-1.5pt]{\scriptsize $\pm$0.99}}
& \shortstack[r]{42.12\\[-1.5pt]{\scriptsize $\pm$2.04}}
& \shortstack[r]{57.97\\[-1.5pt]{\scriptsize $\pm$1.23}}
& \shortstack[r]{64.49\\[-1.5pt]{\scriptsize $\pm$0.80}}
& \shortstack[r]{\textbf{67.12}\\[-1.5pt]{\scriptsize $\pm$0.28}}
& \shortstack[r]{\textbf{68.18}\\[-1.5pt]{\scriptsize $\pm$0.36}}
& 48.83 \\
\bottomrule
\end{tabular*}
\endgroup
\end{table}

\begin{figure}[ht]
    \centering
    \includegraphics[width=\linewidth]{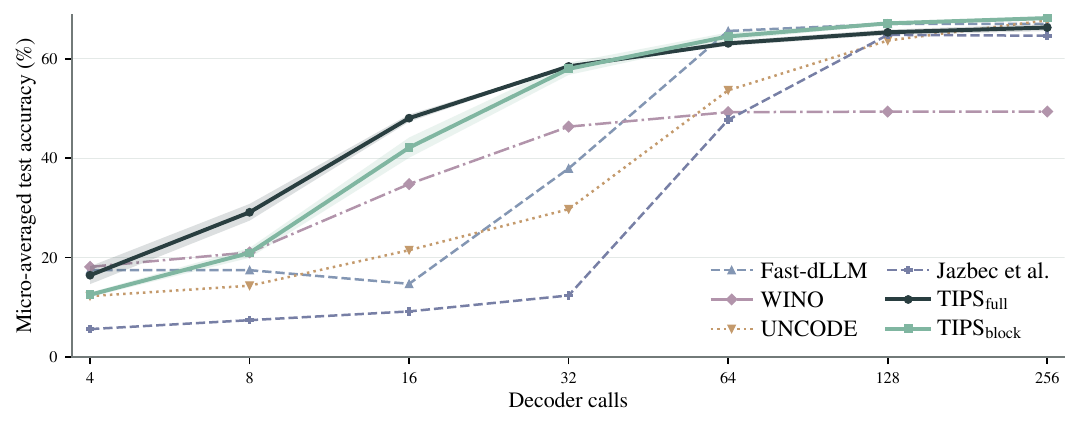}
    \caption{ 
        \textbf{Accuracy--compute tradeoff on mathematical reasoning.}
        Micro-averaged test accuracy on GSM8K and MATH-500 (1,819 problems)
        versus total decoder-calls, shown on a logarithmic scale.
        We compare $\mathrm{TIPS}_{\mathrm{full}}$ ($1\times256$) and
        $\mathrm{TIPS}_{\mathrm{block}}$ ($2\times128$) 
        against Fast-dLLM ($8\times32$), WINO, UNCODE, and the learned unmasking policy of \citet{jazbec2026learning} with expert-steering; the latter three use $1\times256$
        full-canvas decoding.
        The displayed baselines include a highest or competitive accuracy baseline at every evaluated budget from the full comparison suite.
        TIPS curves show means with shaded bands indicating $\pm1$ standard
        error across three training-seed means, each averaged over three
        inference draws. Baselines are shown as point estimates.
        Other adaptive baselines may use fewer calls than the specified budget.
    }
    \label{fig:main-results}
\end{figure}

\paragraph{Baselines. }
We reproduce and compare with a broad suite of baselines. To preserve visual fidelity, we show only a subset of methods, identified by the highest-performing or competitive baselines at each evaluated decoder-call budget to include them in \cref{fig:main-results,tab:main-results}.

This includes WINO \citep{hong2025wino}, Fast-dLLM's confidence-based decoder \citep{wu2025fastdllm}, and UNCODE \citep{huang2025uncode}.
We additionally include the unmasking-only policy from \citet{jazbec2026learning} as a learned-policy comparator.
Results for all evaluated baseline configurations appear in \cref{app:main-results}, with implementation details in \cref{app:main-experiment-details}.

All methods use the exact same setting except for block decomposition, in which we followed the settings found in the method's original manuscripts. For each method, we evaluate the performance at exponentially increasing budgets $T\in\{4,8,16,32,64,128,256\}$. Fixed-horizon methods will use exactly the requested number of decoder calls. For Fast-dLLM, $T$ is a target mean NFE (refer to \cref{app:fastdlm-budget-matching}); other adaptive methods use decoder-call caps and may terminate earlier.

\paragraph{Metrics. }
We primarily measure the micro-averaged accuracy of each method reported across different decoder-call budgets. In \cref{tab:main-results}, we additionally report the standard error of our method's accuracies across three seeds. Further details are given in \cref{app:main-results}. 

Additionally, 
to summarize performance across decoding budgets using a single numerical value, we also estimate the \emph{area under the accuracy log-budget curve} (AUC). Let $\mathrm{Acc}(k)$ be the method's accuracy at NFE budgets $k\in\{4,8,16,32,64,128,256\}$, we use a piecewise-linear interpolation in $\log_2\mathrm{NFE}$ and trapezoidal quadrature, which simplifies to 
\[
    \mathrm{AUC}=\frac{1}{\log_2256-\log_24}\int_{\log_24}^{\log_2256}\mathrm{Acc}(2^u)\,\mathrm{d}u=
    \frac{1}{6}
\sum_{k\in\{4,8,\ldots,128\}}
\frac{\mathrm{Acc}(k)+\mathrm{Acc}(2k)}{2}.
\]
If the method is adaptive and finishes decoding before using up the allocated NFE budget, its highest NFE accuracy is continued as a constant until $k=256$ as the additional budget leaves the method's output and hence performance unchanged.

\subsection{Results}
\label{sec:results}

\begin{table}[t]
\centering
\begin{minipage}[t]{0.54\linewidth}
\vspace{0pt}
\centering
\small
\setlength{\tabcolsep}{3pt}
\caption{\textbf{Zero-shot transfer.}
Test micro-accuracy (\%) on GSM8K and MATH-500, one seed.
Block-size transfer uses the same frozen
$1\!\times\!256$-trained policy.
$T$ is the decoder-call budget.}
\begin{tabular}{@{}lrrr@{}}
\toprule
Method / geometry & $T=16$ & $T=32$ & $T=64$ \\
\midrule
\multicolumn{4}{@{}l@{}}{
    \textit{Dream-7B: zero-shot transfer, $1\!\times\!256$.}
} \\
$\mathrm{TIPS}_{\mathrm{full}}$ & 35.18 & 46.56 & 52.39 \\
Fast-dLLM & 22.87 & 27.43 & 29.96 \\
\midrule
\multicolumn{4}{@{}l@{}}{
    \textit{LLaDA: block-size transfer.}
} \\
$1\!\times\!256$ (training geometry) & 47.00 & 59.10 & 62.67 \\
$2\!\times\!128$ & 36.78 & 55.25 & 62.89 \\
$4\!\times\!64$ & 23.80 & 46.95 & 59.81 \\
$8\!\times\!32$ & 22.43 & 34.03 & 55.25 \\
\bottomrule
\end{tabular}

\label{tab:supp-transfer}
\end{minipage}\hfill%
\begin{minipage}[t]{0.425\linewidth}
\vspace{0pt}
\centering
\small
\setlength{\tabcolsep}{2pt}
\caption{\textbf{Handoff from a heuristic sampler at $T=64$.}
Test micro-accuracy (\%), one seed, switching from non-adjusted confidence
thresholding to TIPS after budget fraction $q$.
Both variants outperform continued heuristic decoding ($q=1$)
at every tested intermediate $q$.}
\begin{tabular}{@{}lcc@{}}
\toprule
Fraction $q$ & $\mathrm{TIPS}_{\mathrm{full}}$
             & $\mathrm{TIPS}_{\mathrm{block}}$ \\
\midrule
$0$ (TIPS)       & 63.39 & 64.54 \\
$1/4$                & 40.85 & 38.70 \\
$1/2$                & 31.34 & 38.15 \\
$3/4$                & 25.73 & 40.19 \\
$1$ (Non-adj. Fast-dLLM) & 23.91 & 34.91 \\
\bottomrule
\end{tabular}

\label{tab:supp-handoff-main}
\end{minipage}
\end{table}
\begin{figure}[ht]
    \centering
    \begin{minipage}[t]{0.49\linewidth}
        \centering
        \includegraphics[width=\linewidth]{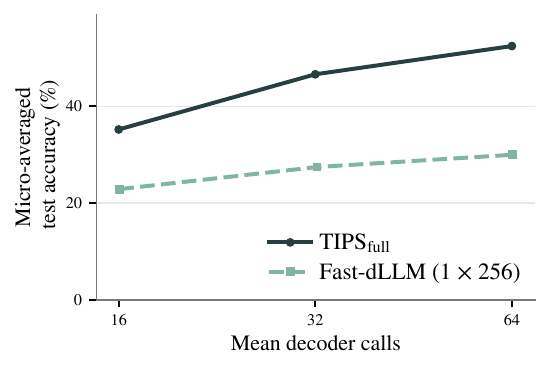}
        \par\small (a) Zero-shot DREAM transfer
    \end{minipage}\hfill
    \begin{minipage}[t]{0.49\linewidth}
        \centering
        \includegraphics[width=\linewidth]{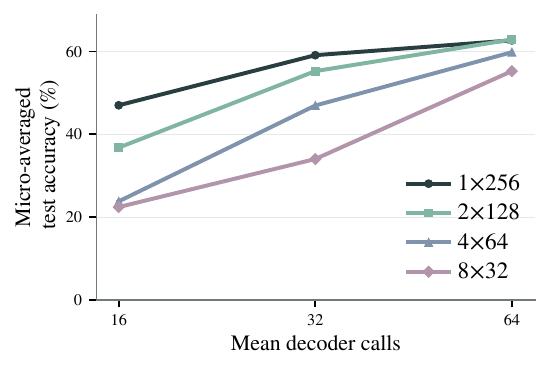}
        \par\small (b) Block-size transfer
    \end{minipage}
    \caption{\textbf{Generalization of ${\mathrm{TIPS}_{\mathrm{full}}}$.} Left: the LLaDA-trained full-canvas policy transfers to Dream-7B. Right: $\mathrm{TIPS}_{\mathrm{full}}$ is applied to different block sizes on LLaDA, with decoding budget shared across blocks.}
    \label{fig:supp-transfer}
\end{figure}

\paragraph{Results. }
\cref{fig:main-results,tab:main-results} presents our main results.
TIPS substantially improves the accuracy--compute tradeoff at low and
moderate decoding budgets, with $\mathrm{TIPS}_{\mathrm{full}}$
achieving the highest accuracy among all methods at
8--32 call caps. Notably, $\mathrm{TIPS}_{\mathrm{full}}$
at a budget of 16 calls already exceeds WINO's accuracy at a budget cap of 32 calls. 
The two TIPS variants exhibit complementary strengths across decoding budgets. While $\mathrm{TIPS}_{\mathrm{full}}$ performs better at 4--32 call caps, $\mathrm{TIPS}_{\mathrm{block}}$ achieves higher accuracy at 64--256 call caps and is often the highest accuracy among the methods in this range. At 256 call cap, $\mathrm{TIPS}_{\mathrm{block}}$ reaches 68.18\%, compared with 67.73\% for UNCODE and 67.01\% for Fast-dLLM: TIPS remains competitive at high budget.%

By looking at the reported AUC values and inspecting \cref{fig:main-results} visually, we see that $\mathrm{TIPS}_{\mathrm{full}}$ and
$\mathrm{TIPS}_{\mathrm{block}}$ achieves substantially higher AUC than the other displayed baselines, with Fast-dLLM being the next best at $40.83$. 
Overall, this demonstrates that both methods are effective at many budget levels without the need to re-fit, and show the value of learning a lightweight controller for the decoding trajectory.

\subsection{Supplementary Experiments}
\label{sec:supplementary-experiments}
We investigate whether TIPS can transfer beyond its decoder backbone during training, different decoding geometries, and whether it can effectively continue decoding from states generated by another sampler. 
These experiments use one seed and, whenever applicable, are evaluated with decoder NFE budgets $T\in\{16,32,64\}$. Unless explicitly mentioned, the mathematical reasoning results are micro-averaged over the test set of GSM8K and MATH-500 as before. 
Additional experiments and results appear in \cref{app:additional-experiments}.

\paragraph{Zero-shot transfer to Dream-7B. }
We apply $\mathrm{TIPS}_{\mathrm{full}}$ policy network to Dream-7B-Instruct without retraining and compare the performance to Fast-dLLM on a single 256-token canvas \citep{ye2025dream}. 
As seen in \cref{fig:supp-transfer} and \cref{tab:supp-transfer}, these gains show that non-trivial visibility decisions can transfer to a different decoder backbone without the need for re-training. 

\paragraph{Zero-shot transfer across block sizes. }
We reuse $\mathrm{TIPS}_{\mathrm{full}}$ for $2\times128$, $4\times64$, and $8\times32$ decoding, keeping the total response length and decoder budget fixed, in order to examine TIPS' ability to generalize to different block sizes which, in turn, also shed light on its performance on different response lengths. Also seen in \cref{fig:supp-transfer} and \cref{tab:supp-transfer}, $\mathrm{TIPS}_{\mathrm{full}}$ can preserve accuracy at slightly smaller block sizes before eventual degradation at $8\times32$. 

\paragraph{Handoff from a heuristic sampler. }
We evaluate whether TIPS can effectively continue decoding
from intermediate states generated by another sampler.
To do so, we run Fast-dLLM's confidence-thresholding sampler with default threshold of 0.9 for a fraction $q\in[0,1]$ of the total forward calls then switch to TIPS. 
Across both TIPS variants and various budgets, 16 out of the 18 settings improve over the baseline confidence-thresholding sampler, demonstrating a consistent ability to recover from states generated from other decoding methods. 
For $T=64$ specifically, as seen in \cref{tab:supp-handoff-main}, all settings $q$ are able to improve over continued heuristic decoding. 
However, available comparisons indicate that
discarding the inherited canvas and restarting TIPS with only
the remaining $(1-q)T$ calls can yield higher accuracy.
Thus, handoff improves on continued heuristic decoding,
but does not necessarily outperform restarting.
See \cref{app:heuristic-full} for details.

We further apply $\mathrm{TIPS}_{\mathrm{full}}$ after non-adjusted Fast-dLLM
has completed its response: four additional decoder evaluations
raise test micro-accuracy from $67.01\%$ to $67.29\%$, correcting
five initially incorrect answers without making any initially
correct answer incorrect in this run
(Appendix~\ref{app:completed-canvas-repair}).

\subsection{Ablations}
\label{sec:ablations}
We ablate TIPS on a held-out validation set of 512 GSM8K and
512 MATH training examples, reporting micro-averaged accuracy.
Its different dataset composition makes these accuracies not
directly comparable to test results. Ablations cover supervised warmup, remasking, visibility-aware
input features, the choice of and sensitivity to $\gamma$,
policy architecture and runtime, and warmup and RL decoding
budgets. Full experimental details and accuracy--compute curves
appear in Appendix~\ref{app:ablations}.

\section{Related Work}
\label{sec:related-work}
\paragraph{Training-free decoding order and adaptive parallelism. }
Masked diffusion allows many reveal orders, which can materially affect generation quality \citep{sahoo2024simple,kim2025ordering}. 
Samplers often prescribe how many positions to reveal at each pass, selecting their identities using decoder statistics. LLaDA uses confidence-ranked unmasking by default \citep{nie2025llada}. More structured approaches spread reveals across dilated position groups or compare short candidate decoding paths before committing tokens \citep{luxembourg2025plan,lee2025lookahead}. Adaptive-parallelism methods include Fast-dLLM, which combines confidence-thresholded decoding with approximate block-wise KV caching \citep{wu2025fastdllm}, and EB-Sampler, which dynamically reveals multiple positions subject to an entropy-based error tolerance \citep{benhamu2025entropy}. Other methods refine the selection criterion: UNCODE/PC-Sampler calibrates uncertainty scores using position- and token-informativeness priors \citep{huang2025uncode}; KLASS incorporates the temporal KL stability of token distributions \citep{kim2025klass}; and DAPD uses a self-attention-derived dependency graph to select weakly coupled positions for parallel decoding \citep{kim2026dapd}.

\paragraph{Learned unmasking schedules. }
A closely related line of work learns a controller around a pretrained denoiser. 
\citet{jazbec2026learning} formulate masked-diffusion sampling as a Markov decision process and use reinforcement learning to train a one-layer transformer that maps decoder confidences to monotonic unmasking decisions.
Their transitions preserve already visible tokens, whereas TIPS extends the action semantics to reversible next-step visibility.
Where-to-Unmask trains a supervised ranker to imitate a ground-truth probability-margin oracle \citep{asano2026where}. Adaptive Order Policies reweight masked-diffusion loss terms by policy probabilities to favor positions at which the denoiser is likely to be correct, considering both frozen-denoiser and joint-training settings \citep{mohamud2026adaptive}. Self-Aware Scheduling (SAS) derives a dense trajectory reward from a bound on sequential decoding mismatch and trains an order policy with group-relative policy optimization while keeping the denoiser fixed \citep{xu2026scheduling}. UPO similarly learns a stochastic position-selection policy for a frozen masked diffusion model, using a KL-regularized, GRPO-style objective around an explicit heuristic reference policy \citep{hong2026upo}. Its actions select only currently masked positions, preserving monotonicity. %

\paragraph{Re-masking and revision. }
Re-masking methods relax monotonicity by allowing earlier token commitments to be withdrawn. P2 and ReMDM allow visible positions to return to the mask state and be regenerated, with ReMDM deriving its sampler from a discrete diffusion process with a custom reverse transition \citep{peng2025pathplanning,wang2025remasking}. Running Confidence Remasking (RCR), introduced alongside MDPO, revisits positions with poor running confidence \citep{he2025mdpo}. WINO uses a training-free parallel draft-and-verify procedure to reveal candidates and re-mask suspicious visible tokens \citep{hong2025wino}, while CoRe identifies revision targets by testing prediction stability under masked-context perturbations \citep{zhai2026core}. These approaches prescribe revision rules rather than learning a separate visibility controller. RemeDi instead modifies the language model to jointly predict tokens and confidence scores, then trains its re-masking behavior through supervised and reinforcement-learning stages \citep{huang2025don}. TIPS also uses a supervised warmup phase followed by reinforcement learning.

\paragraph{Reinforcement learning for diffusion language models. }
Policy optimization has also been applied directly to the dLLM. MDPO updates the masked token model along progressive denoising trajectories using intermediate rewards \citep{he2025mdpo}; mask-aware policy gradients account for both token and masking decisions \citep{raajesh2026maskawarepolicygradientsdiffusion}; and SPG constructs upper- and lower-bound surrogates for the otherwise intractable sequence log-likelihood \citep{wang2025spg}. TIPS instead applies policy gradients only to its auxiliary Bernoulli visibility policy, hence training requires neither sequence likelihood nor updates to decoder parameters.

\section{Discussion}
\label{sec:discussion}

We introduced TIPS, a small policy that learns unmasking and
remasking around a frozen dLLM, separating visibility control
from token prediction while a fixed schedule regulates expected
progress.
On GSM8K and MATH-500, TIPS improves low- and moderate-budget
accuracy and remains competitive at higher budgets without
retraining; full-canvas and blockwise variants offer
complementary strengths.
Ablations, corruption recovery, heuristic-prefix handoffs,
remasking and editing statistics, and transfer experiments
highlight the value of warmup and remasking, suggest that
revision may help by withdrawing harmful context rather than
only replacing tokens, and reveal transferable but
domain-dependent behavior.

\paragraph{Limitations. }
Our evaluation centers on $256$-token mathematical reasoning, and the supplementary transfer experiments use a single seed; longer generations and broader task coverage remain important tests of generality.
TIPS also uses a fixed decoding horizon rather than an adaptive stopping rule, preserving further revision but foregoing savings from early termination.
Developing stopping criteria that account for possible future revisions, and evaluating the resulting accuracy--latency trade-off, are important directions for future work.
Decoder-call budgets also do not directly capture wall-clock cost.

\appendix
\crefalias{section}{appendix}
\crefalias{subsection}{appendix}
\crefalias{subsubsection}{appendix}

\section{Expanded methodology}
\label{app:expanded-methodology}

\subsection{Fitting the visibility schedule}
\label{app:mse-projection-gamma}

We select $\gamma$ by fitting the visibility trajectory of a separately trained, uncalibrated pilot policy. Another feasible approach is to optimize $\gamma$ via hyperparameter search on the validation set. 
We first train this pilot from scratch with RLVR and terminal correctness rewards, without supervised warmup or logit calibration.
We then freeze the pilot and collect $N$ trajectories on the held-out validation prompts, again without calibration.
Let $T_{\mathrm{fit}}$ be the rollout budget, including the deterministic final decoder call, and let $b^{(j)}_{t,i}$ be the pilot's logit at position $i$ and transition $t$ in trajectory $j$.
The empirical mean expected next-state visible fraction is
\begin{equation}
    \bar v_{t+1}
    =
    \frac{1}{N}
    \sum_{j=1}^{N}
    \left[
        \frac{1}{L}
        \sum_{i=1}^{L}
        \sigma\!\left(b^{(j)}_{t,i}\right)
    \right],
    \qquad t=0,\ldots,T_{\mathrm{fit}}-2.
    \label{eq:pilot-expected-visibility}
\end{equation}
Because each Bernoulli action specifies whether a position is visible in the next state, the inner average is that state's conditional expected visible fraction.
Using probabilities instead of sampled mask counts removes the
additional noise from the next Bernoulli sample, while retaining the states visited by the uncalibrated policy.

We fit the endpoint-anchored family $\kappa_\gamma(\tau)=\tau^{\exp(\gamma)}$ by solving the one-dimensional least-squares problem
\begin{equation}
    \widehat\gamma
    \in
    \underset{\gamma\in\mathbb{R}}{\arg\min}
    \;\frac{1}{T_{\mathrm{fit}}-1}
    \sum_{t=0}^{T_{\mathrm{fit}}-2}
    \left[
        \bar v_{t+1}
        -
        \left(
            \frac{t+1}{T_{\mathrm{fit}}}
        \right)^{\exp(\gamma)}
    \right]^2.
    \label{eq:gamma-mse-projection}
\end{equation}
This is a least-squares projection of the pilot's empirical schedule onto
a continuous, monotone family with
$\kappa_\gamma(0)=0$ and $\kappa_\gamma(1)=1$.
The policy observes the current time $t/T_{\mathrm{fit}}$, but the fit
uses the next-state time $(t+1)/T_{\mathrm{fit}}$.
We exclude deterministic finalization because its full visibility is
imposed by the decoding algorithm.
The fit uses neither reference answers nor correctness labels for the
validation prompts.
We then fix $\widehat\gamma$ during subsequent calibrated policy training
and evaluate the same normalized schedule at other inference budgets.
The pilot supplies the schedule shape; it is not the final calibrated
policy.

For blockwise decoding, we apply the same procedure to the active block,
using its length and local decoder-call budget in place of $L$ and
$T_{\mathrm{fit}}$.
We average the expected visible fractions over validation trajectories
and blocks at matching local times, exclude each block's deterministic
finalization, and fit one shared $\gamma$ for that decoding geometry.

\subsection{Fast-dLLM budget matching}
\label{app:fastdlm-budget-matching}
For a target budget $T$, we tune the confidence-decoding threshold using test-set rollouts until the average NFE lies in $[T,T+0.5]$. The one-sided tolerance gives the baseline up to $0.5$ additional decoder calls on average. We use this mean-adjusted setting with $8\times32$ decoding for the main results and $1\times256$ decoding for Dream transfer. Results without this adjustment are labeled Non-adjusted Fast-dLLM or Non-adjusted confidence thresholding, as appropriate, and use decoder call caps. In the main comparison, threshold adjustment covers $T=8,16,32,64,128$; at $T=256$, we reuse the non-adjusted results, whose runs terminate well below their caps, and impute the unavailable 4-call result from the 8-call result. The 4-call result is unavailable because during our sweeps, there was no confidence-threshold hyperparameter found such that the expected NFE was within the range of $[4,4.5]$.

It is important to note that unless otherwise specified, Fast-dLLM refers to the mean-adjusted version, while the non-adjusted version with standard confidence threshold of 0.9 is referred to as the non-adjusted Fast-dLLM. 
\section{Additional Experiments}
\label{app:additional-experiments}

\subsection{Main results with all baselines}
\label{app:main-results}

Table~\ref{tab:full-main-comparison} 
expands the main comparison to all evaluated baseline configurations. Fast-dLLM uses the mean-adjusted $8\times32$ setting described in \cref{sec:experimental-setup}. We also retain Non-adjusted Fast-dLLM with the validation-selected $1\times256$ configuration and the $8\times32$ shared-cap configuration; i.e.\ their thresholds are not adjusted to match mean NFE. 

Across the 256-token configurations, $\mathrm{TIPS}_{\mathrm{full}}$ has the highest mean accuracy at budgets 8--32, while $\mathrm{TIPS}_{\mathrm{block}}$ leads at 128 and 256. Fast-dLLM leads at the target budget of 64, reaching 65.59\% compared with 64.49\% for $\mathrm{TIPS}_{\mathrm{block}}$. The non-adjusted $8\times32$ configuration reaches 37.82\% at its 64-call cap. The $4\times32$ UPO configuration generates only 128 response tokens and is shown separately from the 256-token comparisons.

\begin{table}[htbp]
\centering
\caption{\textbf{Complete accuracy--budget comparison on mathematical reasoning.}
Micro-averaged test accuracy (\%) on GSM8K and MATH-500 (1,819 problems).
All baseline configurations are retained, with blocks denoting number of blocks $\times$ tokens per block.
TIPS entries give the mean and, underneath, its standard error across three training-seed means; each seed mean averages three inference draws.
Baselines are point estimates.
Bold indicates the highest reported mean in each budget column among the 256-token configurations.
Dashes indicate unreported cells.
The non-adjusted Fast-dLLM shared-cap adaptation is retained for reference, and the updated mean-budget adaptation is added as a separate row.
The final row uses only 128 response tokens and is displayed separately, without entering the matched comparison.
}
\label{tab:full-main-comparison}
\begingroup
\definecolor{fullcomparisondark}{HTML}{293E40}
\definecolor{fullcomparisonlight}{HTML}{80B6A1}
\footnotesize
\setlength{\tabcolsep}{2.5pt}
\renewcommand{\arraystretch}{1.12}
\begin{tabular*}{\linewidth}{@{\extracolsep{\fill}}lc*{7}{r}@{}}
\toprule
Method / configuration & Blocks & 4 & 8 & 16 & 32 & 64 & 128 & 256 \\
\midrule
\multicolumn{9}{l}{\textit{Single-canvas configurations: 256 response tokens}} \\
Random / uniform & $1\!\times\!256$ & 11.45 & 16.91 & 25.16 & 33.06 & 38.34 & 41.82 & 42.57 \\
High confidence (fitted) & $1\!\times\!256$ & 16.22 & 16.82 & 23.42 & 34.52 & 39.36 & 32.71 & 28.26 \\
High confidence (linear) & $1\!\times\!256$ & 5.06 & 12.26 & 30.84 & 37.82 & 33.32 & 27.32 & 29.19 \\
Probability margin & $1\!\times\!256$ & 9.84 & 17.59 & 31.61 & 39.53 & 35.51 & 27.54 & 30.40 \\
Predictive entropy & $1\!\times\!256$ & 4.45 & 7.81 & 20.51 & 32.49 & 25.73 & 19.79 & 28.59 \\
Non-adjusted Fast-dLLM & $1\!\times\!256$ & 17.70 & 22.37 & 31.67 & 33.32 & 33.48 & 33.48 & 33.48 \\
EB-Sampler & $1\!\times\!256$ & 12.04 & 16.88 & 19.85 & 21.72 & 25.40 & 27.21 & 31.39 \\
UNCODE / PC-Sampler & $1\!\times\!256$ & 12.26 & 14.35 & 21.50 & 29.74 & 53.66 & 63.66 & 67.73 \\
KLASS & $1\!\times\!256$ & 5.06 & 12.09 & 29.14 & 37.16 & 36.72 & 38.87 & 38.32 \\
DAPD & $1\!\times\!256$ & 8.36 & 14.79 & 21.37 & 28.95 & 50.47 & 50.82 & 50.82 \\
\shortstack[l]{\citet{jazbec2026learning}\\(expert steering, retrained)} & $1\!\times\!256$ & 5.63 & 7.42 & 9.18 & 12.39 & 47.72 & 64.80 & 64.63 \\
\shortstack[l]{\citet{jazbec2026learning}\\(vanilla, retrained)} & $1\!\times\!256$ & 5.81 & 9.05 & 16.40 & 38.34 & 48.30 & 47.32 & 47.90 \\
UPO & $1\!\times\!256$ & 5.94 & 7.15 & 8.03 & 10.23 & 12.37 & 17.59 & 49.70 \\
SAS & $1\!\times\!256$ & 9.84 & 20.95 & 33.53 & 41.56 & 43.65 & 43.65 & 43.76 \\
ReMDM & $1\!\times\!256$ & 0.55 & 4.93 & 21.73 & 32.40 & 41.14 & 40.52 & 33.53 \\
RCR & $1\!\times\!256$ & 8.34 & 19.72 & 33.33 & 38.67 & 35.79 & 33.88 & 36.93 \\
WINO & $1\!\times\!256$ & \textbf{18.14} & 21.06 & 34.80 & 46.34 & 49.26 & 49.37 & 49.37 \\
CoRe & $1\!\times\!256$ & 5.28 & 12.09 & 28.75 & 36.12 & 32.77 & 27.38 & 29.03 \\
\addlinespace[3pt]
\multicolumn{9}{l}{\textit{Block-wise configurations: 256 response tokens}} \\
Non-adjusted Fast-dLLM & $8\!\times\!32$ & 7.20 & 7.75 & 10.12 & 11.76 & 37.82 & 66.85 & 67.01 \\
Fast-dLLM & $8\!\times\!32$ & --- & 17.48 & 14.73 & 37.93 & \textbf{65.59} & 67.01 & 67.01 \\
KLASS (native block-wise) & $4\!\times\!64$ & 14.07 & 21.99 & 28.70 & 40.68 & 59.98 & 66.03 & 66.90 \\
UPO (block-trained; shared cap) & $8\!\times\!32$ & 7.26 & 7.92 & 9.46 & 11.05 & 11.16 & 22.54 & 66.85 \\
SAS (native block-wise) & $8\!\times\!32$ & --- & 17.48 & 23.75 & 33.26 & 53.22 & 61.46 & 66.14 \\
WINO (rescored block-wise) & $2\!\times\!128$ & 8.61 & 10.28 & 13.03 & 38.34 & 63.39 & 65.16 & 65.16 \\
\midrule
\rowcolor{fullcomparisondark!7}
\textcolor{fullcomparisondark}{\textbf{TIPS}$_{\mathrm{full}}$} & $1\!\times\!256$ & \shortstack[r]{16.46\\[-1.5pt]{\scriptsize $\pm$1.79}} & \shortstack[r]{\textbf{29.12}\\[-1.5pt]{\scriptsize $\pm$1.68}} & \shortstack[r]{\textbf{48.05}\\[-1.5pt]{\scriptsize $\pm$0.73}} & \shortstack[r]{\textbf{58.46}\\[-1.5pt]{\scriptsize $\pm$0.10}} & \shortstack[r]{63.09\\[-1.5pt]{\scriptsize $\pm$0.50}} & \shortstack[r]{65.32\\[-1.5pt]{\scriptsize $\pm$0.55}} & \shortstack[r]{66.28\\[-1.5pt]{\scriptsize $\pm$0.71}} \\
\rowcolor{fullcomparisonlight!19}
\textcolor{fullcomparisondark}{\textbf{TIPS}$_{\mathrm{block}}$} & $2\!\times\!128$ & \shortstack[r]{12.53\\[-1.5pt]{\scriptsize $\pm$0.45}} & \shortstack[r]{20.95\\[-1.5pt]{\scriptsize $\pm$0.99}} & \shortstack[r]{42.12\\[-1.5pt]{\scriptsize $\pm$2.04}} & \shortstack[r]{57.97\\[-1.5pt]{\scriptsize $\pm$1.23}} & \shortstack[r]{64.49\\[-1.5pt]{\scriptsize $\pm$0.80}} & \shortstack[r]{\textbf{67.12}\\[-1.5pt]{\scriptsize $\pm$0.28}} & \shortstack[r]{\textbf{68.18}\\[-1.5pt]{\scriptsize $\pm$0.36}} \\
\midrule
\multicolumn{9}{l}{\textit{Different response length: excluded from column-wise ranking}} \\
UPO (128-token response) & $4\!\times\!32$ & --- & --- & --- & --- & --- & 60.25 & --- \\
\bottomrule
\end{tabular*}
\endgroup
\end{table}

\subsection{Ablations}
\label{app:ablations}

\paragraph{Protocol. }
We evaluate the design choices of $\mathrm{TIPS}_{\mathrm{full}}$ on the 1,024-example validation set comprising 512 GSM8K and 512 MATH problems, chosen then excluded from the training data. The same prompts were also used without correctness labels to fit the native schedule exponent. Unless stated otherwise, each comparison uses training seed 123, evaluation seed 42, a 256-token response length, and inference budgets $T\in\{16,32,64\}$. Accuracy is the number of correct answers divided by 1,024. Because this equally weighted dataset mixture differs from the 1,319:500 test mixture, its absolute accuracies should not be compared directly with the main test results. 
The feature-width and finalization-deferral studies are historical exceptions, with their protocols specified separately below.

\subsubsection{Supervised warmup and reinforcement learning}
\label{app:exp-training-stages}
We compare the complete supervised warmup--RLVR recipe with its supervised parent and a policy trained with RLVR alone. The RLVR-trained policies use the same 935-update continuation length, while the warmup-only policy receives no RLVR updates. As shown in Table~\ref{tab:app-exp-training-remasking} and Figure~\ref{fig:app-exp-training-remasking}, both stages contribute to the final performance. Warmup followed by RLVR improves over warmup alone by 14.55, 17.87, and 14.45 percentage points at $T=16,32,64$, respectively. Warmup also provides a useful initialization for RLVR: although the gain over RLVR alone is only 0.49 points at $T=16$, it increases to 8.50 and 7.81 points at $T=32$ and $64$. Thus, the supervised policy is not simply sufficient on its own; supervised initialization and terminal-reward optimization complement one another.

\subsubsection{The contribution of remasking}
\label{app:exp-remasking-ablation}
To separate dependence on an already learned behavior from the value of the action space itself, we evaluate two controls. The first disables remasking at inference for the frozen reference policy. The second retrains a monotone policy with supervised warmup and RLVR. In both cases, visible tokens retain their identities, and only masked positions can become newly visible. If $V$ positions are already visible and $M=L-V$ remain masked, the reveal target is clipped to $[0,M]$ from $L\kappa_t-V$, so the intervention does not demand an infeasible decrease in visibility.

\begin{figure}[htbp]
\centering
\begin{minipage}[t]{0.49\linewidth}\centering
\includegraphics[width=\linewidth]{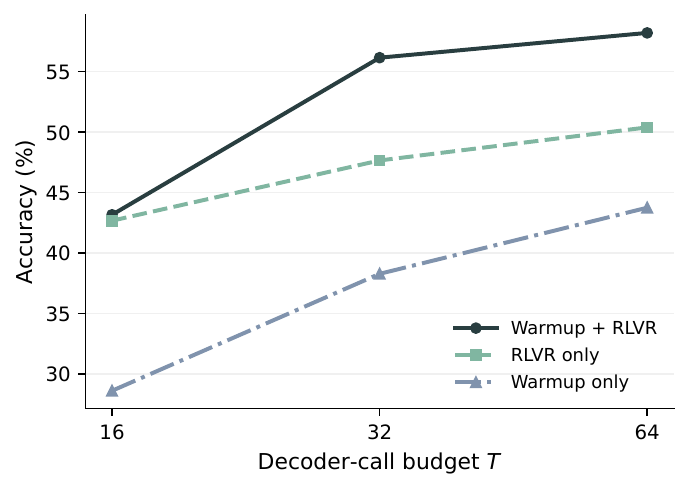}
\par\small (a) Training stages
\end{minipage}\hfill
\begin{minipage}[t]{0.49\linewidth}\centering
\includegraphics[width=\linewidth]{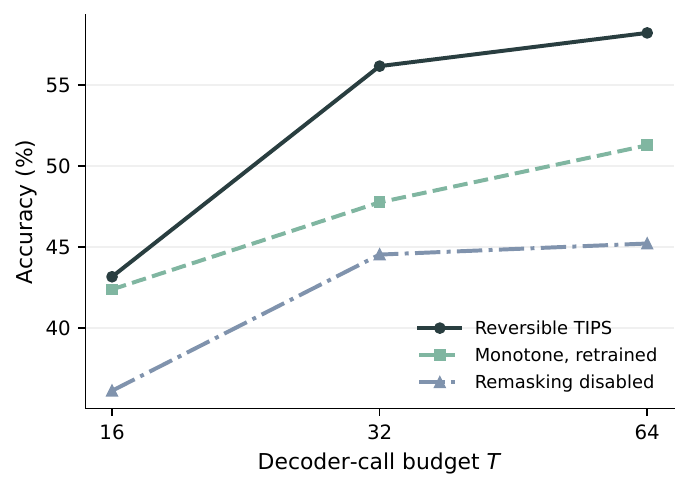}
\par\small (b) Remasking controls
\end{minipage}
\caption{\textbf{Supervised initialization and reversible visibility.} Validation-set accuracy versus decoder-call budget on a logarithmic axis. Each curve is one trained policy evaluated with one inference seed.}
\label{fig:app-exp-training-remasking}
\end{figure}

As seen in \cref{fig:app-exp-training-remasking}, disabling remasking reduces accuracy by 7.03--12.99 percentage points across the three budgets. Retraining recovers part of this loss, but the monotone policy remains 0.78, 8.40, and 6.93 points below reversible TIPS at $T=16,32,64$, respectively. These results support retaining the ability to reconsider visible context.

\begin{table}[htbp]
\centering
\small
\setlength{\tabcolsep}{4.5pt}
\caption{\textbf{Training stages and reversible visibility.} Validation-set micro-accuracy (\%) over 1,024 problems, with one training seed and one evaluation seed. RLVR continuations use the same number of updates.}
\begin{tabular}{lrrr}
\toprule
Configuration & $T=16$ & $T=32$ & $T=64$ \\
\midrule
Warmup $\rightarrow$ RLVR & 43.16 & 56.15 & 58.20 \\
Supervised warmup only & 28.61 & 38.28 & 43.75 \\
RLVR only & 42.68 & 47.66 & 50.39 \\
Remasking disabled (frozen) & 36.13 & 44.53 & 45.21 \\
Monotone policy (retrained) & 42.38 & 47.75 & 51.27 \\
\bottomrule
\end{tabular}

\label{tab:app-exp-training-remasking}
\end{table}
\FloatBarrier

\subsubsection{Visibility-consistent confidence features}
\label{app:exp-visibility-consistency}
At an already visible position, the policy controls whether the current token remains in context, rather than whether the decoder's newly preferred token should be inserted. A global-top-1 confidence can consequently describe the wrong token. 
When analyzing our trajectories, the global argmax differed from the visible token in 73,930 of 3,407,617 recorded visible-position events, or 2.17\%. This observation motivates anchoring the first feature to the current visible token and excluding that token when constructing the strongest alternative. While not necessarily an ablation, we believe that ensuring the policy inputs remain consistent with the current decoding state is important for the semantic correctness of our method. 

\subsubsection{Confidence-feature dimensionality study}
\label{app:exp-historical-features}
We additionally report an earlier study of feature width $K\in\{1,2,10,50\}$. This experiment uses a one-layer full-canvas policy trained with RLVR at $T_{\mathrm{train}}=32$. 
Inputs are raw full-vocabulary probabilities. The first coordinate measures support for the active token, using the greedy proposal at masked positions and the current token at visible positions; larger widths append the strongest alternative probabilities. The $K=1$ and $K=2$ variants therefore isolate whether to expose the strongest competitor.

\begin{figure}[htbp]
\centering
\includegraphics[width=0.92\linewidth]{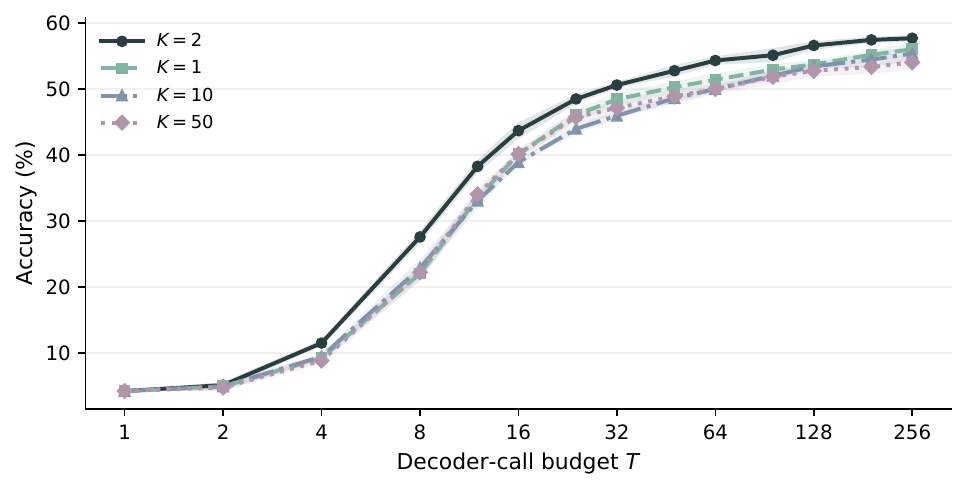}
\caption{\textbf{Confidence-feature width.} Test micro-accuracy versus log decoder-call budget for $K\in\{1,2,10,50\}$ raw-probability features. Markers include intermediate budgets as well as powers of two. Shading is one standard error across five evaluation seeds.}
\label{fig:app-exp-historical-features}
\end{figure}

Figure~\ref{fig:app-exp-historical-features} reports all 14 recorded inference budgets. For each fixed trained policy, we average five seeds. Adding the strongest alternative improves the curve broadly: $K=2$ exceeds $K=1$ at every budget above one call, including gains of 3.56 points at $T=16$, 2.14 points at $T=32$, and 2.92 points at $T=64$. Expanding the input to ten or fifty probabilities does not improve over $K=2$. In this setting, information about the most competitive alternative is more useful than a much wider description of the probability distribution, supporting the choice of a two-channel feature.

\begin{table}[htbp]
\centering
\footnotesize
\setlength{\tabcolsep}{4.5pt}
\caption{\textbf{Feature-width study.} Test micro-accuracy (\%), reported as mean $\pm$ sample standard deviation across five evaluation seeds for each fixed policy.}
\begin{tabular}{lrrrr}
\toprule
Feature width $K$ & $T=16$ & $T=32$ & $T=64$ & $T=256$ \\
\midrule
$1$ & 40.12 $\pm$ 0.93 & 48.44 $\pm$ 1.03 & 51.38 $\pm$ 1.00 & 56.01 $\pm$ 0.96 \\
$2$ & 43.68 $\pm$ 1.27 & 50.59 $\pm$ 0.48 & 54.30 $\pm$ 0.64 & 57.68 $\pm$ 0.48 \\
$10$ & 38.90 $\pm$ 0.35 & 45.94 $\pm$ 0.60 & 49.97 $\pm$ 0.53 & 55.35 $\pm$ 1.08 \\
$50$ & 40.13 $\pm$ 0.42 & 47.12 $\pm$ 1.12 & 49.97 $\pm$ 0.76 & 54.03 $\pm$ 1.10 \\
\bottomrule
\end{tabular}

\label{tab:app-exp-feature-width}
\end{table}
\FloatBarrier

\subsubsection{Occupancy control and finalization deferral}
\label{sec:ablation-schedule}

We revisit an earlier RLVR-only comparison of single-layer policies with one $256$-token response canvas, $T=32$ decoder calls, and three training seeds.
Policies are trained with no occupancy constraint, a soft schedule loss, or fixed power-law calibration, while retaining reversible actions and deterministic finalization.

The soft schedule loss imposes a penalty that is multiplied against the binary terminal correctness reward. Specifically, if, by the finalization pass, the state has masked fraction $m\in[0,1]$, then the assigned terminal correctness reward is multiplied by $(1-m)$. For example, if the response is correct but only $50\%$ of the canvas was unmasked by the finalization pass, the reward becomes $0.5$ for that trajectory. 

\begin{figure}[htbp]
    \centering
    \includegraphics[width=0.90\linewidth]{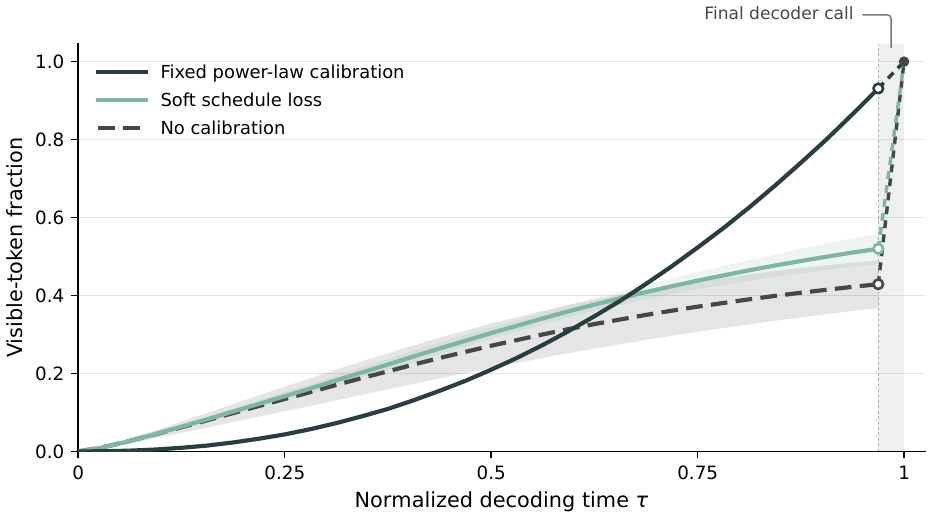}
    \caption{\textbf{Finalization deferral in RLVR-only policies.}
    Realized visible-token fraction at $T=32$: mean $\pm$ one standard deviation over training seeds $123$, $124$, and $125$, with evaluation seed $42$ and example-weighted pooling of GSM8K test and MATH-500.
    The shaded final interval and dashed terminal segments denote deterministic finalization.
    The origin is the prescribed all-mask state.}
    \label{fig:app-finalization-deferral}
\end{figure}

Figure~\ref{fig:app-finalization-deferral} shows that the uncalibrated trajectory levels off before the final call.
After the last learned transition, $57.11\%$ of positions remain masked, versus $48.01\%$ with the soft loss and $6.89\%$ with fixed calibration (Table~\ref{tab:app-finalization-deferral}).
The subsequent jump is imposed by deterministic completion, not learned visibility decisions: this is \emph{finalization deferral}.
We find the soft schedule loss, while non-constraining, insufficient for alleviating the finalization deferral phenomenon. 
Fixed calibration reduces this terminal workload and is accompanied by higher accuracy, from $45.90\%$ to $50.83\%$.

\begin{table}[htbp]
    \centering
    \small
    \setlength{\tabcolsep}{5pt}
    \caption{\textbf{Occupancy control reduces the finalization burden.}
    At $L=256$ and $T=32$, mask counts are means $\pm$ one standard deviation across three training seeds, measured at $\tau=31/32$ with evaluation seed $42$.}
    \begin{tabular}{@{}lrrr@{}}
        \toprule
        Occupancy control & Masks before finalization & Masked (\%) & Accuracy (\%) \\
        \midrule
        None & $146.20\pm15.68$ & $57.11$ & $45.90$ \\
        Soft schedule loss & $122.91\pm9.96$ & $48.01$ & $47.18$ \\
        Fixed power-law calibration & $17.65\pm0.02$ & $6.89$ & $50.83$ \\
        \bottomrule
    \end{tabular}

    \label{tab:app-finalization-deferral}
\end{table}
\FloatBarrier

\subsubsection{Schedule shape and logit calibration}
\label{app:exp-schedules}
\begin{figure}[htbp]
\centering
\begin{minipage}[t]{0.49\linewidth}\centering
\includegraphics[width=\linewidth]{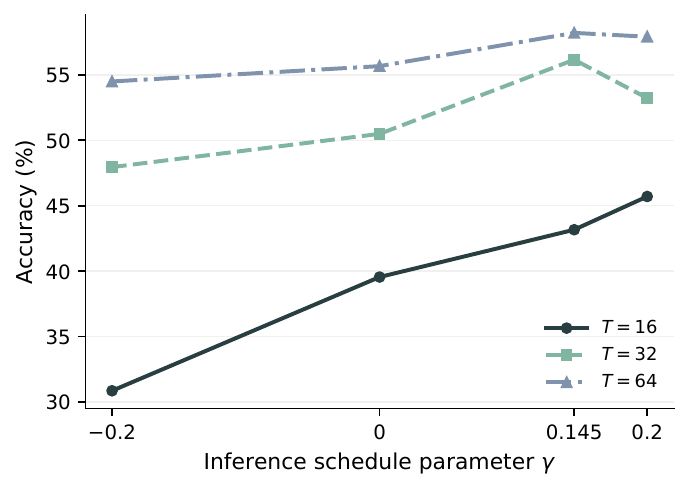}
\par\small (a) Schedule-shape sensitivity
\end{minipage}\hfill
\begin{minipage}[t]{0.49\linewidth}\centering
\includegraphics[width=\linewidth]{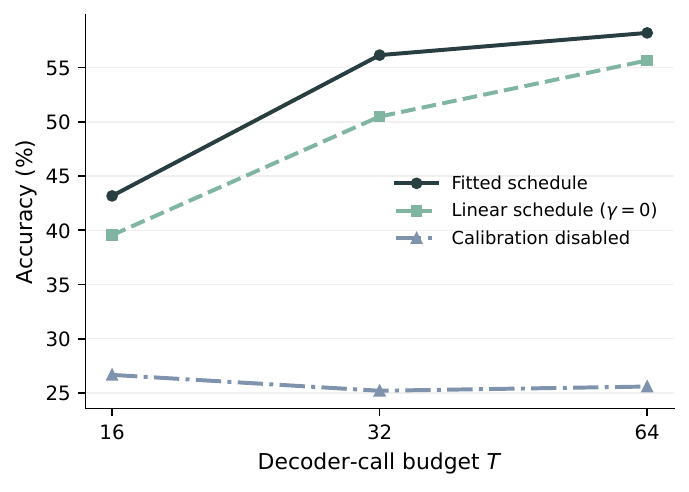}
\par\small (b) Calibration intervention
\end{minipage}
\caption{\textbf{Schedule shape and occupancy calibration have distinct effects.} Left: validation-set accuracy at the four measured $\gamma$ values, with a separate curve for each budget. Right: the fitted schedule, calibrated linear schedule, and calibration-disabled policy versus log budget. All settings use the same frozen checkpoint.}
\label{fig:app-exp-schedules}
\end{figure}

We freeze the reference policy and vary the inference schedule $\kappa(\tau)=\tau^{\exp(\gamma)}$. Table~\ref{tab:app-exp-schedules} and \cref{fig:app-exp-schedules} distinguishes changes in $\gamma$ from removing calibration altogether. The fitted value is $\gamma=0.14544$; the other evaluated settings are $-0.2$, $0$, and $0.2$. Note that $\gamma=0$ is a calibrated linear schedule, not an uncalibrated policy.

The effect of schedule shape depends on the inference budget. Moving from the fitted value to $\gamma=0.2$ raises accuracy at $T=16$ from 43.16\% to 45.70\%, but reduces it at $T=32$ from 56.15\% to 53.22\%; the $T=64$ difference is only 0.29 points. More front-loaded visibility at $\gamma=-0.2$ is worse at all three budgets, particularly at $T=16$. Therefore, nearby schedules can preserve much of the performance, but the schedule is not irrelevant and the best measured setting need not be the same at every budget.

\paragraph{Removing calibration. }
We next bypass the scalar occupancy-matching shift while keeping the checkpoint and deterministic finalization unchanged. Accuracy falls to 26.66\%, 25.20\%, and 25.59\% at $T=16,32,64$, respectively. In this intervention, additional decoder calls no longer yield the improvement observed with calibrated TIPS. The experiment establishes the importance of calibration for the learned policy's use of intermediate states.

\begin{table}[htbp]
\centering
\small
\setlength{\tabcolsep}{4.5pt}
\caption{\textbf{Inference-time schedule interventions.} Validation-set accuracy (\%) for frozen $\mathrm{TIPS}_{\mathrm{full}}$. All $\gamma$ settings retain occupancy calibration; the final row bypasses the scalar logit shift entirely.}
\begin{tabular}{lrrr}
\toprule
Frozen-policy setting & $T=16$ & $T=32$ & $T=64$ \\
\midrule
Fitted: $\gamma=0.14544$ & 43.16 & 56.15 & 58.20 \\
$\gamma=-0.2$ & 30.86 & 47.95 & 54.49 \\
$\gamma=0$ (linear) & 39.55 & 50.49 & 55.66 \\
$\gamma=0.2$ & 45.70 & 53.22 & 57.91 \\
Calibration disabled & 26.66 & 25.20 & 25.59 \\
\bottomrule
\end{tabular}

\label{tab:app-exp-schedules}
\end{table}
\FloatBarrier

\subsubsection{Policy architecture and runtime}
\label{app:exp-architecture}
We replace the two-layer DiT with either a one-layer DiT or a parameter-matched per-token multilayer perceptron, while retaining the training and evaluation protocol. The per-token network cannot exchange information between response positions within the controller. It nevertheless reaches 54.88\% accuracy at $T=64$, compared with 58.20\% for the two-layer DiT. Thus, a substantial fraction of the policy's effectiveness can be retained with local processing of contextual decoder features. Cross-position processing still helps: the two-layer DiT exceeds the MLP at all three budgets and exceeds the one-layer DiT at $T=32$ and $64$.

Table~\ref{tab:app-exp-architecture} also reports the corresponding controller costs. At $T=32$, cumulative policy time per generated example is 8.06\,ms for the two-layer DiT, 6.88\,ms for the one-layer DiT, and 6.22\,ms for the MLP. These measurements sum the controller's work across the complete rollout.

\begin{table}[htbp]
\centering
\footnotesize
\setlength{\tabcolsep}{4.5pt}
\caption{\textbf{Architecture and controller cost.} Validation-set accuracy (\%) and total CUDA-measured policy time per generated example at $T=32$. Timings accumulate all 31 policy decisions; they were collected on an H100 80GB with float32 policy weights and batch size at most 16. The decoder and feature-extraction cost are excluded from the policy-time column.}
\begin{tabular}{lrrrrr}
\toprule
Policy & Parameters & $T=16$ & $T=32$ & $T=64$ & Policy time (ms) \\
\midrule
Two-layer DiT & 628,739 & 43.16 & 56.15 & 58.20 & 8.06 \\
One-layer DiT & 331,395 & 45.12 & 53.03 & 55.86 & 6.88 \\
Per-token MLP & 634,835 & 38.18 & 48.24 & 54.88 & 6.22 \\
\bottomrule
\end{tabular}

\label{tab:app-exp-architecture}
\end{table}

\subsubsection{Training with a longer decoding budget}
\label{app:exp-training-horizon}
We compare $\mathrm{TIPS}_{\mathrm{full}}$ trained with $T_{\mathrm{train}}=32$ and $T_{\mathrm{train}}=128$ on the same validation problems. Unfortunately, computational constraints prevented a fully matched experiment, and the $T_{\mathrm{train}}=128$ configuration uses a different $\gamma$ of approximately $0.8$.

The $T_{\mathrm{train}}=32$ policy is more accurate at every measured inference budget, with gains of 4.00, 6.84, and 4.98 percentage points at $T=16,32,64$, respectively. A possible explanation is that a longer training trajectory requires more visibility decisions to share the same terminal correctness signal, making credit assignment more difficult.

\begin{table}[htbp]
\centering
\small
\setlength{\tabcolsep}{4.5pt}
\caption{\textbf{Training-budget comparison for $\mathrm{TIPS}_{\mathrm{full}}$.} Validation-set accuracy (\%) at the indicated inference budgets.}
\begin{tabular}{lrrr}
\toprule
Training configuration & $T=16$ & $T=32$ & $T=64$ \\
\midrule
$T_{\mathrm{train}}=32$ & 43.16 & 56.15 & 58.20 \\
$T_{\mathrm{train}}=128$ & 39.16 & 49.32 & 53.22 \\
\bottomrule
\end{tabular}

\label{tab:app-exp-horizon}
\end{table}
\begin{figure}[htbp]
\centering
\begin{minipage}[t]{0.49\linewidth}\centering
\includegraphics[width=\linewidth]{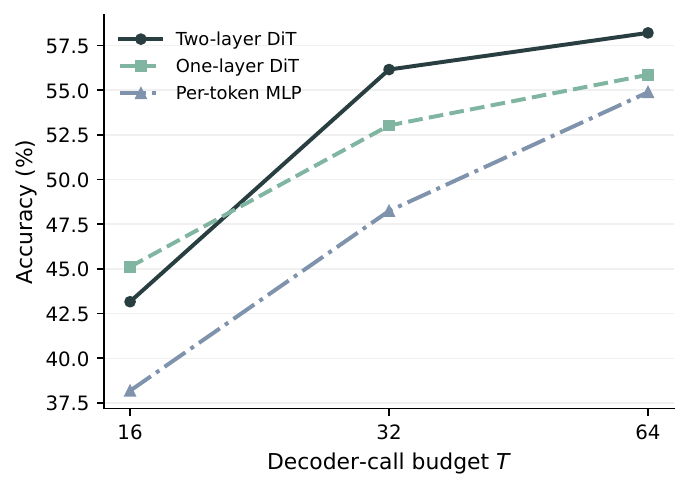}
\par\small (a) Policy architecture
\end{minipage}\hfill
\begin{minipage}[t]{0.49\linewidth}\centering
\includegraphics[width=\linewidth]{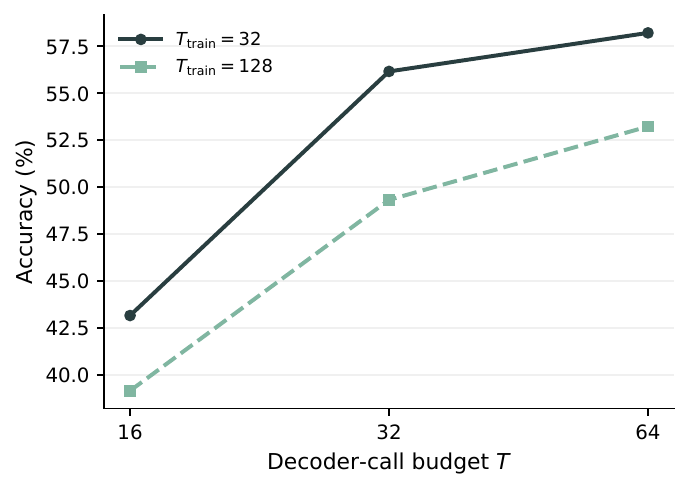}
\par\small (b) Training decoding budget
\end{minipage}
\caption{\textbf{Controller architecture and training budget.} Validation-set accuracy versus log decoder-call budget.}
\label{fig:app-exp-architecture-horizon}
\end{figure}
\FloatBarrier
\subsubsection{Depth-matched comparison with the learned unmasking policy}
\label{app:exp-depth-matching}
To examine whether the comparison with \citet{jazbec2026learning} is explained by controller depth, we evaluate one- and two-layer policies for both methods. We also vary the depth of $\mathrm{TIPS}_{\mathrm{block}}$ without changing its $2\times128$ decoding geometry. These depth-pair experiments use the same 1,024-example validation set and one training/evaluation seed pair. Within each method, the intended architectural change is the number of DiT layers; the training recipe is otherwise retained. Across methods, matching depth almost matches the parameter count: TIPS has 331,395 parameters at one layer and 628,739 at two, compared with 331,265 and 628,609 for the unmasking-only policy.

TIPS retains its advantage at either matched depth (Table~\ref{tab:app-exp-depth}). At $T=32$, the one-layer full-canvas TIPS policy reaches 53.03\%, compared with 43.85\% for the one-layer comparison policy; at two layers, the respective accuracies are 56.15\% and 40.33\%. Moreover, doubling the comparison policy's depth does not reproduce the improvement from TIPS. Its accuracy falls from 32.32\% to 20.70\% at $T=16$, and from 43.85\% to 40.33\% at $T=32$, while the two depths become similar at larger decoder budgets. These single-seed comparisons show that the reported advantage is not removed by matching controller depth, without attributing it to any one of TIPS's other design choices. 

For this depth comparison, we use the vanilla (non-expert-steered) variant of \citet{jazbec2026learning} at both depths, motivated by its stronger low-budget performance in our reproduction.
\cref{tab:main-results} instead reports the expert-steered variant, which achieves higher accuracy at large budgets. The validation split additionally
places greater weight on the harder MATH dataset (50\%, compared
with 27.48\% in the test mixture): it involves high-school competition-level mathematics requiring complex, multi-step symbolic reasoning, whereas GSM8K comprises elementary- and middle-school-level math word problems involving only basic arithmetic and simple multi-step algebraic manipulation. Thus, differences from \cref{tab:main-results} reflect changes in both the baseline variant and evaluation population. 

The compute interpretation also differs between methods. Both TIPS depths use their entire allocated horizon, whereas the comparison policy may terminate early. At cap 64, its deeper controller consumes 50.30 decoder calls on average, compared with 35.42 for the one-layer controller, despite their nearly identical accuracies. We therefore report mean calls alongside the cap-based accuracy curves.

\begin{table}[htbp]
\centering
\scriptsize
\setlength{\tabcolsep}{4.5pt}
\caption{\textbf{Matched policy-depth comparisons.} Validation-set accuracy (\%) at the requested decoder-call caps. Each TIPS decoding geometry is unchanged when depth is varied.}
\begin{tabular}{lrrrrrrrr}
\toprule
Method & Layers & Parameters & $4$ & $8$ & $16$ & $32$ & $64$ & $128$ \\
\midrule
$\mathrm{TIPS}_{\mathrm{full}}$ & 1 & 331,395 & 17.38 & 25.98 & 45.12 & 53.03 & 55.86 & 59.47 \\
$\mathrm{TIPS}_{\mathrm{full}}$ & 2 & 628,739 & 18.26 & 28.42 & 43.16 & 56.15 & 58.20 & 59.67 \\
$\mathrm{TIPS}_{\mathrm{block}}$ & 1 & 331,395 & 12.70 & 20.90 & 40.43 & 52.64 & 58.79 & 61.33 \\
$\mathrm{TIPS}_{\mathrm{block}}$ & 2 & 628,739 & 15.04 & 23.54 & 41.60 & 53.03 & 59.18 & 61.33 \\
\citet{jazbec2026learning} & 1 & 331,265 & 7.32 & 14.16 & 32.32 & 43.85 & 43.75 & 43.95 \\
\citet{jazbec2026learning} & 2 & 628,609 & 7.23 & 11.62 & 20.70 & 40.33 & 43.65 & 44.82 \\
\bottomrule
\end{tabular}

\label{tab:app-exp-depth}
\end{table}
\begin{table}[htbp]
\centering
\small
\setlength{\tabcolsep}{4.5pt}
\caption{\textbf{Consumed decoder calls in the depth comparison.} Mean calls per validation sample. TIPS uses every allocated call, whereas the comparison policy may stop before reaching the cap. Equal caps therefore do not always imply equal consumed compute.}
\begin{tabular}{lrrrr}
\toprule
Method & Cap $16$ & Cap $32$ & Cap $64$ & Cap $128$ \\
\midrule
TIPS (either geometry or depth) & 16.00 & 32.00 & 64.00 & 128.00 \\
Jazbec et al., one layer & 16.00 & 29.52 & 35.42 & 35.92 \\
Jazbec et al., two layers & 16.00 & 31.95 & 50.30 & 51.38 \\
\bottomrule
\end{tabular}

\label{tab:app-exp-depth-compute}
\end{table}
\begin{figure}[htbp]
\centering
\includegraphics[width=0.92\linewidth]{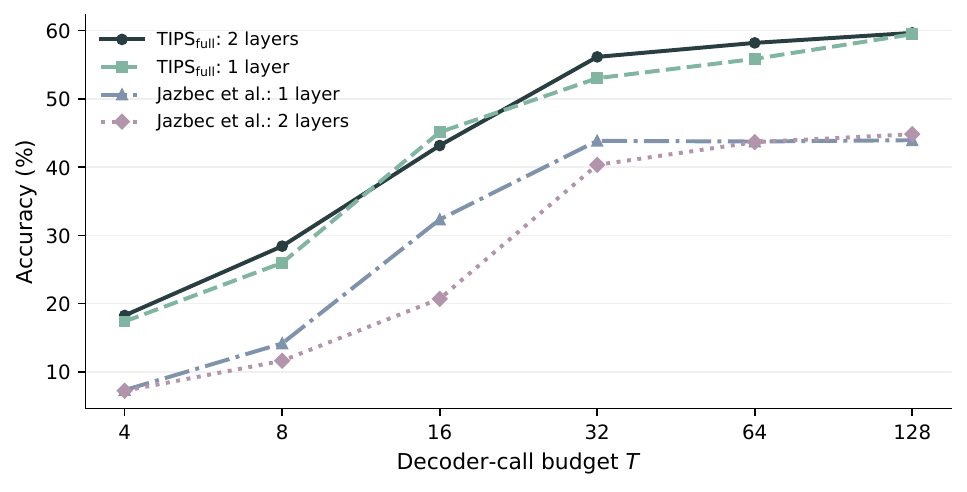}
\caption{\textbf{Full-canvas comparison at matched policy depth.} Validation-set accuracy for one- and two-layer TIPS and the learned unmasking policy of \citet{jazbec2026learning}. The horizontal axis is the requested decoder-call cap; consumed calls for adaptive policies are reported in Table~\ref{tab:app-exp-depth-compute}.}
\label{fig:app-exp-depth-full}
\end{figure}
\begin{figure}[htbp]
\centering
\includegraphics[width=0.64\linewidth]{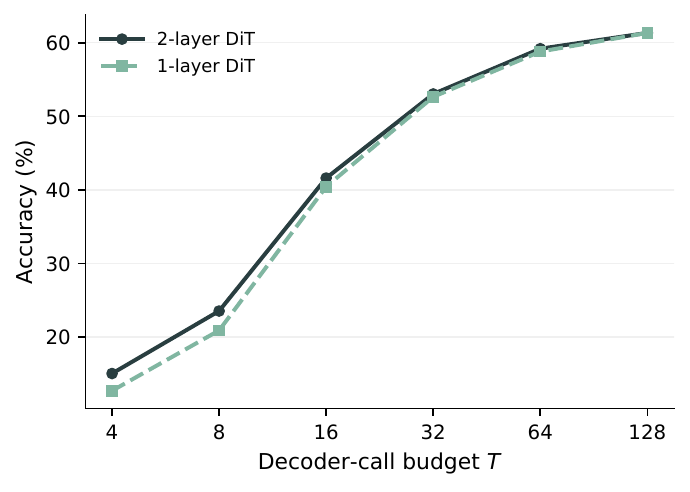}
\caption{\textbf{Policy depth under two-block decoding.} Both curves use the same $2\times128$ geometry and exact global decoder-call budget. The one-layer variant approaches the two-layer reference at larger budgets and matches it at $T=128$ in this single-seed comparison.}
\label{fig:app-exp-depth-block}
\end{figure}
\FloatBarrier
\subsection{Supplementary analyses of visibility and revision}
\label{app:exp-mechanistic}

\paragraph{Protocol. }
These experiments use $\mathrm{TIPS}_{\mathrm{full}}$ and $\mathrm{TIPS}_{\mathrm{block}}$ checkpoints with training seed 123, one evaluation seed, and the full GSM8K and MATH-500 test populations. Except for the explicitly selected corruption-reference pools, each reported mathematical-reasoning accuracy is micro-averaged over all 1,819 problems. The decoder remains frozen and uses greedy token proposals. We preserve each checkpoint's native schedule and timing convention, including the block policy's next-state local time input. All budgets count decoder calls, including deterministic finalization.

\subsubsection{Does TIPS select useful reveal and remask positions?}
\label{app:exp-position-selection}
We examine a low-budget case, $\mathrm{TIPS}_{\mathrm{block}}$ at $T=16$, by intervening on one decision along an otherwise on-distribution trajectory. At within-block phases $0.25$, $0.5$, and $0.75$ in each of the two blocks, we preserve the numbers of revealed and remasked positions but randomize the reveal positions, remask positions, or both. The random selections remain within the corresponding eligible masked or visible sets. Each continuation is evaluated on all test problems, including cases where an intervention leaves the action unchanged.

\begin{figure}[htbp]
\centering
\includegraphics[width=0.94\linewidth]{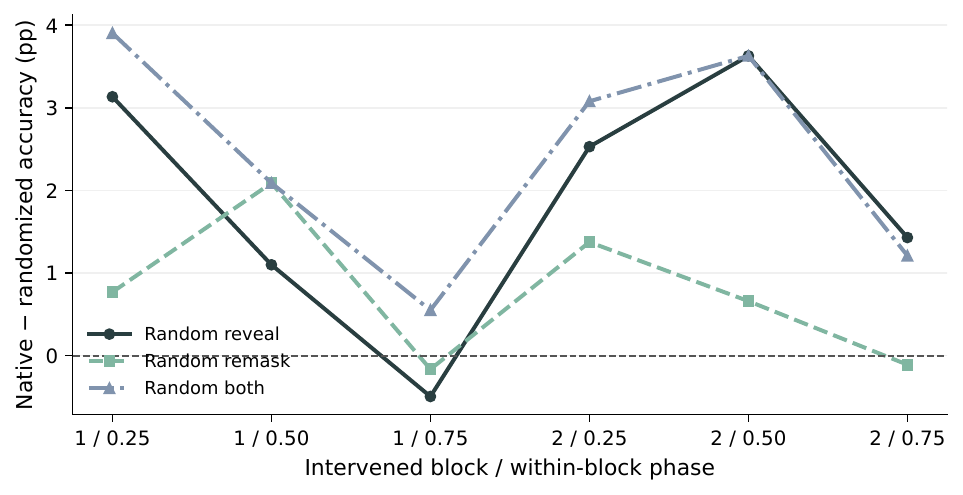}
\caption{\textbf{Count-preserving position randomization for $\mathrm{TIPS}_{\mathrm{block}}$ at $T=16$.} Positive values mean that native position selection yields higher final accuracy. All six intervention sites are shown, with one-based block indices. Reveal counts and remask counts are separately preserved at the intervened decision.}
\label{fig:app-exp-positions}
\end{figure}

As seen in \cref{fig:app-exp-positions} and \cref{tab:app-exp-positions}, randomizing both sets reduces accuracy at all six sites in this case study. For example, at phase $0.25$ of the first block, accuracy falls from 44.42\% to 40.52\%; at phase $0.5$ of the second block, it falls to 40.79\%. The separate interventions are less uniform: randomizing reveal or remask positions at phase $0.75$ of the first block slightly improves accuracy. Thus, the selected example provides evidence that the locations chosen by TIPS can matter beyond their counts, particularly early in the trajectory, while retaining cases where a particular native decision is not beneficial. It is a focused case study rather than a summary of all geometries and budgets.

\begin{table}[htbp]
\centering
\footnotesize
\setlength{\tabcolsep}{4.5pt}
\caption{\textbf{Position-selection case study: $\mathrm{TIPS}_{\mathrm{block}}$, $T=16$.} Test micro-accuracy (\%) for a count-preserving intervention at one decision. All six sites for this selected setting are included. Block indices are one-based.}
\begin{tabular}{rrrrrr}
\toprule
Block & Phase & Native & Random reveal & Random remask & Random both \\
\midrule
1 & 0.25 & 44.42 & 41.29 & 43.65 & 40.52 \\
1 & 0.50 & 44.42 & 43.32 & 42.33 & 42.33 \\
1 & 0.75 & 44.42 & 44.91 & 44.58 & 43.87 \\
2 & 0.25 & 44.42 & 41.89 & 43.05 & 41.34 \\
2 & 0.50 & 44.42 & 40.79 & 43.76 & 40.79 \\
2 & 0.75 & 44.42 & 42.99 & 44.53 & 43.21 \\
\bottomrule
\end{tabular}

\label{tab:app-exp-positions}
\end{table}
\FloatBarrier

\subsubsection{Recovery from controlled token corruption}
\label{app:exp-corruption-recovery}
We test whether a policy can recover from deliberately introduced token errors. For each policy, we first generate one $T=32$ response per test problem and retain verifier-accepted references. At within-block phase $0.75$, we construct matched clean and corrupted partial states, retaining only references with at least eight eligible visible tokens. We replace $k\in\{1,2,4,8\}$ visible tokens with uniformly drawn vocabulary tokens, excluding the original token, the mask, and tokenizer-defined special tokens. Corruption sets are nested across $k$, and every $k$ for a given policy and intervention block uses the same reference responses.

The intervention leaves eight calls for the full-canvas policy. For the two-block policy, corrupting the first block leaves four calls in that block and sixteen in the second, while corrupting the second block leaves four calls. Table~\ref{tab:app-exp-corruption-coverage} gives the resulting populations. 

\paragraph{Token restoration and answer correctness. }
TIPS frequently restores the exact substituted token, even though neither the policy nor the decoder receives the reference during continuation. 
As seen in \cref{tab:app-exp-corruption-recovery}, for first-block corruption, $\mathrm{TIPS}_{\mathrm{block}}$ restores 76.53\% of substituted tokens at $k=1$ and 66.55\% at $k=8$; the corresponding second-block rates are 66.35\% and 57.23\%. The full-canvas rates range from 42.50\% to 46.41\%. Exact restoration and answer correctness nevertheless measure different outcomes: an alternative token can preserve a correct solution, and restoring one token need not repair every consequence of the corruption. All corrupted conditions have lower answer accuracy than their matched clean continuations.

\paragraph{Restoration pathway and collateral drift. }
Most finally restored tokens last return through a learned reveal rather than deterministic finalization (Table~\ref{tab:app-exp-corruption-pathway}). For example, with eight corrupted tokens in the first block, 5,011 of the 5,830 finally restored tokens return through learned reveals, and 819 return at finalization. However, corruption can also affect positions that were not substituted. Table~\ref{tab:app-exp-corruption-drift} compares untouched-token disagreement with the reference against its paired clean-control rate. Excess disagreement is positive in all twelve corrupted conditions and generally grows with $k$. These measurements show both recovery and propagation of perturbations.

\begin{table}[htbp]
\centering
\footnotesize
\setlength{\tabcolsep}{4.5pt}
\caption{\textbf{Reference selection for controlled corruption.} One $T=32$ reference attempt is made per test problem for each policy. We retain verifier-accepted responses with at least eight eligible visible tokens at the intervention site. The same eligible pool is used for all $k$ within a row.}
\begin{tabular}{lrrr}
\toprule
Policy / active block & Accepted references & Eligible & Too few tokens \\
\midrule
$\mathrm{TIPS}_{\mathrm{full}}$ & 1047/1819 & 1047 & 0 \\
$\mathrm{TIPS}_{\mathrm{block}}$, block 1 & 1095/1819 & 1095 & 0 \\
$\mathrm{TIPS}_{\mathrm{block}}$, block 2 & 1095/1819 & 948 & 147 \\
\bottomrule
\end{tabular}

\label{tab:app-exp-corruption-coverage}
\end{table}
\begin{table}[htbp]
\centering
\scriptsize
\setlength{\tabcolsep}{4.5pt}
\caption{\textbf{Controlled corruption and recovery.} Answer correctness is measured over eligible problems; exact restoration is measured over the $k$ substituted tokens per problem. $k=0$ denotes the reconstructed clean continuation. $\Delta$ compares answer accuracy with that matched clean control.}
\begin{tabular}{lrrrrr}
\toprule
Policy / active block & $k$ & Correct / eligible & Answer (\%) & Restored (\%) & $\Delta$ (pp) \\
\midrule
$\mathrm{TIPS}_{\mathrm{full}}$ & 0 & 1038/1047 & 99.14 & --- & +0.00 \\
$\mathrm{TIPS}_{\mathrm{full}}$ & 1 & 1030/1047 & 98.38 & 42.50 & -0.76 \\
$\mathrm{TIPS}_{\mathrm{full}}$ & 2 & 1022/1047 & 97.61 & 42.50 & -1.53 \\
$\mathrm{TIPS}_{\mathrm{full}}$ & 4 & 1007/1047 & 96.18 & 43.79 & -2.96 \\
$\mathrm{TIPS}_{\mathrm{full}}$ & 8 & 979/1047 & 93.51 & 46.41 & -5.64 \\
\midrule
$\mathrm{TIPS}_{\mathrm{block}}$, block 1 & 0 & 1076/1095 & 98.26 & --- & +0.00 \\
$\mathrm{TIPS}_{\mathrm{block}}$, block 1 & 1 & 1041/1095 & 95.07 & 76.53 & -3.20 \\
$\mathrm{TIPS}_{\mathrm{block}}$, block 1 & 2 & 1045/1095 & 95.43 & 75.39 & -2.83 \\
$\mathrm{TIPS}_{\mathrm{block}}$, block 1 & 4 & 1015/1095 & 92.69 & 71.80 & -5.57 \\
$\mathrm{TIPS}_{\mathrm{block}}$, block 1 & 8 & 1001/1095 & 91.42 & 66.55 & -6.85 \\
\midrule
$\mathrm{TIPS}_{\mathrm{block}}$, block 2 & 0 & 948/948 & 100.00 & --- & +0.00 \\
$\mathrm{TIPS}_{\mathrm{block}}$, block 2 & 1 & 935/948 & 98.63 & 66.35 & -1.37 \\
$\mathrm{TIPS}_{\mathrm{block}}$, block 2 & 2 & 925/948 & 97.57 & 65.14 & -2.43 \\
$\mathrm{TIPS}_{\mathrm{block}}$, block 2 & 4 & 899/948 & 94.83 & 63.79 & -5.17 \\
$\mathrm{TIPS}_{\mathrm{block}}$, block 2 & 8 & 838/948 & 88.40 & 57.23 & -11.60 \\
\bottomrule
\end{tabular}

\label{tab:app-exp-corruption-recovery}
\end{table}
\begin{figure}[htbp]
\centering
\begin{minipage}[t]{0.49\linewidth}\centering
\includegraphics[width=\linewidth]{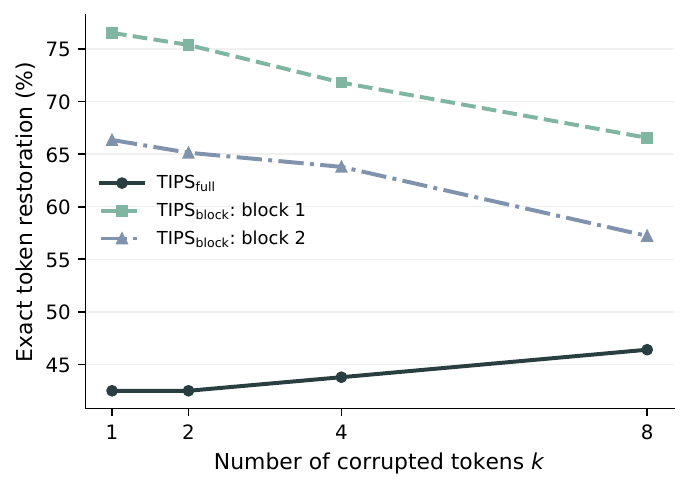}
\par\small (a) Exact substituted-token restoration
\end{minipage}\hfill
\begin{minipage}[t]{0.49\linewidth}\centering
\includegraphics[width=\linewidth]{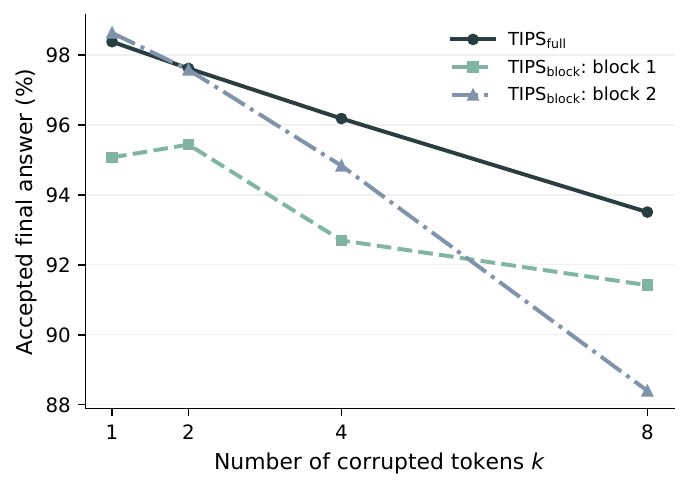}
\par\small (b) Final-answer correctness
\end{minipage}
\caption{\textbf{Recovery from controlled corruption.} The horizontal axis is the number of substituted tokens. Each curve uses its own accepted-reference population and remaining decoding horizon, as detailed in the text. Panel (a) weights corrupted tokens; panel (b) weights eligible problems.}
\label{fig:app-exp-corruption}
\end{figure}
\FloatBarrier
\begin{table}[htbp]
\centering
\scriptsize
\setlength{\tabcolsep}{4.5pt}
\caption{\textbf{Where restoration occurs.} The learned-reveal and finalization columns partition the finally restored tokens by their last restoration event. The last column counts tokens restored temporarily but not present in their reference form at termination.}
\begin{tabular}{lrrrrr}
\toprule
Policy / active block & $k$ & Restored & Learned reveal & Finalization & Transient only \\
\midrule
$\mathrm{TIPS}_{\mathrm{full}}$ & 1 & 445 & 434 & 11 & 0 \\
$\mathrm{TIPS}_{\mathrm{full}}$ & 2 & 890 & 857 & 33 & 0 \\
$\mathrm{TIPS}_{\mathrm{full}}$ & 4 & 1834 & 1767 & 67 & 0 \\
$\mathrm{TIPS}_{\mathrm{full}}$ & 8 & 3887 & 3747 & 140 & 4 \\
$\mathrm{TIPS}_{\mathrm{block}}$, block 1 & 1 & 838 & 736 & 102 & 0 \\
$\mathrm{TIPS}_{\mathrm{block}}$, block 1 & 2 & 1651 & 1442 & 209 & 2 \\
$\mathrm{TIPS}_{\mathrm{block}}$, block 1 & 4 & 3145 & 2724 & 421 & 3 \\
$\mathrm{TIPS}_{\mathrm{block}}$, block 1 & 8 & 5830 & 5011 & 819 & 7 \\
$\mathrm{TIPS}_{\mathrm{block}}$, block 2 & 1 & 629 & 536 & 93 & 0 \\
$\mathrm{TIPS}_{\mathrm{block}}$, block 2 & 2 & 1235 & 1067 & 168 & 0 \\
$\mathrm{TIPS}_{\mathrm{block}}$, block 2 & 4 & 2419 & 2094 & 325 & 0 \\
$\mathrm{TIPS}_{\mathrm{block}}$, block 2 & 8 & 4340 & 3647 & 693 & 1 \\
\bottomrule
\end{tabular}

\label{tab:app-exp-corruption-pathway}
\end{table}
\begin{table}[htbp]
\centering
\scriptsize
\setlength{\tabcolsep}{4.5pt}
\caption{\textbf{Drift at untouched positions.} We compare identical initially visible, nonspecial positions in each corrupted/clean pair, excluding substitutions. The first two rates and $\Delta_{\rm token}$ weight token positions; $\Delta_{\rm example}$ averages per-problem drift fractions. Differences are in percentage points. A token mismatch does not necessarily change the final answer.}
\begin{tabular}{lrrrrr}
\toprule
Policy / active block & $k$ & Corrupt (\%) & Clean (\%) & $\Delta_{\rm token}$ & $\Delta_{\rm example}$ \\
\midrule
$\mathrm{TIPS}_{\mathrm{full}}$ & 1 & 0.15 & 0.06 & +0.08 & +0.08 \\
$\mathrm{TIPS}_{\mathrm{full}}$ & 2 & 0.19 & 0.06 & +0.13 & +0.13 \\
$\mathrm{TIPS}_{\mathrm{full}}$ & 4 & 0.30 & 0.06 & +0.24 & +0.24 \\
$\mathrm{TIPS}_{\mathrm{full}}$ & 8 & 0.54 & 0.07 & +0.47 & +0.49 \\
$\mathrm{TIPS}_{\mathrm{block}}$, block 1 & 1 & 0.30 & 0.14 & +0.16 & +0.16 \\
$\mathrm{TIPS}_{\mathrm{block}}$, block 1 & 2 & 0.37 & 0.14 & +0.23 & +0.23 \\
$\mathrm{TIPS}_{\mathrm{block}}$, block 1 & 4 & 0.73 & 0.15 & +0.58 & +0.58 \\
$\mathrm{TIPS}_{\mathrm{block}}$, block 1 & 8 & 1.20 & 0.15 & +1.05 & +1.06 \\
$\mathrm{TIPS}_{\mathrm{block}}$, block 2 & 1 & 0.13 & 0.02 & +0.11 & +0.13 \\
$\mathrm{TIPS}_{\mathrm{block}}$, block 2 & 2 & 0.23 & 0.02 & +0.21 & +0.24 \\
$\mathrm{TIPS}_{\mathrm{block}}$, block 2 & 4 & 0.41 & 0.02 & +0.39 & +0.53 \\
$\mathrm{TIPS}_{\mathrm{block}}$, block 2 & 8 & 0.85 & 0.02 & +0.82 & +1.21 \\
\bottomrule
\end{tabular}

\label{tab:app-exp-corruption-drift}
\end{table}
\begin{figure}[htbp]
\centering
\includegraphics[width=0.66\linewidth]{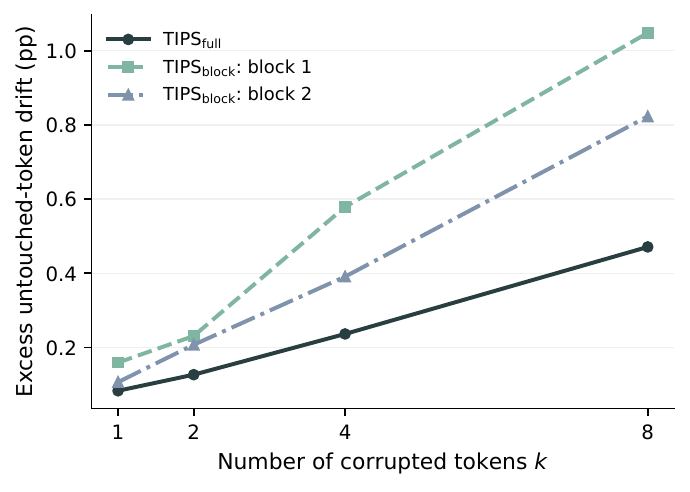}
\caption{\textbf{Excess drift at initially untouched positions.} Token-weighted disagreement with the reference after a corrupted continuation, minus disagreement at the identical untouched positions in its matched clean continuation. Block numbers are one-based.}
\label{fig:app-exp-corruption-drift}
\end{figure}
\FloatBarrier

\subsubsection{Remasking is more frequent than token editing}
\label{app:exp-editing-statistics}
We distinguish a remasking event from a completed token edit. A remasking episode starts when a visible token becomes masked and ends when that position next becomes visible; it is an edit only if the returning token differs from the original. Table~\ref{tab:app-exp-remask-frequency} reports remasking under four denominators, preventing trajectory-level incidence from being confused with a per-position rate. Every observed trajectory contains at least one remask, but the fraction of visible position-opportunities that are remasked is approximately 18\% for the full-canvas policy and 31--38\% for the block policy.

Most remasking episodes do not change token identity. Across the measured budgets, 91.52--95.17\% of full-canvas episodes and 92.91--98.16\% of blockwise episodes return the original token. Every recorded episode resolves by the end of decoding, and the mean delay before re-reveal is 2.25--3.13 calls for full-canvas decoding and 1.20--1.41 calls for blockwise decoding (Table~\ref{tab:app-exp-remask-edits}). Thus, remasking often temporarily withdraws context rather than replacing its content. Together with the remasking ablation, these observations motivate the hypothesis that hiding unreliable context can help other positions even when the hidden token ultimately returns unchanged. 
Another potential explanation is simply due to sampling noise, as each re-masking decision is sampled from a Bernoulli distribution informed by the policy network. 
The episode counts do not isolate that mechanism from the effects of actual token edits.

\begin{table}[htbp]
\centering
\scriptsize
\setlength{\tabcolsep}{4.5pt}
\caption{\textbf{How often TIPS remasks.} All 1,819 test problems are included in each row. The four rates use, respectively, visible position-opportunities, all active position-opportunities, learned steps containing a remask, and trajectories containing a remask. A position-opportunity is one active position at one learned decision. Deterministic finalization is excluded.}
\begin{tabular}{lrrrrrr}
\toprule
Policy & $T$ & Remasks & Visible (\%) & All positions (\%) & Steps (\%) & Trajectories (\%) \\
\midrule
$\mathrm{TIPS}_{\mathrm{full}}$ & 16 & 503,803 & 18.04 & 7.21 & 93.29 & 100.00 \\
$\mathrm{TIPS}_{\mathrm{full}}$ & 32 & 1,119,399 & 17.97 & 7.75 & 96.57 & 100.00 \\
$\mathrm{TIPS}_{\mathrm{full}}$ & 64 & 2,457,136 & 18.71 & 8.38 & 97.97 & 100.00 \\
$\mathrm{TIPS}_{\mathrm{block}}$ & 16 & 324,263 & 30.50 & 9.95 & 85.31 & 100.00 \\
$\mathrm{TIPS}_{\mathrm{block}}$ & 32 & 929,911 & 34.23 & 13.31 & 92.91 & 100.00 \\
$\mathrm{TIPS}_{\mathrm{block}}$ & 64 & 2,315,040 & 38.11 & 16.04 & 96.02 & 100.00 \\
\bottomrule
\end{tabular}

\label{tab:app-exp-remask-frequency}
\end{table}
\begin{table}[htbp]
\centering
\scriptsize
\setlength{\tabcolsep}{4.5pt}
\caption{\textbf{What happens after remasking.} An episode begins when a visible token is remasked and ends at its next reveal. Changed and same-token percentages use completed episodes; every recorded episode completes. Delay is the mean number of global forward-call intervals until re-reveal. Finalized counts episodes ending at deterministic finalization. Special is the fraction of remask events involving tokenizer-defined special tokens, without opportunity adjustment.}
\begin{tabular}{lrrrrrrr}
\toprule
Policy & $T$ & Changed & Changed (\%) & Same (\%) & Delay & Finalized & Special (\%) \\
\midrule
$\mathrm{TIPS}_{\mathrm{full}}$ & 16 & 42,714 & 8.48 & 91.52 & 2.245 & 29,553 & 45.36 \\
$\mathrm{TIPS}_{\mathrm{full}}$ & 32 & 72,426 & 6.47 & 93.53 & 2.699 & 16,413 & 36.39 \\
$\mathrm{TIPS}_{\mathrm{full}}$ & 64 & 118,743 & 4.83 & 95.17 & 3.126 & 8,332 & 31.96 \\
$\mathrm{TIPS}_{\mathrm{block}}$ & 16 & 22,988 & 7.09 & 92.91 & 1.198 & 48,771 & 25.16 \\
$\mathrm{TIPS}_{\mathrm{block}}$ & 32 & 31,264 & 3.36 & 96.64 & 1.294 & 31,638 & 18.69 \\
$\mathrm{TIPS}_{\mathrm{block}}$ & 64 & 42,633 & 1.84 & 98.16 & 1.412 & 17,001 & 14.76 \\
\bottomrule
\end{tabular}

\label{tab:app-exp-remask-edits}
\end{table}
\FloatBarrier

\subsubsection{Continuing heuristic trajectories with TIPS}
\label{app:heuristic-full}
We extend the main-text handoff experiment to all budgets and switching fractions. A non-adjusted confidence-threshold sampler based on Fast-dLLM first runs for $qT$ global decoder calls, with $q\in\{0,1/4,1/2,3/4,1\}$, after which TIPS uses the remaining budget. The heuristic reveals eligible masked proposals with confidence strictly above 0.9 and, when none qualifies, reveals the highest-confidence eligible proposal for each example. TIPS resumes at the corresponding point of its native schedule for normalized time input and logit calibration purposes.

For a matched budget comparison, all arms use exactly $T$ real forward passes, including calls that leave the canvas unchanged and deterministic finalizations. In particular, the $q=1$ endpoint is this fixed-horizon heuristic. We retain both endpoints and all three hybrid settings in Table~\ref{tab:app-exp-handoff}.

Sixteen of the eighteen hybrid conditions improve over the matched heuristic endpoint. At $T=64$, full-canvas handoff at $q=1/4$ raises accuracy from 23.91\% to 40.85\%, while blockwise handoff at $q=3/4$ raises it from 34.91\% to 40.19\%. The paired counts show that these gains can include both repaired and newly incorrect final answers: the former full-canvas condition changes 491 heuristic failures into correct answers while changing 183 heuristic successes into failures. No hybrid exceeds pure TIPS at the same budget. The result therefore supports TIPS's ability to improve continuations from another sampler, rather than recommending a heuristic prefix over starting with TIPS.

The results also support a comparison
at the remaining budget. For $T=64$ and $q=1/2$, handoff
leaves 32 calls for TIPS and yields $31.34\%$ and $38.15\%$
accuracy for $\mathrm{TIPS}_{\mathrm{full}}$ and
$\mathrm{TIPS}_{\mathrm{block}}$, respectively.
Their from-scratch 32-call runs instead achieve $57.56\%$
and $60.20\%$, suggesting that restarting from an all-mask
canvas can be preferable to retaining the heuristic state.

\begin{table}[htbp]
\centering
\small
\setlength{\tabcolsep}{4.5pt}
\caption{\textbf{Complete heuristic-to-TIPS handoff results.} Test micro-accuracy (\%) at all global switching fractions. $q=0$ is pure TIPS; $q=1$ is the matched fixed-horizon non-adjusted confidence-threshold heuristic. Every row uses exactly $T$ real decoder calls, including no-op and finalization calls.}
\begin{tabular}{lrrrrrr}
\toprule
Policy & $T$ & $q=0$ & $q=1/4$ & $q=1/2$ & $q=3/4$ & $q=1$ \\
\midrule
$\mathrm{TIPS}_{\mathrm{full}}$ & 16 & 46.23 & 24.02 & 16.99 & 17.26 & 17.70 \\
$\mathrm{TIPS}_{\mathrm{full}}$ & 32 & 57.56 & 31.45 & 20.56 & 19.79 & 19.74 \\
$\mathrm{TIPS}_{\mathrm{full}}$ & 64 & 63.39 & 40.85 & 31.34 & 25.73 & 23.91 \\
$\mathrm{TIPS}_{\mathrm{block}}$ & 16 & 44.42 & 17.98 & 20.45 & 20.62 & 15.94 \\
$\mathrm{TIPS}_{\mathrm{block}}$ & 32 & 60.20 & 24.85 & 22.76 & 26.06 & 20.67 \\
$\mathrm{TIPS}_{\mathrm{block}}$ & 64 & 64.54 & 38.70 & 38.15 & 40.19 & 34.91 \\
\bottomrule
\end{tabular}

\label{tab:app-exp-handoff}
\end{table}
\begin{figure}[htbp]
\centering
\begin{minipage}[t]{0.49\linewidth}\centering
\includegraphics[width=\linewidth]{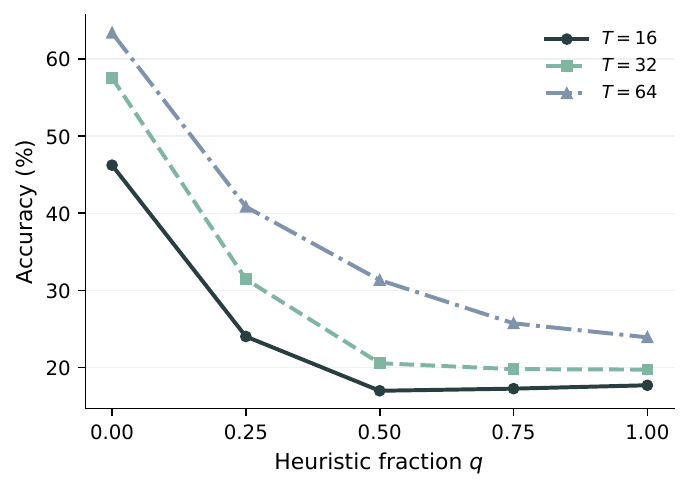}
\par\small (a) $\mathrm{TIPS}_{\mathrm{full}}$
\end{minipage}\hfill
\begin{minipage}[t]{0.49\linewidth}\centering
\includegraphics[width=\linewidth]{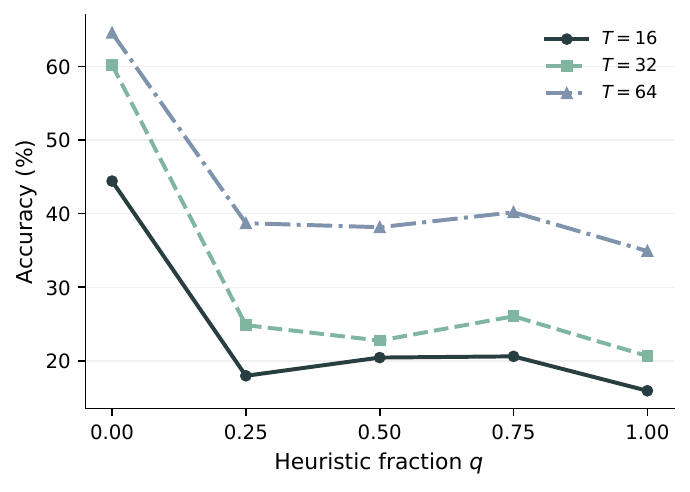}
\par\small (b) $\mathrm{TIPS}_{\mathrm{block}}$
\end{minipage}
\caption{\textbf{Heuristic prefixes followed by TIPS.} Test micro-accuracy at every evaluated switching fraction, including pure TIPS ($q=0$) and the matched heuristic ($q=1$). Each curve has a fixed total decoder-call budget.}
\label{fig:app-exp-handoff}
\end{figure}
\begin{table}[htbp]
\centering
\footnotesize
\setlength{\tabcolsep}{4.5pt}
\caption{\textbf{Paired final-answer changes after handoff.} Counts compare each hybrid with the matched pure-heuristic endpoint on the same 1,819 problems. Correctness refers to terminal answers. Both beneficial and adverse transitions are retained.}
\begin{tabular}{lrrrrr}
\toprule
Policy & $T$ & $q$ & Wrong $\to$ correct & Correct $\to$ wrong & $\Delta$ (pp) \\
\midrule
$\mathrm{TIPS}_{\mathrm{full}}$ & 16 & 0.25 & 267 & 152 & +6.32 \\
$\mathrm{TIPS}_{\mathrm{full}}$ & 16 & 0.50 & 91 & 104 & -0.71 \\
$\mathrm{TIPS}_{\mathrm{full}}$ & 16 & 0.75 & 42 & 50 & -0.44 \\
$\mathrm{TIPS}_{\mathrm{full}}$ & 32 & 0.25 & 394 & 181 & +11.71 \\
$\mathrm{TIPS}_{\mathrm{full}}$ & 32 & 0.50 & 137 & 122 & +0.82 \\
$\mathrm{TIPS}_{\mathrm{full}}$ & 32 & 0.75 & 50 & 49 & +0.05 \\
$\mathrm{TIPS}_{\mathrm{full}}$ & 64 & 0.25 & 491 & 183 & +16.93 \\
$\mathrm{TIPS}_{\mathrm{full}}$ & 64 & 0.50 & 284 & 149 & +7.42 \\
$\mathrm{TIPS}_{\mathrm{full}}$ & 64 & 0.75 & 88 & 55 & +1.81 \\
$\mathrm{TIPS}_{\mathrm{block}}$ & 16 & 0.25 & 186 & 149 & +2.03 \\
$\mathrm{TIPS}_{\mathrm{block}}$ & 16 & 0.50 & 195 & 113 & +4.51 \\
$\mathrm{TIPS}_{\mathrm{block}}$ & 16 & 0.75 & 143 & 58 & +4.67 \\
$\mathrm{TIPS}_{\mathrm{block}}$ & 32 & 0.25 & 246 & 170 & +4.18 \\
$\mathrm{TIPS}_{\mathrm{block}}$ & 32 & 0.50 & 184 & 146 & +2.09 \\
$\mathrm{TIPS}_{\mathrm{block}}$ & 32 & 0.75 & 163 & 65 & +5.39 \\
$\mathrm{TIPS}_{\mathrm{block}}$ & 64 & 0.25 & 316 & 247 & +3.79 \\
$\mathrm{TIPS}_{\mathrm{block}}$ & 64 & 0.50 & 205 & 146 & +3.24 \\
$\mathrm{TIPS}_{\mathrm{block}}$ & 64 & 0.75 & 167 & 71 & +5.28 \\
\bottomrule
\end{tabular}

\label{tab:app-exp-handoff-pairs}
\end{table}
\FloatBarrier

\providecolor{tipsRepairDark}{HTML}{293E40}
\providecolor{tipsRepairLight}{HTML}{80B6A1}

\subsubsection{Repairing completed non-adjusted Fast-dLLM responses}
\label{app:completed-canvas-repair}

\begin{figure}[htbp]
    \centering
    \includegraphics[width=0.94\linewidth]
        {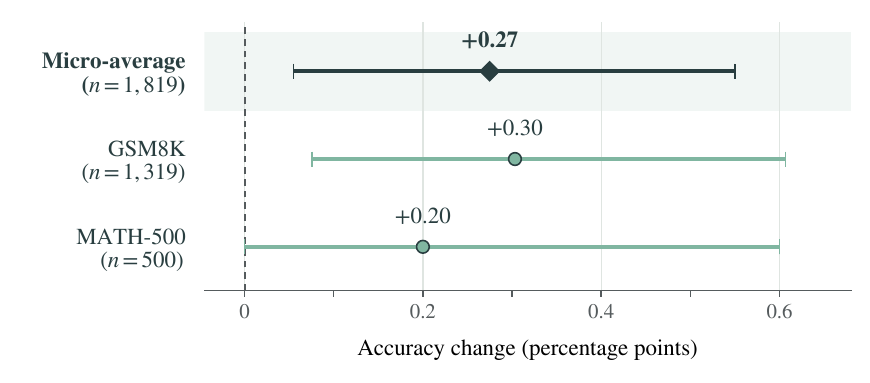}
    \caption{\textbf{Accuracy changes after four additional repair calls.}
    Points show the paired accuracy change from completed non-adjusted Fast-dLLM
    responses to their $\mathrm{TIPS}_{\mathrm{full}}$ revisions.
    Whiskers are $95\%$ paired bootstrap intervals from $10{,}000$
    resamples, sampling problems within each benchmark at its
    original size. The dashed line marks no accuracy change.}
    \label{fig:completed-canvas-repair}
\end{figure}

We next test whether TIPS can improve responses that another sampler
has already completed, rather than continue a partially decoded
trajectory. We first run non-adjusted Fast-dLLM on frozen LLaDA-8B-Instruct with
$8\times32$ decoding, confidence threshold $0.9$, no caching, and a
global cap of $256$ decoder calls. We then apply the frozen
$\mathrm{TIPS}_{\mathrm{full}}$ policy to every completed $256$-token
canvas on the full GSM8K and MATH-500 test sets, using exactly four
additional decoder evaluations. Repair retains the native schedule
and follows the final four calls of a virtual $32$-call horizon:
three policy transitions receive normalized times $28/32$, $29/32$,
and $30/32$, with next-state visibility targets
$((t+1)/32)^{\exp(\gamma_{\mathrm{full}})}$ for $t\in\{28,29,30\}$,
followed by deterministic finalization. The decoder and policy
remain frozen, and correctness feedback is used only for scoring. As shown in
Table~\ref{tab:completed-canvas-repair} and
Figure~\ref{fig:completed-canvas-repair}, micro-accuracy increases
from $67.01\%$ to $67.29\%$, a gain of $0.27$ percentage points.
TIPS corrects five of the $600$ initially incorrect answers---four
on GSM8K and one on MATH-500---while all $1{,}219$ initially correct
answers remain correct. This single-seed result shows a modest improvement from
revisiting completed responses, extending revision beyond unfinished
trajectories.

\begin{table}[htbp]
    \centering
    \small
    \setlength{\tabcolsep}{5.5pt}
    \renewcommand{\arraystretch}{1.08}
    \caption{\textbf{Four-call repair of completed non-adjusted Fast-dLLM responses.}
    Test accuracy before and after applying
    $\mathrm{TIPS}_{\mathrm{full}}$ to the same saved responses,
    with training and evaluation seed $123$. 
    W$\to$C and C$\to$W count wrong-to-correct and correct-to-wrong
    final-answer transitions, respectively. Every response is
    repaired, regardless of its original correctness. Non-adjusted Fast-dLLM
    has a $256$-call global cap but terminates naturally on every
    problem. Differences are computed from unrounded
    accuracies.}
    \begin{tabular}{lrrrrrr}
        \toprule
        & & \multicolumn{2}{c}{Accuracy (\%)}
        & & \multicolumn{2}{c}{Answer transitions} \\
        \cmidrule(lr){3-4}\cmidrule(lr){6-7}
        Dataset & $N$ & Non-adjusted Fast-dLLM & $+\mathrm{TIPS}_{\mathrm{full}}$
        & $\Delta$ (pp) & W$\to$C & C$\to$W \\
        \midrule
        GSM8K & 1,319 & 79.30 & 79.61 & +0.30 & 4 & 0 \\
        MATH-500 & 500 & 34.60 & 34.80 & +0.20 & 1 & 0 \\
        \midrule
        \rowcolor{tipsRepairLight!15}
        Micro-average & 1,819 & 67.01 & \textbf{67.29}
        & \textbf{+0.27} & 5 & 0 \\
        \bottomrule
    \end{tabular}

    \label{tab:completed-canvas-repair}
\end{table}

\subsection{Generalization beyond the training configuration}
\label{app:exp-generalization}

\paragraph{Protocol. }
We evaluate transfer with one training lineage and one evaluation seed, a 256-token response, and budgets $T\in\{16,32,64\}$. Mathematical-reasoning results use the complete GSM8K and MATH-500 test sets. Coding results use all 164 HumanEval and 500 MBPP test tasks and report pass@1 from one generation per task. Policy weights remain frozen for the zero-shot experiments; coding-trained policies are evaluated separately. Baseline settings are fixed independently of coding test outcomes. We retain a reduced baseline roster -- selected based on achieving the best non-TIPS results in mathematical reasoning in \cref{tab:main-results} -- due to computational constraints.

\subsubsection{Zero-shot transfer to Dream-7B}
\label{app:exp-dream-transfer}
We apply the LLaDA-trained $\mathrm{TIPS}_{\mathrm{full}}$ policy to Dream-v0-Instruct-7B without retraining. This comparison uses a single $1\times256$ canvas for both TIPS and Fast-dLLM and preserves the Dream logit-to-position alignment. As the policy observes confidence features rather than token identities or decoder hidden states, its input interface does not require a shared vocabulary or hidden-state dimension.

The transferred policy reaches 35.18\%, 46.56\%, and 52.39\% test micro-accuracy at $T=16,32,64$, compared with 22.87\%, 27.43\%, and 29.96\% for the matched full-canvas Fast-dLLM baseline. Table~\ref{tab:app-exp-dream} expands the main-text comparison with per-dataset scores and actual mean decoder calls. These results demonstrate transfer to another frozen backbone in this setting; they do not require adapting the policy on Dream outputs or selecting its parameters from Dream test scores.

\begin{table}[htbp]
\centering
\footnotesize
\setlength{\tabcolsep}{5pt}
\caption{\textbf{Dream transfer by dataset and consumed compute.} Columns correspond to requested budgets 16, 32, and 64. Both methods decode a single 256-token canvas. The pooled row weights individual problems equally.}
\begin{tabular}{llrrrrrr}
\toprule
Dataset & Method & \multicolumn{3}{c}{Accuracy (\%)} & \multicolumn{3}{c}{Mean decoder calls} \\
\cmidrule(lr){3-5}\cmidrule(lr){6-8}
 &  & $16$ & $32$ & $64$ & $16$ & $32$ & $64$ \\
\midrule
GSM8K & $\mathrm{TIPS}_{\mathrm{full}}$ & 41.77 & 55.72 & 63.84 & 16.00 & 32.00 & 64.00 \\
GSM8K & Fast-dLLM & 25.32 & 30.78 & 34.57 & 15.48 & 32.75 & 66.33 \\
MATH-500 & $\mathrm{TIPS}_{\mathrm{full}}$ & 17.80 & 22.40 & 22.20 & 16.00 & 32.00 & 64.00 \\
MATH-500 & Fast-dLLM & 16.40 & 18.60 & 17.80 & 17.50 & 30.70 & 58.21 \\
Micro-average & $\mathrm{TIPS}_{\mathrm{full}}$ & 35.18 & 46.56 & 52.39 & 16.00 & 32.00 & 64.00 \\
Micro-average & Fast-dLLM & 22.87 & 27.43 & 29.96 & 16.03 & 32.19 & 64.10 \\
\bottomrule
\end{tabular}

\label{tab:app-exp-dream}
\end{table}
\begin{figure}[htbp]
\centering
\includegraphics[width=0.65\linewidth]{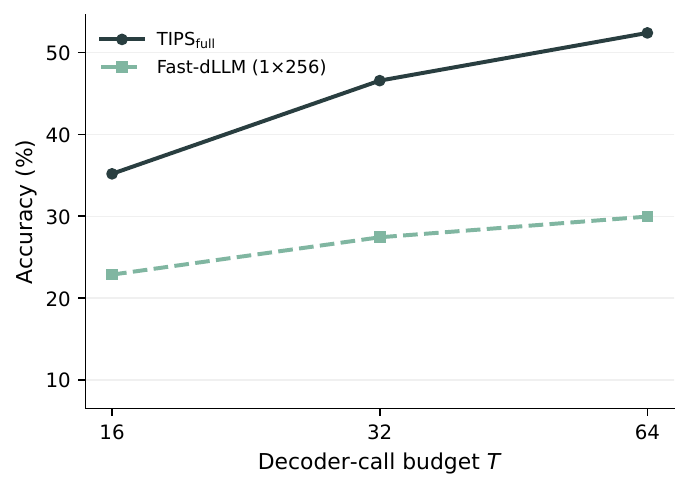}
\caption{\textbf{Zero-shot backbone transfer on a matched full canvas.} Test micro-accuracy for the frozen LLaDA-trained $\mathrm{TIPS}_{\mathrm{full}}$ policy and Fast-dLLM on Dream-7B. Both use $1\times256$ decoding. The horizontal axis is the decoder-call budget; actual baseline consumption is reported in Table~\ref{tab:app-exp-dream}.}
\label{fig:app-exp-dream}
\end{figure}
\FloatBarrier

\subsubsection{Zero-shot transfer from mathematics to coding}
\label{app:exp-coding-zero-shot}
We next apply the math-trained full-canvas and blockwise policies to coding without adaptation. Unlike the backbone-transfer experiment, this changes the task domain while retaining LLaDA. Table~\ref{tab:app-exp-coding-zero-shot} reports every retained baseline individually: UNCODE/PC-Sampler on $1\times256$, WINO on $1\times256$ and $2\times128$, and Non-adjusted Fast-dLLM on $1\times256$ and $8\times32$. These baselines retain their fixed mathematical-reasoning presets or assets rather than being tuned on coding test scores.

Transfer is mixed. At $T=16$, $\mathrm{TIPS}_{\mathrm{full}}$ achieves the highest measured HumanEval score among these methods, 16.46\%, but it does not maintain that advantage as the budget grows. At $T=64$, Non-adjusted Fast-dLLM with $8\times32$ decoding reaches 40.24\% on HumanEval, compared with 22.56\% and 29.27\% for the full and block TIPS policies. On MBPP, the two policies reach 29.00\% and 33.00\%, compared with 38.20\% for that baseline. Thus, the mathematical-reasoning policy transfers nontrivially, but the gains observed in its training domain do not translate into consistent superiority on coding.

\begin{table}[htbp]
\centering
\footnotesize
\setlength{\tabcolsep}{5pt}
\caption{\textbf{Zero-shot coding transfer.} TIPS uses frozen math-trained policies. Pass@1 (\%), one generation per test task, against each of the five retained baselines individually. Baseline presets/assets are inherited from mathematical reasoning. The $8\!\times\!32$ baseline uses a shared global cap. Adaptive methods may consume fewer than $T$ calls.}
\begin{tabular}{llrrrrrr}
\toprule
Method & Geometry & \multicolumn{3}{c}{HumanEval (164)} & \multicolumn{3}{c}{MBPP (500)} \\
\cmidrule(lr){3-5}\cmidrule(lr){6-8}
 &  & $T=16$ & $T=32$ & $T=64$ & $T=16$ & $T=32$ & $T=64$ \\
\midrule
UNCODE / PC-Sampler & $1\times256$ & 4.27 & 6.10 & 16.46 & 0.00 & 2.80 & 15.20 \\
WINO & $1\times256$ & 15.24 & 21.95 & 24.39 & 16.20 & 26.80 & 32.40 \\
WINO & $2\times128$ & 12.20 & 23.78 & 30.49 & 1.80 & 11.40 & 29.80 \\
Non-adjusted Fast-dLLM & $1\times256$ & 15.24 & 21.95 & 23.78 & 16.80 & 23.40 & 26.20 \\
Non-adjusted Fast-dLLM & $8\times32$ & 14.63 & 32.93 & 40.24 & 1.20 & 13.40 & 38.20 \\
$\mathrm{TIPS}_{\mathrm{full}}$ & $1\times256$ & 16.46 & 23.78 & 22.56 & 10.60 & 21.80 & 29.00 \\
$\mathrm{TIPS}_{\mathrm{block}}$ & $2\times128$ & 13.41 & 27.44 & 29.27 & 8.00 & 23.60 & 33.00 \\
\bottomrule
\end{tabular}

\label{tab:app-exp-coding-zero-shot}
\end{table}
\begin{figure}[htbp]
\centering
\includegraphics[width=0.86\linewidth]{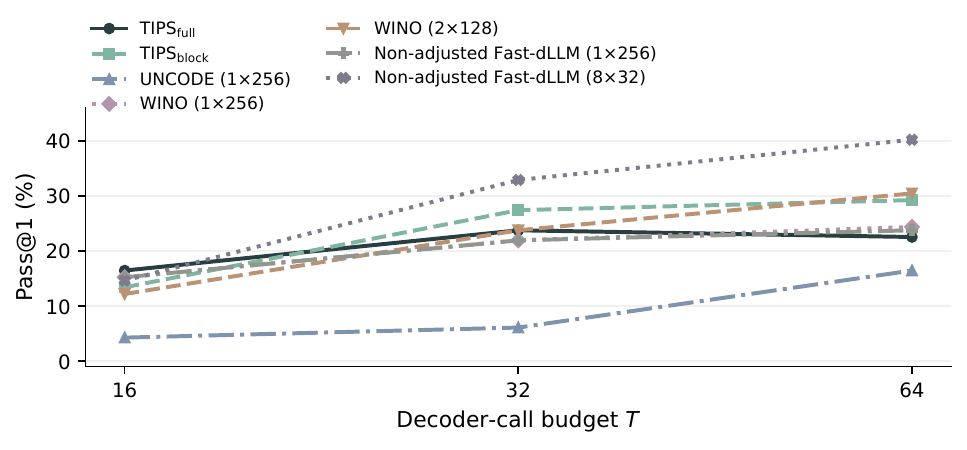}
\par\small (a) HumanEval
\medskip
\includegraphics[width=0.86\linewidth]{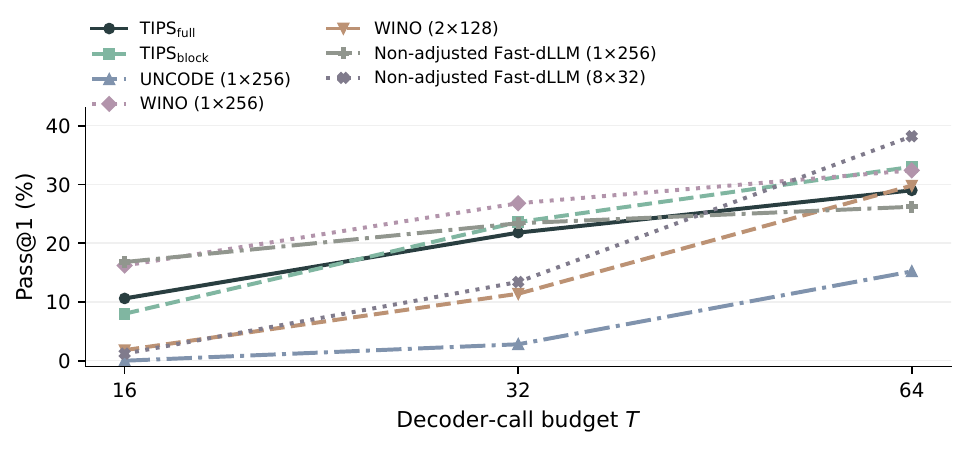}
\par\small (b) MBPP
\caption{\textbf{Zero-shot coding transfer against individual baselines.} One-generation pass@1 versus log decoder-call cap. TIPS uses its frozen math-trained policies.}
\label{fig:app-exp-coding-zero-shot}
\end{figure}
\FloatBarrier

\subsubsection{Training TIPS for coding}
\label{app:exp-coding-trained}
We separately train full-canvas and blockwise policies for coding while keeping the decoder frozen, using supervised warmup followed by an RLVR continuation to the fixed 935-update endpoint. Table~\ref{tab:app-exp-coding-trained} compares these policies with the same five baselines, so changes relative to the zero-shot table reflect the policy's domain adaptation.

Coding training improves the full-canvas policy in every measured dataset--budget condition. Its HumanEval pass@1 increases to 19.51\%, 25.61\%, and 27.44\%, while its MBPP score increases to 24.00\%, 31.80\%, and 37.80\%. The largest gains over the corresponding math-trained full-canvas policy occur on MBPP, reaching 13.40 percentage points at $T=16$. The coding-trained full policy exceeds every retained baseline on HumanEval at $T=16$ and on MBPP at $T=16$ and $32$. Nevertheless, Non-adjusted Fast-dLLM with $8\times32$ decoding remains stronger on HumanEval at $T=32$ and $64$, and slightly stronger on MBPP at $T=64$.

The block policy responds less consistently to coding training: it improves at $T=16$ on both tasks but falls below its zero-shot counterpart at $T=64$. These results support domain-specific policy training, especially for full-canvas decoding, while leaving room for a coding-specific recipe and hyperparameter study. We do not infer that the present coding configuration is optimal. Table~\ref{tab:app-exp-coding-compute} additionally reports actual mean calls, since adaptive baselines may finish below their requested caps.

\begin{table}[htbp]
\centering
\footnotesize
\setlength{\tabcolsep}{5pt}
\caption{\textbf{Coding-trained TIPS.} TIPS policies are trained separately on coding; the decoder remains frozen. Pass@1 (\%), one generation per test task, against each of the five retained baselines individually. Baseline presets/assets and TIPS settings are inherited from mathematical reasoning and untuned. The $8\!\times\!32$ baseline uses a shared global cap. Adaptive methods may consume fewer than $T$ calls.}
\begin{tabular}{llrrrrrr}
\toprule
Method & Geometry & \multicolumn{3}{c}{HumanEval (164)} & \multicolumn{3}{c}{MBPP (500)} \\
\cmidrule(lr){3-5}\cmidrule(lr){6-8}
 &  & $T=16$ & $T=32$ & $T=64$ & $T=16$ & $T=32$ & $T=64$ \\
\midrule
UNCODE / PC-Sampler & $1\times256$ & 4.27 & 6.10 & 16.46 & 0.00 & 2.80 & 15.20 \\
WINO & $1\times256$ & 15.24 & 21.95 & 24.39 & 16.20 & 26.80 & 32.40 \\
WINO & $2\times128$ & 12.20 & 23.78 & 30.49 & 1.80 & 11.40 & 29.80 \\
Non-adjusted Fast-dLLM & $1\times256$ & 15.24 & 21.95 & 23.78 & 16.80 & 23.40 & 26.20 \\
Non-adjusted Fast-dLLM & $8\times32$ & 14.63 & 32.93 & 40.24 & 1.20 & 13.40 & 38.20 \\
$\mathrm{TIPS}_{\mathrm{full}}$ & $1\times256$ & 19.51 & 25.61 & 27.44 & 24.00 & 31.80 & 37.80 \\
$\mathrm{TIPS}_{\mathrm{block}}$ & $2\times128$ & 16.46 & 26.83 & 21.95 & 13.00 & 23.60 & 32.40 \\
\bottomrule
\end{tabular}

\label{tab:app-exp-coding-trained}
\end{table}
\begin{figure}[htbp]
\centering
\includegraphics[width=0.86\linewidth]{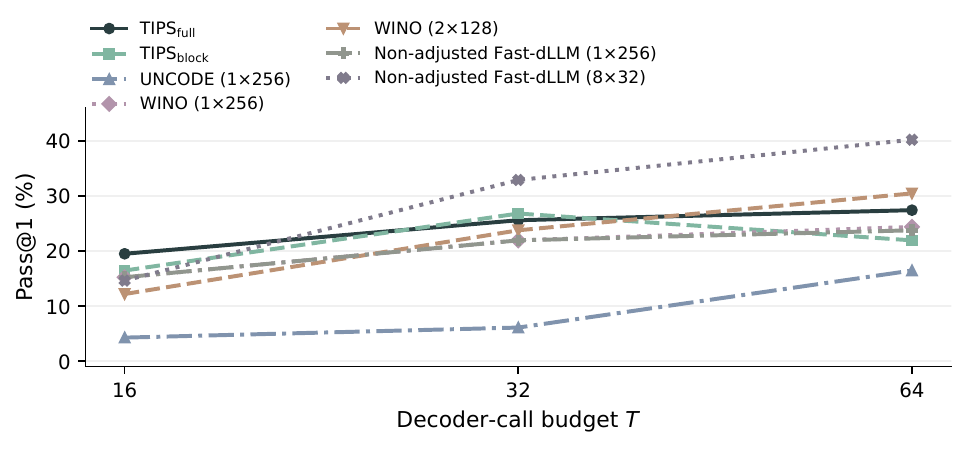}
\par\small (a) HumanEval
\medskip
\includegraphics[width=0.86\linewidth]{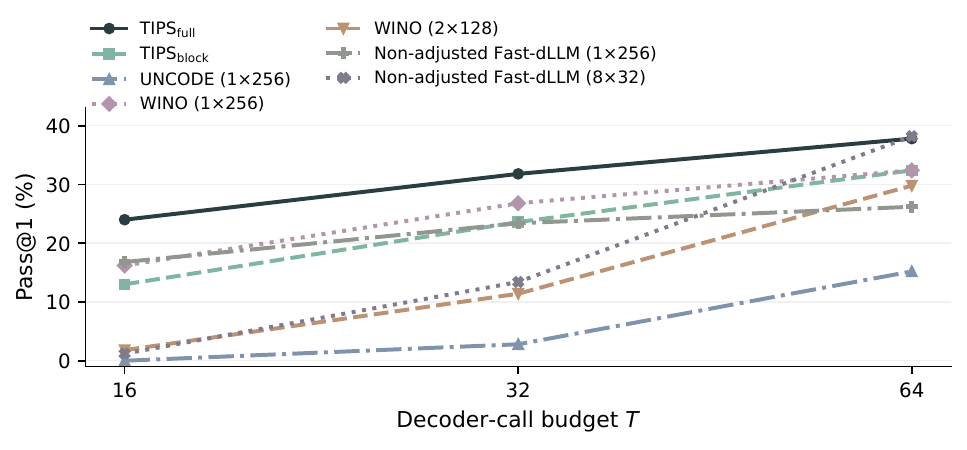}
\par\small (b) MBPP
\caption{\textbf{Coding-trained TIPS against individual baselines.} The decoder is unchanged; only the TIPS policies are trained for coding. Baselines are identical to those in Figure~\ref{fig:app-exp-coding-zero-shot}. Curves show one evaluation sample.}
\label{fig:app-exp-coding-trained}
\end{figure}
\FloatBarrier
\begin{table}[htbp]
\centering
\footnotesize
\setlength{\tabcolsep}{5pt}
\caption{\textbf{Consumed compute in coding experiments.} Mean decoder calls at the indicated caps, for the same conditions as Tables~\ref{tab:app-exp-coding-zero-shot} and~\ref{tab:app-exp-coding-trained}. Both math-trained and coding-trained TIPS use the exact requested horizon.}
\begin{tabular}{llrrrrrr}
\toprule
Method & Geometry & \multicolumn{3}{c}{HumanEval (164)} & \multicolumn{3}{c}{MBPP (500)} \\
\cmidrule(lr){3-5}\cmidrule(lr){6-8}
 &  & $T=16$ & $T=32$ & $T=64$ & $T=16$ & $T=32$ & $T=64$ \\
\midrule
UNCODE / PC-Sampler & $1\times256$ & 16.00 & 32.00 & 64.00 & 16.00 & 32.00 & 64.00 \\
WINO & $1\times256$ & 15.84 & 25.26 & 31.24 & 15.94 & 25.20 & 33.71 \\
WINO & $2\times128$ & 16.00 & 24.84 & 31.19 & 16.00 & 28.73 & 41.09 \\
Non-adjusted Fast-dLLM & $1\times256$ & 13.21 & 18.83 & 21.01 & 14.95 & 23.45 & 28.65 \\
Non-adjusted Fast-dLLM & $8\times32$ & 15.93 & 28.51 & 37.82 & 16.00 & 31.76 & 53.07 \\
TIPS (both policies) & native & 16.00 & 32.00 & 64.00 & 16.00 & 32.00 & 64.00 \\
\bottomrule
\end{tabular}

\label{tab:app-exp-coding-compute}
\end{table}
\FloatBarrier

\subsubsection{Zero-shot transfer across block sizes}
\label{app:exp-block-transfer}
We reuse the full-canvas-trained policy for target geometries $1\times256$, $2\times128$, $4\times64$, and $8\times32$, without retraining or fitting a geometry-specific schedule. The total response length remains 256 tokens, and the global decoder budget is divided equally across blocks, including one deterministic finalization per block. Consequently, increasing the number of blocks changes both the visible-context structure and the number of learned decisions available within each block.

Transfer is strongest when sufficient within-block compute remains available. At $T=64$, the original full-canvas and transferred two-block configurations reach 62.67\% and 62.89\% micro-accuracy, respectively; the four- and eight-block configurations retain 59.81\% and 55.25\%. At $T=16$, however, accuracy decreases from 47.00\% on the full canvas to 22.43\% with eight blocks. In the latter case, each block receives only two calls: one learned decision and one finalization. The results therefore show useful transfer to smaller blocks, particularly at larger budgets, while also exposing the cost of reducing the number of decisions per block.

\begin{table}[htbp]
\centering
\small
\setlength{\tabcolsep}{5pt}
\caption{\textbf{Frozen-policy block-size transfer.} All rows use the full-canvas-trained checkpoint, without block-specific policy training. Values are test accuracy (\%). The total 256-token response and total decoder-call budget remain fixed.}
\begin{tabular}{llrrr}
\toprule
Target geometry & Dataset & $T=16$ & $T=32$ & $T=64$ \\
\midrule
$1\times256$ & GSM8K & 57.39 & 71.72 & 74.75 \\
$1\times256$ & MATH-500 & 19.60 & 25.80 & 30.80 \\
$1\times256$ & Micro-average & 47.00 & 59.10 & 62.67 \\
$2\times128$ & GSM8K & 44.73 & 66.41 & 75.74 \\
$2\times128$ & MATH-500 & 15.80 & 25.80 & 29.00 \\
$2\times128$ & Micro-average & 36.78 & 55.25 & 62.89 \\
$4\times64$ & GSM8K & 27.90 & 57.01 & 72.02 \\
$4\times64$ & MATH-500 & 13.00 & 20.40 & 27.60 \\
$4\times64$ & Micro-average & 23.80 & 46.95 & 59.81 \\
$8\times32$ & GSM8K & 26.46 & 40.71 & 66.26 \\
$8\times32$ & MATH-500 & 11.80 & 16.40 & 26.20 \\
$8\times32$ & Micro-average & 22.43 & 34.03 & 55.25 \\
\bottomrule
\end{tabular}

\label{tab:app-exp-block-transfer}
\end{table}
\begin{figure}[htbp]
\centering
\includegraphics[width=0.65\linewidth]{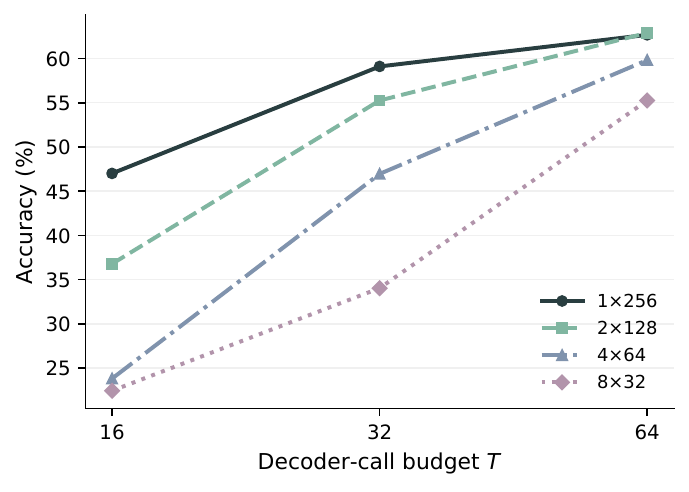}
\caption{\textbf{Frozen-policy transfer across decoding geometries.} Every curve uses the same full-canvas-trained checkpoint and the same global 256-token response length. The decoder-call budget is shared across blocks.}
\label{fig:app-exp-block-transfer}
\end{figure}
\FloatBarrier

\subsubsection{Sensitivity to decoder proposal temperature}
\label{app:exp-temperature}
We vary token-proposal temperature over $\{0,0.1,0.5,1.0\}$ for the frozen full-canvas policy, leaving policy-action sampling and the schedule unchanged. Finalization remains greedy in every condition. Small proposal noise is competitive: temperature 0.1 gives 48.32\% at $T=16$ and 63.33\% at $T=64$, slightly above the corresponding greedy scores of 47.00\% and 62.67\%, while greedy proposals are better at $T=32$. Increasing temperature to 1.0 reduces accuracy at all three budgets, with the largest loss at $T=16$ (37.77\% versus 47.00\%). These measurements support greedy decoding as a strong default.

\begin{table}[htbp]
\centering
\small
\setlength{\tabcolsep}{5pt}
\caption{\textbf{Sensitivity to proposal temperature.} Frozen $\mathrm{TIPS}_{\mathrm{full}}$ on the pooled math test set. Only token-proposal temperature changes; finalization remains greedy. Values are single-seed micro-accuracy (\%).}
\begin{tabular}{rrrr}
\toprule
Proposal temperature & $T=16$ & $T=32$ & $T=64$ \\
\midrule
0.0 & 47.00 & 59.10 & 62.67 \\
0.1 & 48.32 & 58.11 & 63.33 \\
0.5 & 46.89 & 56.02 & 62.62 \\
1.0 & 37.77 & 51.35 & 60.75 \\
\bottomrule
\end{tabular}

\label{tab:app-exp-temperature}
\end{table}
\begin{figure}[htbp]
\centering
\includegraphics[width=0.65\linewidth]{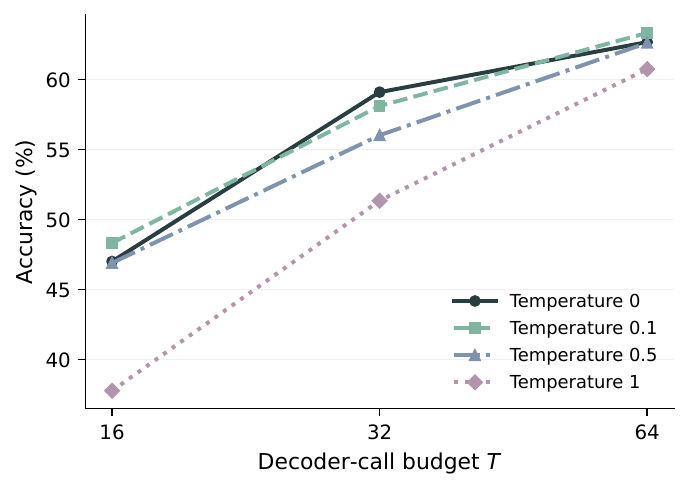}
\caption{\textbf{Sensitivity to proposal temperature.} Test micro-accuracy for the same frozen $\mathrm{TIPS}_{\mathrm{full}}$ policy. Only token proposals change temperature; finalization remains greedy, and visibility actions retain the same sampling rule.}
\label{fig:app-exp-temperature}
\end{figure}
\FloatBarrier
\subsection{Training diagnostics}
\label{app:exp-training-diagnostics}
\begin{figure}[htbp]
\centering
\includegraphics[width=0.98\linewidth]{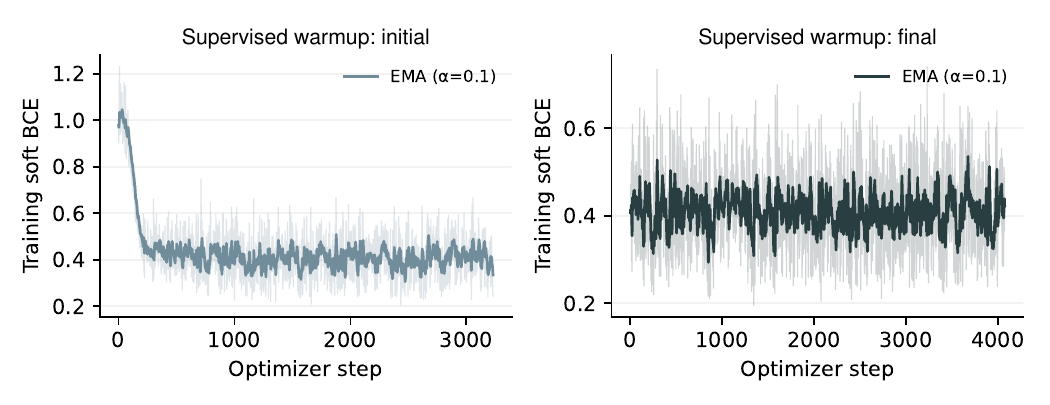}
\caption{\textbf{Supervised warmup on mathematics.} Recorded training soft binary cross-entropy in the initial and final warmup phases. Faint traces are raw observations; the darker curves are exponential moving averages with smoothing coefficient 0.1. The phases have separate optimizer-step axes.}
\label{fig:app-exp-warmup-math}
\end{figure}

We report training traces separately from downstream test performance. In the mathematical-reasoning run, supervised warmup comprises an initial 3,230-update phase followed by a final 4,074-update phase after refreshing the state dataset. Figure~\ref{fig:app-exp-warmup-math} shows the recorded optimized loss in each phase.

\begin{figure}[htbp]
\centering
\includegraphics[width=0.98\linewidth]{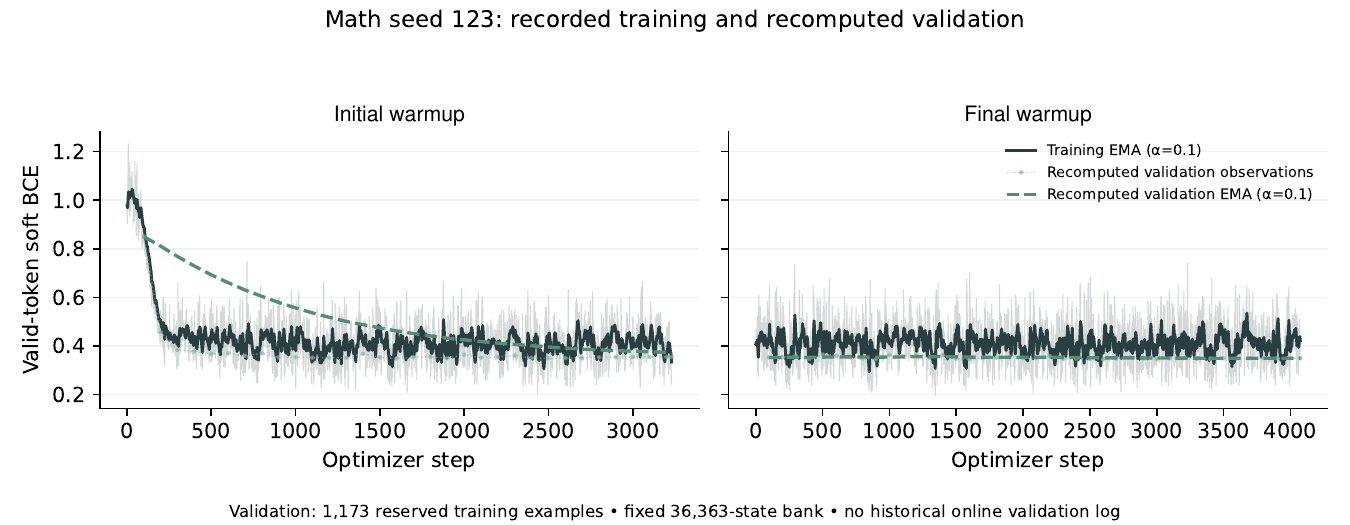}
\caption{\textbf{Recomputed warmup validation curves.} Recorded mathematical-reasoning training loss and validation observations recomputed on a fixed state bank, with exponential moving averages for visualization.}
\label{fig:app-exp-warmup-validation}
\end{figure}

\begin{figure}[htbp]
\centering
\includegraphics[width=0.98\linewidth]{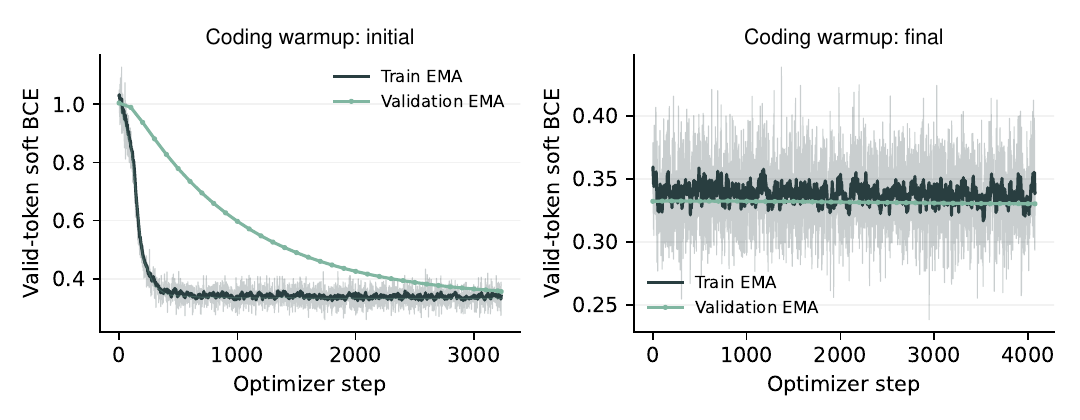}
\caption{\textbf{Supervised warmup on coding.} Supplied training and validation soft binary cross-entropy traces for the initial and final warmup phases. The displayed averages summarize within-run loss observations.}
\label{fig:app-exp-warmup-coding}
\end{figure}

Figure~\ref{fig:app-exp-warmup-validation} additionally reports recomputed mathematical-reasoning validation observations on a fixed dataset of 36,363 states from 1,173 reserved training examples. These are post-hoc diagnostic evaluations and use a different population from the 1,024-example validation ablations. Figure~\ref{fig:app-exp-warmup-coding} retains the coding run's recorded training and validation loss curves. All three figures concern supervised warmup of the small policy network.

\FloatBarrier

\begin{figure}[htbp]
\centering
\begin{minipage}[t]{0.49\linewidth}\centering
\includegraphics[width=\linewidth]{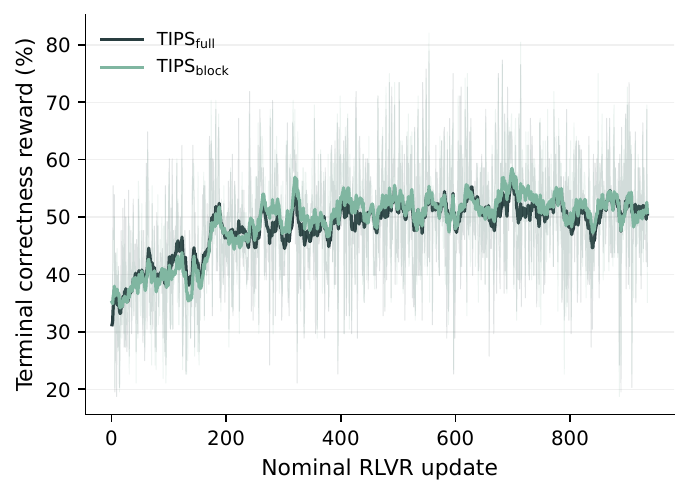}
\par\small (a) Mathematical reasoning
\end{minipage}\hfill
\begin{minipage}[t]{0.49\linewidth}\centering
\includegraphics[width=\linewidth]{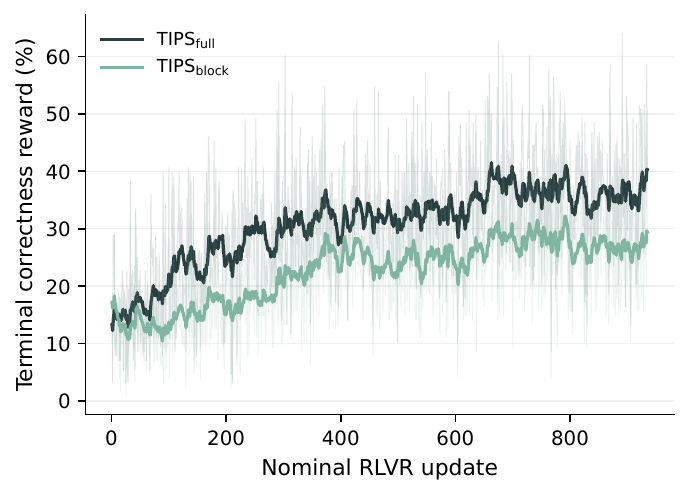}
\par\small (b) Coding
\end{minipage}
\caption{\textbf{RLVR training curves.} Terminal correctness reward per nominal update. Faint lines show the raw observations; darker lines show exponential moving averages with smoothing coefficient 0.1.}
\label{fig:app-exp-rlvr-training}
\end{figure}

\paragraph{RLVR trajectories. }
Figure~\ref{fig:app-exp-rlvr-training} shows raw terminal correctness rewards and exponential moving averages for the mathematical-reasoning and coding runs. Each nominal update contains 128 terminal outcomes, and each continuation ends at nominal update 935. Table~\ref{tab:app-exp-training-diagnostics} summarizes the first and last 50 updates without smoothing. For mathematics, mean reward increases from approximately 38\% to 50\% for both geometries. For coding, it increases from 16.28\% to 37.55\% for the full policy and from 13.66\% to 27.41\% for the block policy. In the coding full-canvas run, two nominal updates are skipped by the original optimizer rule, leaving 933 optimizer steps; the plot retains all 935 nominal updates and their recorded outcomes.

\begin{table}[htbp]
\centering
\small
\setlength{\tabcolsep}{5pt}
\caption{\textbf{RLVR training diagnostics.} Mean raw terminal-correctness reward (\%) over the first and last 50 recorded updates of each run.}
\begin{tabular}{llrr}
\toprule
Domain & Policy & First 50 updates & Last 50 updates \\
\midrule
Math & $\mathrm{TIPS}_{\mathrm{full}}$ & 38.19 & 50.73 \\
Math & $\mathrm{TIPS}_{\mathrm{block}}$ & 37.55 & 50.41 \\
Coding & $\mathrm{TIPS}_{\mathrm{full}}$ & 16.28 & 37.55 \\
Coding & $\mathrm{TIPS}_{\mathrm{block}}$ & 13.66 & 27.41 \\
\bottomrule
\end{tabular}

\label{tab:app-exp-training-diagnostics}
\end{table}
\FloatBarrier
\subsection{Illustrative decoding trajectories}
\label{app:exp-trajectories}
We show four trajectories with verifier-accepted final answers---one for each policy and mathematical-reasoning dataset---and one unsuccessful full-canvas trajectory. All use $T=32$. These examples were selected for readable outputs and visible ordinary-token edits. In particular, an accepted terminal answer does not certify every intermediate reasoning step or establish that a displayed edit caused the success.

The canvases retain the recorded token-level displays. A square denotes a masked position; a prefixed $+$ marks a reveal, a prefixed $-$ a remask, and a prefixed $*$ a changed re-reveal. These prefixes are display annotations. Each snapshot is labeled by the number of decoder calls already consumed, while active-block indices in the recorded canvases are zero-based. Deterministic finalizations are identified explicitly, and earlier finalized blocks remain frozen. The prompt instructions are omitted from the figures, while the task is given in each caption.
\clearpage

\begin{figure}[p]
\centering
\includegraphics[width=\linewidth,height=0.82\textheight,keepaspectratio]{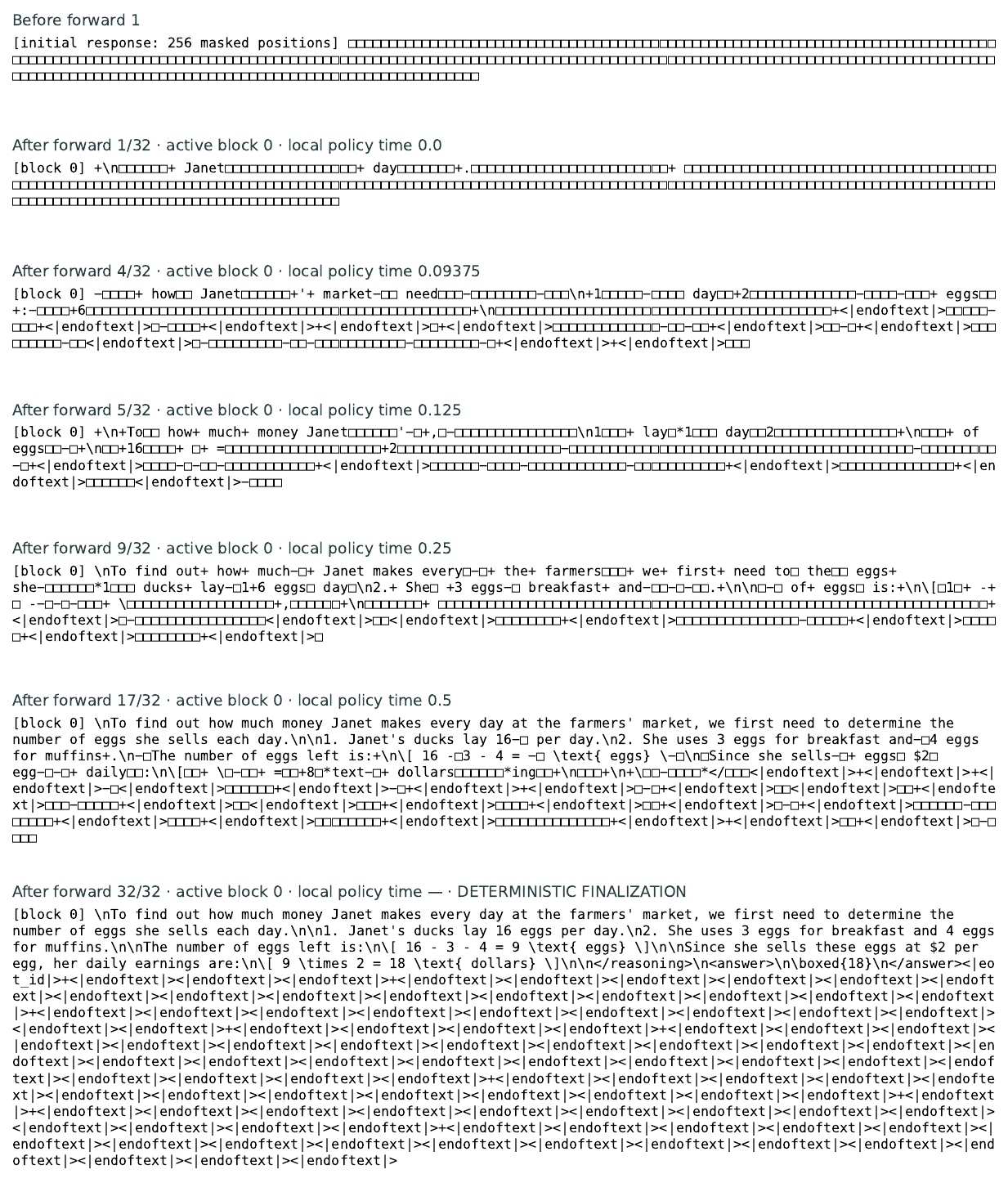}
\caption{\textbf{Accepted full-canvas trajectory on GSM8K.} Janet's ducks lay 16 eggs daily; she uses three for breakfast and four for muffins, then sells the remainder for two dollars each. The final response correctly computes $(16-3-4)\times2=18$ dollars. The snapshots show progressive visibility, remasking, and finalization.}
\label{fig:app-exp-trajectory-full-gsm8k}
\end{figure}
\clearpage

\begin{figure}[p]
\centering
\includegraphics[width=\linewidth,height=0.82\textheight,keepaspectratio]{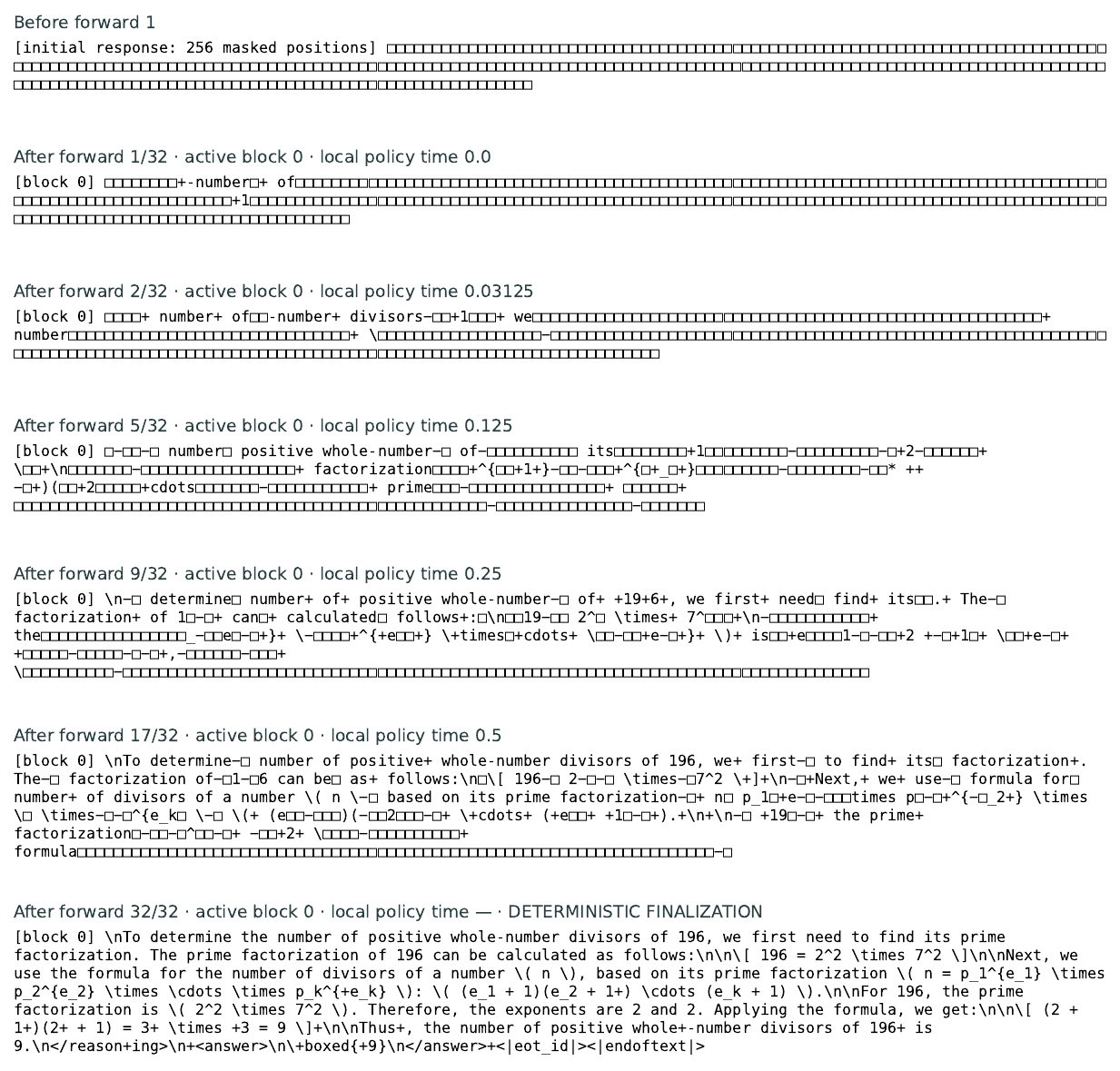}
\caption{\textbf{Accepted full-canvas trajectory on MATH-500.} The task asks how many positive whole-number divisors 196 has. The response arrives at $196=2^2\times7^2$ and the accepted answer $(2+1)(2+1)=9$. Token-level reveal and remask annotations are retained in the displayed intermediate and final canvases.}
\label{fig:app-exp-trajectory-full-math}
\end{figure}
\clearpage

\begin{figure}[p]
\centering
\includegraphics[width=\linewidth,height=0.82\textheight,keepaspectratio]{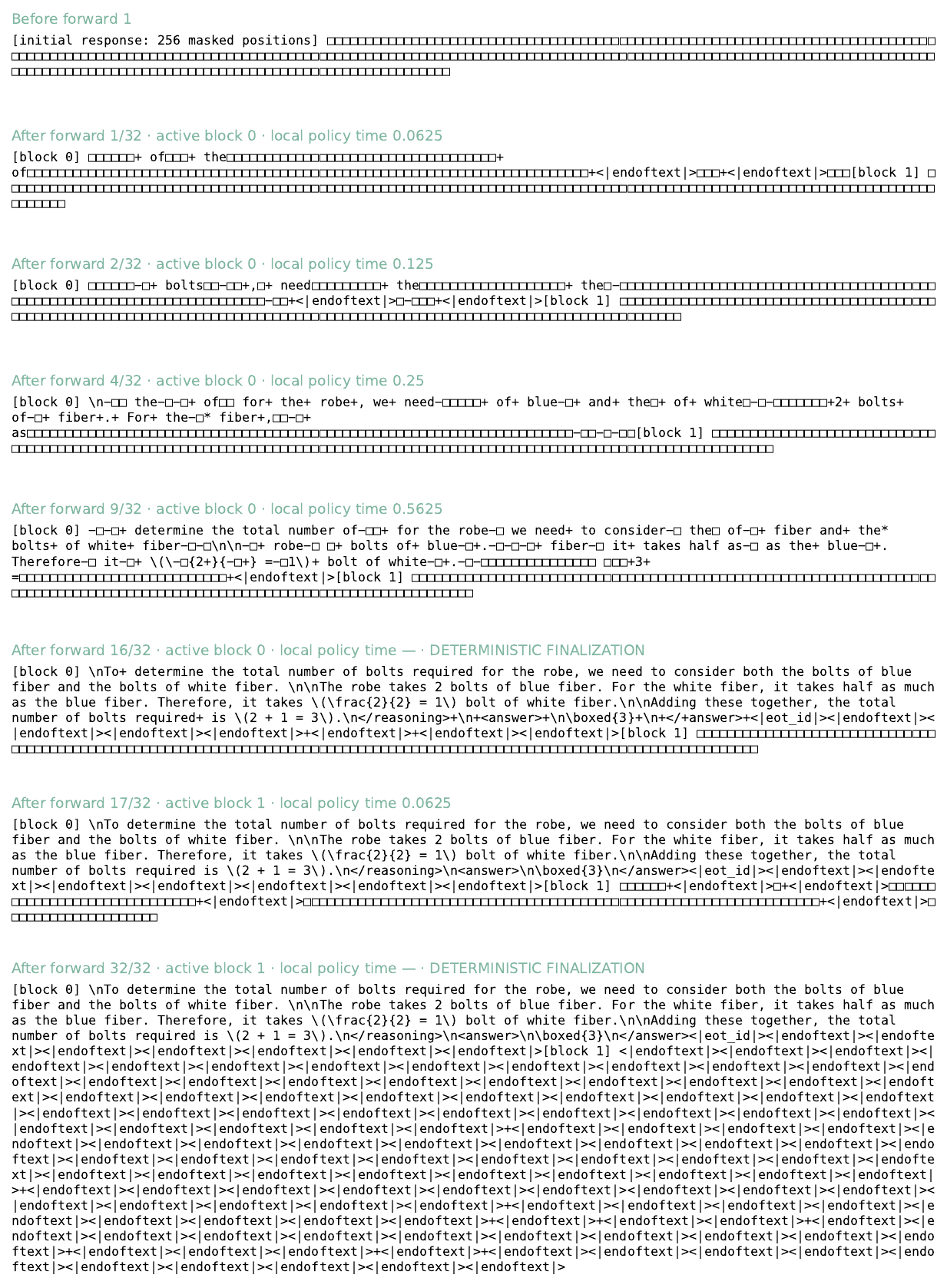}
\caption{\textbf{Accepted blockwise trajectory on GSM8K.} A robe requires two bolts of blue fiber and half as much white fiber, for a total of three bolts. The first block contains the solution and is finalized after 16 calls; the second block is then decoded under the same global budget. This example also illustrates that TIPS continues to the allocated horizon even when the answer-bearing text appears earlier.}
\label{fig:app-exp-trajectory-block-gsm8k}
\end{figure}
\clearpage

\begin{figure}[p]
\centering
\includegraphics[width=\linewidth,height=0.82\textheight,keepaspectratio]{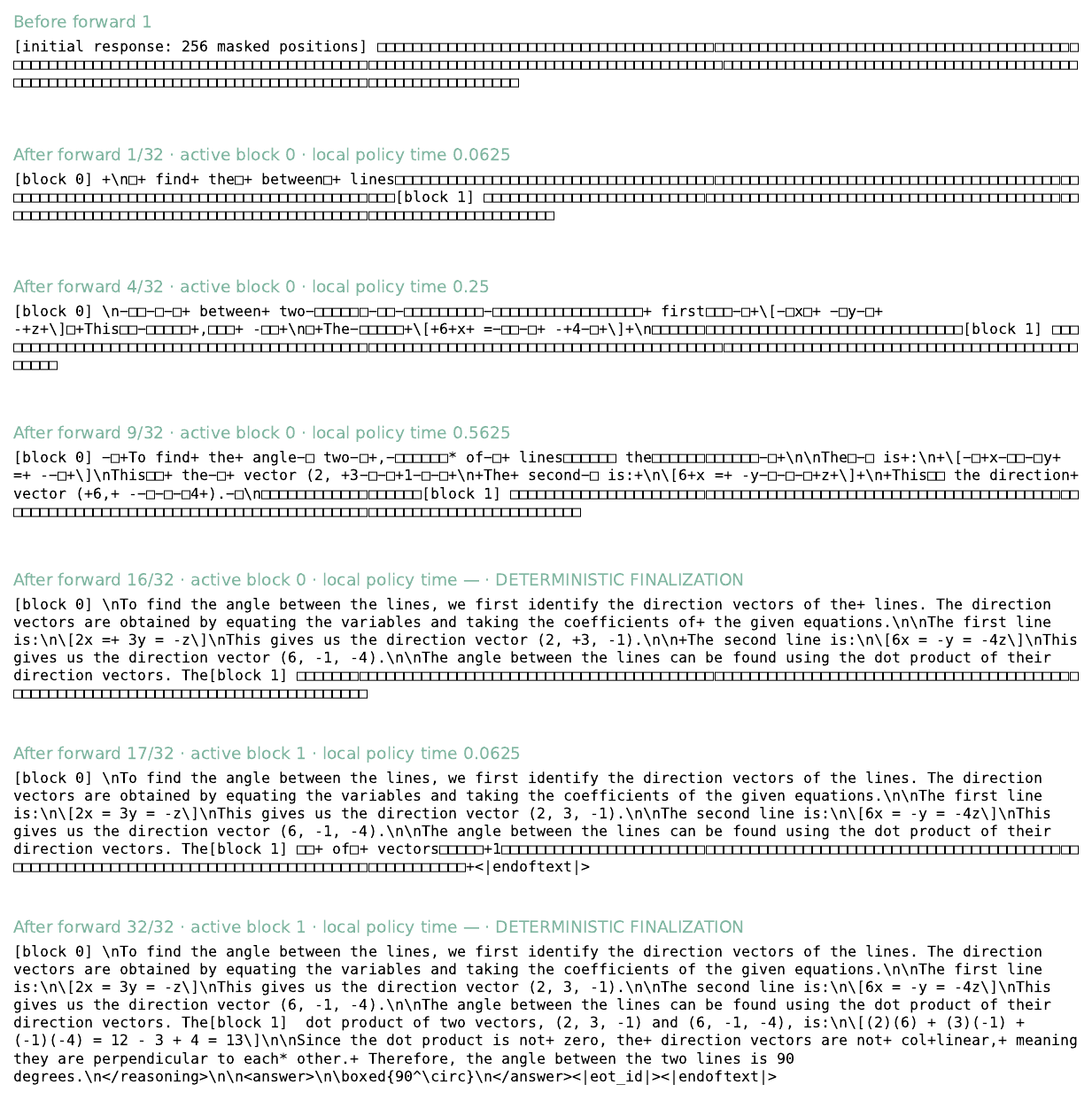}
\caption{\textbf{Verifier-accepted blockwise answer with flawed intermediate reasoning.} The task asks for the angle between the lines $2x=3y=-z$ and $6x=-y=-4z$. The final answer of $90^\circ$ is accepted, but the displayed derivation uses incorrect direction vectors and an invalid inference from their dot product. We retain the example to distinguish final-answer verification from verification of the reasoning trace.}
\label{fig:app-exp-trajectory-block-math}
\end{figure}
\clearpage

\begin{figure}[p]
\centering
\includegraphics[width=\linewidth,height=0.82\textheight,keepaspectratio]{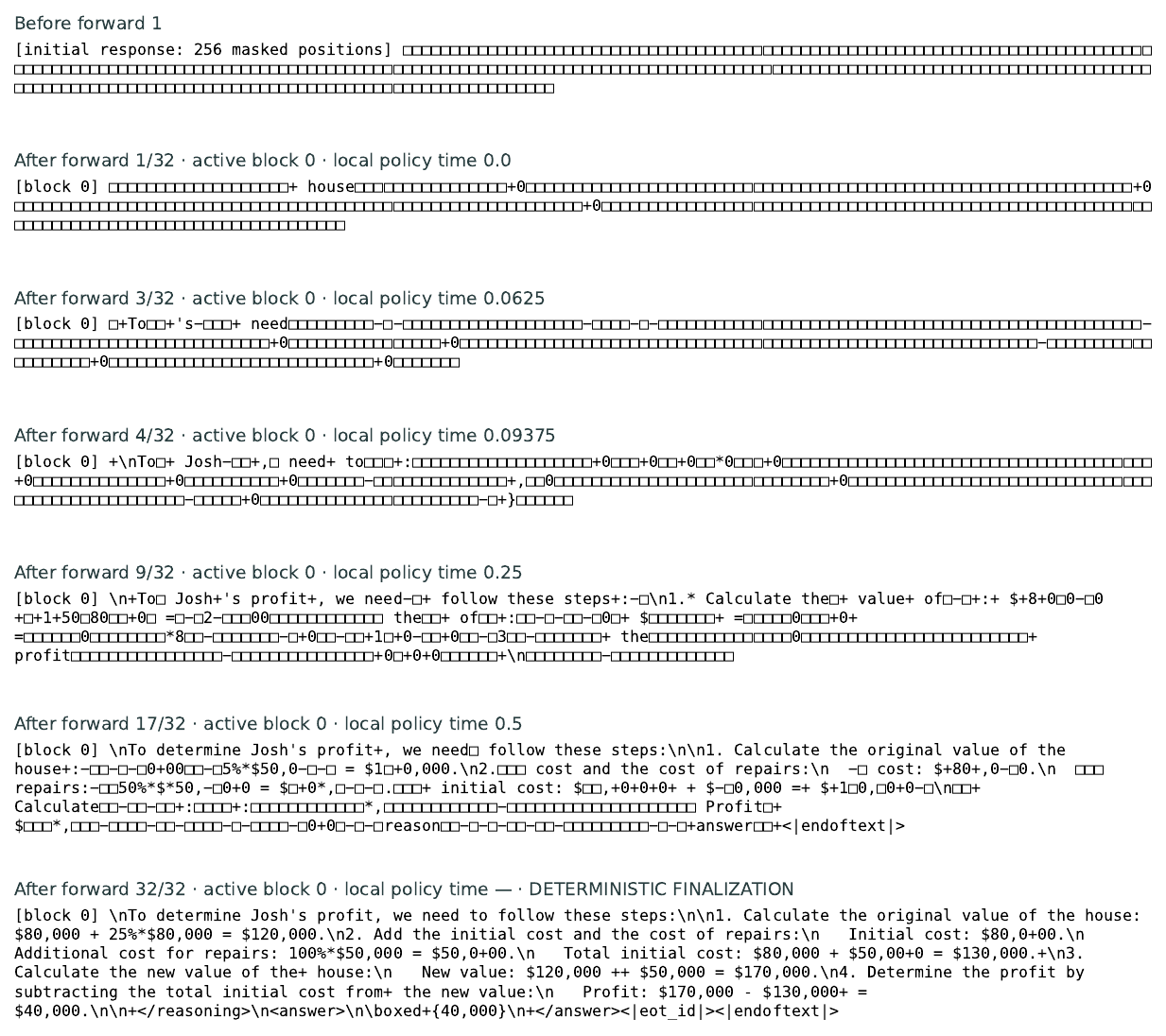}
\caption{\textbf{Unsuccessful full-canvas revision on GSM8K.} Josh buys a house for \$80,000, spends \$50,000 on repairs, and increases its value by 150\%. Despite repeated visibility changes, the response ends with an incorrect calculation and a generated answer of \$40,000. The trajectory demonstrates that remasking is an opportunity to revise, not a guarantee of correct reasoning or a correct final answer.}
\label{fig:app-exp-trajectory-failure}
\end{figure}
\clearpage
\section{Experimental and implementation details}
\label{app:implementation-details}

This section specifies the architecture, numerical implementation, and training procedure for the mathematical-reasoning checkpoints used in the main comparison. We distinguish the full-canvas policy, $\mathrm{TIPS}_{\mathrm{full}}$, from the two-block policy, $\mathrm{TIPS}_{\mathrm{block}}$, while identifying the supervised initialization shared by the two variants. Evaluation populations, compute accounting, and statistical summaries are specified in \cref{app:evaluation-protocols}.

\subsection{Frozen decoder and policy architecture}
\label{app:policy-network-architecture}

We use the frozen LLaDA-8B-Instruct decoder \citep{nie2025llada}, identified as \texttt{GSAI-ML/LLaDA-8B-Instruct}. Only the visibility controller is optimized: decoder parameters are excluded from the optimizer, and the recorded decoder weights remain unchanged throughout training.

The controller is a two-layer RoPE-DiT \citep{peebles2023scalable,su2024roformer} with hidden width 128, two attention heads per layer, and feed-forward width 512. A linear projection maps each two-dimensional confidence feature to the hidden width. A 128-dimensional sinusoidal time embedding is processed by a two-layer multilayer perceptron with SiLU activation, and a learned two-state embedding distinguishes masked from visible positions. The DiT layers use time- and state-conditioned adaptive layer normalization, bidirectional self-attention over response-position inputs, and GELU feed-forward activations. Rotary position embeddings are applied to attention queries and keys, with base $10{,}000$. A final layer normalization and scalar projection produce one visibility logit per position.

We use zero dropout and no separate conditioning on the absolute decoding budget. The adaptive normalization parameters are initialized to zero, the shared output bias is initialized to zero, and learned state-dependent logit offsets are initialized to $-2$ for masked positions and $+4$ for visible positions. Each final controller has 628,739 trainable parameters; fixed buffers are excluded from this count. The two response blocks of $\mathrm{TIPS}_{\mathrm{block}}$ share controller weights, so response-block count does not multiply the number of parameters.

\begin{table}[htbp]
\centering
\small
\caption{Configuration of the final mathematical-reasoning policies. The architecture is shared; decoding geometry, local timing, and fitted schedule differ.}
\label{tab:app-policy-configuration}
\begin{tabular}{@{}lcc@{}}
\toprule
Setting & $\mathrm{TIPS}_{\mathrm{full}}$ & $\mathrm{TIPS}_{\mathrm{block}}$ \\
\midrule
Response geometry & $1\times256$ & $2\times128$ \\
DiT layers / hidden width & $2 / 128$ & $2 / 128$ \\
Attention heads / head width & $2 / 64$ & $2 / 64$ \\
Feed-forward width & 512 & 512 \\
Confidence channels & 2 & 2 \\
Time-embedding width & 128 & 128 \\
Dropout & 0 & 0 \\
Trainable policy parameters & 628,739 & 628,739 \\
Training decoder-call budget & 32 & 32 \\
Schedule power $\alpha=\exp(\gamma)$ & 1.1565491821 & 1.2060408032 \\
Schedule parameter $\gamma$ & 0.1454407285 & 0.1873429312 \\
\bottomrule
\end{tabular}
\end{table}

\subsection{Confidence features and visibility updates}
\label{app:decoding-implementation}

We compute the features in \cref{eq:tips-features} from the decoder distribution before applying the visibility transition. At a masked position, the active token is the greedy proposal; at a visible position, it is the token already stored on the canvas. The strongest alternative is obtained from the two highest-probability vocabulary entries: it is the second entry when the active token is first, and the first entry otherwise. Thus, alternative probabilities are not renormalized over a reduced vocabulary.

For numerical stability, feature calculations are performed in float32 with $\epsilon=10^{-6}$. The active probability is clipped to $[\epsilon,1-\epsilon]$ for its log-odds, whereas both probabilities are clipped only from below by $\epsilon$ for the log-ratio. The implementation uses the ordering returned by the vocabulary top-two operation without an additional tie-breaking rule.

All position-wise visibility decisions are applied synchronously. A newly revealed position receives its proposal from the pre-transition decoder evaluation; a visible position selected to remain visible retains its existing token, even when the decoder now prefers another token. At deterministic finalization, one fresh decoder evaluation fills the remaining masks greedily and leaves already visible tokens unchanged. This finalization does not invoke the policy or contribute an action likelihood.

\paragraph{Block-local execution.}
Let $B$ be the number of response blocks and $H=T/B$ the number of decoder calls allocated to each block. Each block receives $H-1$ learned transitions and one deterministic finalization. The prompt and finalized prefix remain immutable, while future blocks remain masked until they become active. For local learned step $j\in\{0,\ldots,H-2\}$, both variants target occupancy $((j+1)/H)^{\alpha}$ within the active block. The saved policies use different time inputs: $\mathrm{TIPS}_{\mathrm{full}}$ receives $j/H$, whereas $\mathrm{TIPS}_{\mathrm{block}}$ receives $(j+1)/H$, with the clock reset at each block. We preserve these checkpoint-specific conventions during evaluation.

\subsection{Numerical calibration of the visibility logits}
\label{app:logit-calibration}

For the set $\mathcal I$ of positions eligible for scheduling, define
\begin{equation}
 f(\delta)
 =\frac{1}{|\mathcal I|}\sum_{i\in\mathcal I}\sigma(b_i+\delta)
 -\bar\kappa,
 \qquad
 \bar\kappa=\operatorname{clip}(\kappa,10^{-6},1-10^{-6}).
 \label{eq:app-calibration-residual}
\end{equation}
The scalar shift is initialized as
$\delta^{(0)}=\operatorname{logit}(\bar\kappa)-|\mathcal I|^{-1}\sum_{i\in\mathcal I}b_i$.
We first perform ten float32 Newton iterations,
\begin{equation}
\begin{split}
 \delta^{(r+1)}
 &=\delta^{(r)}-
 \operatorname{clip}\!\left(
 \frac{f(\delta^{(r)})}{\max\{f'(\delta^{(r)}),10^{-8}\}},-10,10
 \right),\\
 f'(\delta)
 &=\frac{1}{|\mathcal I|}\sum_{i\in\mathcal I}
 \sigma(b_i+\delta)\bigl[1-\sigma(b_i+\delta)\bigr].
\end{split}
\label{eq:newton-calibration}
\end{equation}
For rows whose resulting absolute residual is nonfinite or at least $10^{-6}$, we select a safeguarded ten-iteration solution. Its initial bracket is
$[\operatorname{logit}(\bar\kappa)-\max_{i\in\mathcal I}b_i,
\operatorname{logit}(\bar\kappa)-\min_{i\in\mathcal I}b_i]$.
Each safeguarded iteration updates the bracket and chooses the lower-residual candidate between a clipped Newton proposal and the bracket midpoint. Both solution paths remain in the automatic-differentiation graph, so gradients pass through the selected calibration computation during training. The detached teacher solve used for supervised targets is a separate procedure.

\subsection{Supervised warmup}
\label{app:supervised-warmup-config}

\paragraph{Reference responses.}
The training input contains 12,776 problems: 6,280 from GSM8K and 6,496 from MATH. We use each problem's original dataset answer or solution as its reference response. Each reference view ends at the first EOS token, or has EOS appended when none is present, and is then capped at 256 tokens. All training problems are retained, including the 2,113 whose reference view exceeds the canvas length before truncation. Reference inclusion is not conditioned on acceptance by the generated-answer verifier. Padded positions are marked invalid for the direct supervised loss.

\paragraph{Teacher and loss.}
The teacher uses the reference-token utility $u_{t,i}$ in \cref{eq:warmup-utility}. The masked-position margin is computed by subtracting the strongest non-reference logit from the reference-token logit, which preserves the sign of the margin when the reference is the decoder's top prediction. In the implementation, valid-position utilities are transformed to $(u_{t,i}+1)/2$ and multiplied by 10 before applying the occupancy-matching offset. Consequently, the sharpness in the raw-utility convention of the main text is $\beta=5$.

Binary cross-entropy is averaged over reference-valid positions within each state and then equally over the states in the minibatch. Reference tokens and reference margins are used only to construct supervision; the student receives the decoder-derived confidence features, mask states, and time inputs.

\paragraph{Initial and refreshed state banks.}
Warmup has two successive stages, both using full-canvas decoding at $T_{\mathrm{train}}=32$. The first stage trains from a fresh initialization on an initial bank of 396,056 states. We then refresh the bank with 99,014 additional states and continue from the first-stage policy using the combined 495,070-state bank. The optimizer, learning-rate schedule, and local update counter restart for the second stage. Each stage visits its bank once, with seed-dependent packet order and within-packet state order. Thus, the two stages together comprise 7,304 optimizer updates.

\begin{table}[htbp]
\centering
\small
\caption{Training stages per seed. State presentations count supervised examples processed, not distinct training problems.}
\label{tab:app-training-stages}
\begin{tabular}{@{}lrrl@{}}
\toprule
Stage & Updates & State presentations & Initialization \\
\midrule
Initial supervised warmup & 3,230 & 396,056 & Fresh policy \\
Refreshed supervised warmup & 4,074 & 495,070 & Initial warmup endpoint \\
Full-canvas RLVR & 935 & --- & Refreshed warmup endpoint \\
Two-block RLVR & 935 & --- & Same warmup endpoint \\
\bottomrule
\end{tabular}
\end{table}

\subsection{RLVR and optimization settings}
\label{app:rlvr-configuration}

For each training seed, the final supervised policy initializes separate full-canvas and two-block RLVR runs. Each run uses a global batch of 128 trajectories, formed from 16 prompt groups with eight trajectories per prompt, and a decoding horizon of 32 calls. Rollouts and likelihood replay use microbatches of 16 trajectories. Terminal reward is binary final-answer correctness, and the advantage for a trajectory is its reward minus the mean reward of its eight-trajectory group. 

At each learned step, the action log-probability sums Bernoulli log-probabilities over the positions on which the policy acts. The new-to-old joint action-likelihood ratio is clipped to $[0.5,1.5]$ in the GRPO surrogate, which is averaged over learned steps and trajectories. Finalization contributes neither an action nor a likelihood term. Hence, a training rollout contains 31 learned transitions for the full-canvas policy and 30 for the two-block policy. Both final mathematical-reasoning branches complete 935 nominal updates and 935 optimizer steps. Each branch therefore uses $935\times128\times32=3{,}829{,}760$ example-level decoder evaluations for training rollouts; this count excludes bootstrap training and supervised state collection.

\begin{table}[htbp]
\centering
\small
\caption{Optimization settings.}
\label{tab:app-optimization}
\begin{tabular}{@{}lcc@{}}
\toprule
Setting & Supervised warmup & RLVR \\
\midrule
Optimizer & AdamW & AdamW \\
Peak learning rate & $3\times10^{-5}$ & $3\times10^{-5}$ \\
Adam moments & $(0.9,0.999)$ & $(0.9,0.99)$ \\
Adam numerical constant & $10^{-8}$ & $10^{-8}$ \\
Weight decay & 0 & 0.1; 0 for bias/norm \\
Learning-rate warmup & 100 updates & 100 updates \\
Subsequent schedule & Cosine decay & Cosine decay \\
Global gradient-norm clipping & 0.2 & 0.2 \\
Trajectories per prompt & --- & 8 \\
Global trajectory batch & --- & 128 \\
Likelihood-ratio clipping & --- & $[0.5,1.5]$ \\
\bottomrule
\end{tabular}
\end{table}

We evaluate the fixed 935-update endpoints for training seeds $123$, $124$, and $125$. The checkpoint indices $1$, $2$, and $3$ correspond to these seeds, respectively, for both geometries. Full-canvas and block policies with the same index share their supervised warmup ancestral weights, but their RLVR weights are trained separately.

\subsection{Baseline configurations}
\label{app:main-experiment-details}

The mathematical-reasoning comparison contains the configurations listed in \cref{tab:app-baseline-roster}. Multiple configurations of one method differ in response geometry or in the saved training or decoding variant; they are therefore retained as separate comparisons.

\begin{table}[htbp]
\centering
\small
\caption{Baseline configurations. A geometry denotes number of blocks $\times$ tokens per block. Separate entries within a row are distinct saved configurations. The $4\times32$ UPO configuration uses a 128-token response and is recorded only at $T=128$.}
\label{tab:app-baseline-roster}
\begin{tabular}{@{}p{0.31\linewidth}p{0.64\linewidth}@{}}
\toprule
Method & Configurations \\
\midrule
Random / uniform & $1\times256$ \\
High confidence & $1\times256$, fitted schedule; $1\times256$, linear schedule \\
Probability margin & $1\times256$ \\
Predictive entropy & $1\times256$ \\
Fast-dLLM & $8\times32$, mean-adjusted as described in \cref{sec:experimental-setup} \\
Non-adjusted Fast-dLLM & $1\times256$; $8\times32$ \\
EB-Sampler & $1\times256$ \\
UNCODE / PC-Sampler & $1\times256$ \\
KLASS & $1\times256$; $4\times64$ \\
DAPD & $1\times256$ \\
\citet{jazbec2026learning} & $1\times256$, expert-steering; $1\times256$, vanilla \\
UPO & $1\times256$; $8\times32$, block-trained; $4\times32$, 128-token response \\
SAS & $1\times256$; $8\times32$ \\
ReMDM & $1\times256$ \\
RCR & $1\times256$ \\
WINO & $1\times256$; $2\times128$ \\
CoRe & $1\times256$ \\
\bottomrule
\end{tabular}
\end{table}
\section{Evaluation protocols}
\label{app:evaluation-protocols}

We first specify the mathematical-reasoning evaluation of the main TIPS policies, including its validation/test separation and aggregation over trained policies. We then describe the populations, interventions, and uncertainty measures used in the supplementary experiments, whose statistical units differ from those of the main result.

\subsection{Datasets, validation split, and prompts}
\label{app:datasets-and-prompts}

The mathematical-reasoning test population is the union of all 1,319 GSM8K test problems and all 500 MATH-500 problems \citep{cobbe2021gsm8k,hendrycks2021math,lightman2024verify}. The 1,024-problem validation set contains 512 GSM8K and 512 MATH problems drawn from the training sources and excluded from the gradient-training input. The recorded training, validation, and test identifier sets are pairwise disjoint.

\begin{table}[htbp]
\centering
\small
\caption{Problem populations for the main mathematical-reasoning training and evaluation. The validation population is also used, without correctness labels, to fit the schedule.}
\label{tab:app-data-populations}
\begin{tabular}{@{}lrrr@{}}
\toprule
Population & GSM8K & MATH / MATH-500 & Total \\
\midrule
Gradient-training input & 6,280 & 6,496 & 12,776 \\
Validation and schedule fitting & 512 & 512 & 1,024 \\
Final mathematical-reasoning test & 1,319 & 500 & 1,819 \\
\bottomrule
\end{tabular}
\end{table}

The same validation prompts support both label-free schedule fitting and design ablations. Their equal task mixture also differs from the $1{,}319{:}500$ test mixture, so their absolute accuracies are not directly comparable to main-test accuracies.

\paragraph{Prompt construction.}
The main TIPS test prompts are zero-shot. We place the task instruction and question in one user message, separated by two newline characters, apply the tokenizer's chat template with \texttt{add\_generation\_prompt=True}, and append the literal prefix \texttt{<reasoning>}. The GSM8K instruction is:
\begin{quote}
\small
You are a math expert. You will be given a question to solve. Solve it step by step. Wrap the final answer in a \texttt{\textbackslash boxed\{\}}.\\
Respond in the following format:
\begin{verbatim}
<reasoning>
Your reasoning here
</reasoning>
<answer>
\boxed{...}
</answer>
\end{verbatim}
\end{quote}
The saved MATH-500 instruction has the same text, with a literal double-quotation mark and a newline after the closing \texttt{</answer>} tag. We retain this exact task-specific prompt rather than normalizing it after evaluation. The full tokenized prompt is preserved. The response canvas is a separate fixed-length sequence of 256 tokens.

\subsection{Decoder-call budgets and actual consumption}
\label{app:compute-accounting}
The main budgets are $T\in\{4,8,16,32,64,128,256\}$. For TIPS, these are exact horizons: each example consumes all $T$ calls, including deterministic finalization, even if an intermediate canvas is already fully visible. Full-canvas decoding uses $T-1$ learned transitions and one finalization; two-block decoding uses $T-2$ learned transitions and two finalizations. For example, the two-block policy still makes two learned transitions at $T=4$.

The horizontal coordinate in the main comparison denotes a decoder-call cap, not necessarily the number of calls consumed by an adaptive baseline. The displayed $8\times32$ Fast-dLLM configuration uses the mean-adjusted confidence threshold to match budget in expectation (refer to \cref{app:fastdlm-budget-matching}). Other naturally terminating decoding can consume fewer calls than its cap. We therefore distinguish capped accuracy comparisons from measurements plotted against mean actual decoder calls. Controller evaluations and numerical calibration do not add decoder calls, although their runtime is not represented by this metric. Equal decoder-call counts alone do not imply equal wall-clock time or floating-point operation counts across response geometries.

\subsection{Output processing and answer verification}
\label{app:answer-verification}

For the main TIPS evaluation, we decode the complete 256-token response canvas with \texttt{skip\_special\_tokens=True}, without an additional stop-string truncation. 
GSM8K reference targets are converted to finite numerical values, using the final \texttt{\#\#\#\#} target when a full dataset answer is supplied. MATH targets are normalized with the mathematical-answer parser, which extracts an answer from boxed, framed, or answer-tagged reference text when applicable. Generated responses are evaluated by the corresponding task-specific extraction and correctness functions. Each verification has a five-second timeout. Missing or nonfinite predictions, answer mismatches, timeouts, and verifier exceptions count as incorrect; they are not removed from the evaluation denominator.

\subsection{Randomness, aggregation, and uncertainty}
\label{app:evaluation-metrics}

For each TIPS geometry and budget, the main evaluation uses three trained policies, with seeds $123$, $124$, and $125$, and three inference streams per policy, with identifiers $42$, $43$, and $44$. Token proposals are greedy, whereas visibility actions are independent Bernoulli draws conditional on the policy probabilities. 

Let $c_{sri}(T)\in\{0,1\}$ denote correctness on problem $i$ for training seed $s$ and inference stream $r$, and let $N=1{,}819$. We first compute each trained policy's accuracy, averaging its three inference draws, and then average the three policy accuracies:
\begin{equation}
 A_s(T)=\frac{1}{3N}\sum_{r=1}^{3}\sum_{i=1}^{N}c_{sri}(T),
 \qquad
 \bar A(T)=\frac{1}{3}\sum_{s=1}^{3}A_s(T).
 \label{eq:app-main-accuracy}
\end{equation}
This is a problem-level micro-average over GSM8K and MATH-500, not an equally weighted average of their dataset-level accuracies. For the main table and shaded bands, we report the standard error across the three trained-policy means:
\begin{equation}
 \operatorname{SE}(T)
 =\frac{1}{\sqrt{3}}
 \left[\frac{1}{2}\sum_{s=1}^{3}
 \bigl(A_s(T)-\bar A(T)\bigr)^2\right]^{1/2}.
 \label{eq:app-main-se}
\end{equation}

\end{document}